\documentclass[10pt,twocolumn,letterpaper]{article}
\usepackage{cvpr}
\usepackage{booktabs}
\usepackage{makecell}
\usepackage{tabularx}
\usepackage{graphicx}
\usepackage{array}
\usepackage{multirow}
\usepackage{float}
\usepackage{etoc}
\usepackage{amsmath}
\usepackage{wrapfig}
\usepackage{adjustbox}
\usepackage{ragged2e}
\usepackage{graphbox}
\usepackage{tcolorbox}
\tcbuselibrary{skins}
\definecolor{citecolor}{HTML}{0071BC}
\definecolor{linkcolor}{HTML}{ED1C24}
\usepackage[pagebackref=false, breaklinks=true, letterpaper=true, colorlinks, citecolor=citecolor, linkcolor=linkcolor, bookmarks=false]{hyperref}

\usepackage[scaled]{helvet}

\renewcommand{\paragraph}[1]{\vspace{1.25mm}\noindent\textbf{#1}}

\newcolumntype{x}[1]{>{\centering\arraybackslash}p{#1pt}}
\newcolumntype{y}[1]{>{\raggedright\arraybackslash}p{#1pt}}
\newcolumntype{z}[1]{>{\raggedleft\arraybackslash}p{#1pt}}

\newlength\savewidth\newcommand\shline{\noalign{\global\savewidth\arrayrulewidth
  \global\arrayrulewidth 1pt}\hline\noalign{\global\arrayrulewidth\savewidth}}

\newcommand{\tablestyle}[2]{\setlength{\tabcolsep}{#1}\renewcommand{\arraystretch}{#2}\centering\footnotesize}

\definecolor{cvprblue}{rgb}{0.21,0.49,0.74}
\def\confName{CVPR}
\def\confYear{2026}

\title{Memorization in 3D Shape Generation: An Empirical Study}

\author{
Shu Pu$^{1}$ \quad Boya Zeng$^{1}$ \quad Kaichen Zhou$^{2}$ \quad Mengyu Wang$^{2}$ \quad Zhuang Liu$^{1}$ \\[0.5em]
$^{1}$Princeton University \quad $^{2}$Harvard University}

\begin{document}
\maketitle

\begin{abstract}
Generative models are increasingly used in 3D vision to synthesize novel shapes, yet it remains unclear whether their generation relies on memorizing training shapes. Understanding their memorization could help prevent training data leakage and improve the diversity of generated results.
In this paper, we design an evaluation framework to quantify memorization in 3D generative models and study the influence of different data and modeling designs on memorization. We first apply our framework to quantify memorization in existing methods.
Next, through controlled experiments with a latent vector-set (Vecset) diffusion model, we find that, on the data side, memorization depends on data modality, and increases with data diversity and finer-grained conditioning; on the modeling side, it peaks at a moderate guidance scale and can be mitigated by longer Vecsets and simple rotation augmentation.
Together, our framework and analysis provide an empirical understanding of memorization in 3D generative models and suggest simple yet effective strategies to reduce it without degrading generation quality. Our code is available at {\small \href{https://github.com/zlab-princeton/3d_mem}{\texttt{github.com/zlab-princeton/3d\_mem}}}.
\end{abstract}

\section{Introduction}
\label{sec:intro}

3D shape creation is crucial for various industries, including digital twins, robotics, and AR/VR. Since manual modeling is slow and labor-intensive, the community has developed a variety of AI-powered models for 3D shape generation~\citep{poole2022dreamfusion, lin2023magic3d, hong2023lrm, Genesis, yang2025holopart, huang2025midi}. Recently, this field has benefited from feed-forward diffusion models trained directly on 3D shape datasets~\citep{cheng2023sdfusion,zhao2023michelangelo,zhang2024clay,li2025triposg,Trellis,hunyuan3d2025hunyuan3d,wu2025direct3d2} and compact 3D latent representations~\citep{3DShape2Vecset, chen20253dtopia, chen2025dora}, which together enable fast and scalable synthesis of high-fidelity 3D assets.

\begin{figure}[t]
    \centering
    \includegraphics[width=\linewidth]{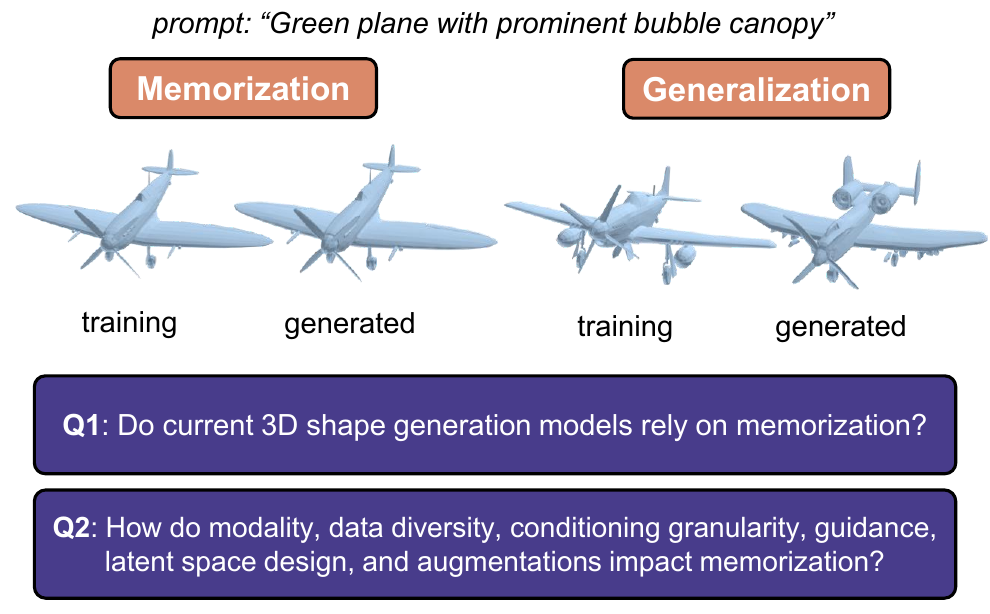}
    \caption{\textbf{Examples of generated 3D shapes that illustrate memorization \vs generalization relative to the training shapes.}
    In this paper, we propose a framework to evaluate memorization in 3D shape generation, use it to quantify memorization in existing methods, and conduct controlled experiments to study how data and modeling designs impact memorization.}
    \label{fig:teaser}
\end{figure}

Despite this rapid progress, the impressive performance of these models calls for closer examination. In particular, if these models achieve high fidelity by memorizing training shapes, they may risk training data leakage and fail to generalize to unseen geometries. Prior empirical studies of memorization in generative models have largely focused on image generation~\citep{somepalli2022diffusion,gu2023memorization,kadkhodaie2023generalization}. These studies investigate how memorization scales with factors such as dataset size and model capacity. However, memorization in 3D generative models has not yet been systematically studied. Given the complex geometric data and novel modeling factors in the 3D modality (\eg, 3D representations and rotation augmentations), it is a natural testbed for studying the impact of data and modeling designs that prior work has not explored.

Currently, in 3D shape generation, there is no standard metric to quantify the novelty of generated shapes relative to the training set. Existing evaluation metrics, such as Uni3D-Score~\citep{ma2023geodream} and Fréchet Point Cloud Distance (FPD)~\citep{shu20193d}, evaluate only shape fidelity and distributional similarity. In this work, we introduce an evaluation framework to quantify memorization in 3D generative models.

Concretely, we define \textit{object-level} memorization as generating shapes that are visually near-identical to training shapes. See Figure~\ref{fig:teaser} for an example. By benchmarking metrics commonly used for 3D object retrieval (3DOR) against human judgments, we find that Light Field Distance (LFD)~\citep{chen2003LFD} is the most accurate among the evaluated metrics for identifying replicas of training shapes. We then define \textit{model-level} memorization as the generated set being closer to the training set than a held-out test set is. Building on LFD, we propose an evaluation framework that employs the z-score $Z_U$ from a statistical test~\citep{mann1947test,meehan2020non} to quantify this memorization in practice, alongside Fréchet Distance (FD)~\citep{FID} to decouple it from generation quality.

Applying this framework to existing 3D generators, we find that models trained on smaller datasets (\eg, a single ShapeNet category)~\citep{hui2022wavelet, NFD, zheng2023las} show clear memorization, producing near-exact replicas, whereas recent large-scale conditional generative models~\citep{3DShape2Vecset,zhao2023michelangelo,Trellis} exhibit limited memorization and an enhanced ability to generalize.

To understand what drives memorization and to promote generalization, we conduct controlled experiments using the vector-set (Vecset) representation~\citep{3DShape2Vecset}, a latent shape representation widely used in recent state-of-the-art 3D generative models~\citep{zhang2024clay, li2024craftsman3d, wu2024direct3d, li2025triposg, hunyuan3d2025hunyuan3d}. Our results highlight multiple factors that affect a model's memorization. From the data perspective, rendered images are more likely to be memorized than 3D shapes in our setup; higher semantic diversity and finer-grained conditioning result in stronger memorization. From the modeling perspective, memorization is highest at a moderate classifier-free guidance scale, while increasing the latent Vecset length and applying simple rotation augmentation can effectively reduce it.

In summary, we introduce an evaluation framework to assess memorization in 3D generative models. Using this framework, we evaluate memorization in existing methods and design controlled experiments to study the influence of various data and modeling designs on memorization. Our findings suggest that specific strategies, such as applying rotation augmentation and increasing latent Vecset length, can help mitigate memorization while maintaining generation quality. We hope these insights offer valuable guidance for future research into generalizable 3D synthesis.

\section{Related Work}
\label{sec:related_work}

\paragraph{Shape generation in 3D vision.} 3D shape data have different representations, diverse structures, and conditional generation paradigms that condition on various modalities. Recently, 3DShape2VecSet~\citep{3DShape2Vecset} proposes Vecset, which enables training 3D generative models in a compact latent vector-set space and makes diffusion models easier to train and scale. As a result, many current state-of-the-art models are built on this representation~\citep{zhang2024clay, li2024craftsman3d, wu2024direct3d, li2025triposg, hunyuan3d2025hunyuan3d}.

\paragraph{Memorization in generative models.}
Motivated largely by copyright infringement concerns, a growing line of work studies data replication and memorization in generative models~\citep{somepalli2022diffusion, gu2023memorization,zeng2025generative}. These works typically analyze how factors such as dataset scale, model capacity, and training dynamics correlate with a model's tendency to generate samples nearly identical to its training data.
More recent studies~\citep{kadkhodaie2023generalization,song2025selective,bertrand2025closed} take a more mechanistic view of diffusion models, arguing that their ability to generate novel samples arises from structured inductive biases and from how they selectively fit or approximate the target score.

\paragraph{3D Object Retrieval (3DOR).} 3DOR is fundamental in AR/VR~\citep{tsai2023world,giunchi2018model}, the gaming industry~\citep{zuo2023peek}, and medical applications~\citep{li2023medshapenet}. Traditional 3D distance metrics, such as Chamfer Distance (CD) and LFD~\citep{chen2003LFD}, compute distances based on raw 3D geometry or renderings. Modern 3DOR methods are built on point cloud encoders~\citep{qi2017pointnet++,xue2024ulip,zhou2023uni3d} or multi-view encoders~\citep{su2015multi,hamdi2021mvtn}, with the latter often trained on relatively small rendered-image datasets. Recently, 3DOR has been explored in various settings~\citep{van2024finetuning, wang2025dac}.

\section{Methodology}
\label{sec:retrieval_methods}
To date, methods for quantifying and analyzing memorization in generative models of 3D shapes remain underexplored.
To study this memorization, we first identify an accurate retrieval method that can reliably find the closest training shape for each generated shape. Using this retrieval method, we employ the Mann-Whitney z-score $Z_U$~\citep{mann1947test,meehan2020non} to quantify memorization in generative models. Finally, we combine the z-score with a generation quality measure to establish an evaluation framework capable of distinguishing true generalization from low-quality generation.

\subsection{Distance Metrics}
\label{subsec:eval}

We construct a benchmark to identify the most effective distance metric for memorization detection. This benchmark contains 133 3D shapes generated in roughly equal numbers from four ShapeNet categories (\textit{chair}, \textit{car}, \textit{airplane}, and \textit{table}). We generate these shapes using models~\citep{hui2022wavelet, NFD} that, upon manual inspection, exhibit limited generalization.

For each generated shape and each distance metric, we retrieve the nearest neighbor from the training set, and we mark the retrieval as correct if the generated shape is visually near-identical to its retrieved shape according to manual inspection. Generated shapes for which none of the metrics retrieves a visually similar result are excluded from the final accuracy computation.
We evaluate seven widely used point cloud-based and view-based metrics on this benchmark. Implementation details are in Appendix~\ref{app:Dist_metric}.

\paragraph{Point cloud-based metrics} mainly assess geometric similarity by either directly computing distances between point clouds (\eg, Chamfer Distance) or comparing features produced by point cloud encoders (\eg, classical encoders like PointNet++~\citep{qi2017pointnet++} and image-text-aligned encoders like ULIP-2~\citep{xue2024ulip} and Uni3D~\citep{zhou2023uni3d}). Across all settings, we consistently sample 4096 points for each shape.
 
\paragraph{View-based metrics} mainly assess perceptual similarity by comparing images of 3D objects from multiple viewpoints, using either handcrafted image descriptors (\eg, LFD~\citep{chen2003LFD}) or image encoders (\eg, DinoV2~\citep{oquab2023dinov2} and SSCD~\citep{pizzi2022self}).

LFD renders object silhouettes from 10 canonical viewpoints. For each silhouette, it extracts a feature vector using handcrafted image descriptors~\citep{zhang2002integrated}. The distance between two shapes is then defined as the L1 distance between their feature vectors. For image encoders, we use SSCD and DinoV2 as suggested in~\citep{somepalli2022diffusion}. Following prior work~\citep{zheng2022sdf,Trellis}, we render 12 views per shape and obtain the final embedding by average-pooling the [CLS] tokens from each view.

\paragraph{Human evaluation results.}
The results in Table~\ref{tab:retrieval_accuracy} show that, among all seven retrieval metrics, LFD achieves the highest accuracy. Recent point cloud encoders such as Uni3D also deliver competitive performance, whereas image encoders trained on image datasets (\ie, DinoV2 and SSCD) still struggle to perform well in the 3D setting. Therefore, we adopt LFD as our primary retrieval metric. 

\begin{table}[htbp]
\centering
\tablestyle{11pt}{1.05}
\setlength{\tabcolsep}{2pt} 
\footnotesize               
\resizebox{\columnwidth}{!}{%
\begin{tabular}{@{}lccccccc@{}}
metric        & LFD        & Uni3D      & ULIP-2    & CD         & DinoV2     & PointNet++ & SSCD  \\
\shline
acc. {\scriptsize (\%)} & \textbf{78.4} & 74.8     & 66.2     & 46.8      & 20.1      & 18.7      & 7.2 \\
\end{tabular}%
}
\caption{\textbf{Top-1 retrieval accuracy (\%)} of distance metrics on 133 generated shapes from four ShapeNet categories (\textit{chair}, \textit{car}, \textit{airplane}, and \textit{table}). LFD achieves the highest accuracy.}
\vspace{-0.75em}
\label{tab:retrieval_accuracy}
\end{table}

\begin{table*}[t]
\vspace{-2.5em}
\centering
\small
\setlength{\tabcolsep}{1pt}
\renewcommand{\arraystretch}{1.1}
\begin{tabular}{@{}lccccc@{}}
\toprule
method
& LAS-Diffusion {\scriptsize (uncond.)}
& LAS-Diffusion {\scriptsize (class)}
& Wavelet Generation
& 3DShape2VecSet
& Michelangelo
\\
\midrule
$Z_U$
& -7.02
& -4.93
& -1.35
& 4.56
& 9.25
\\[0.2em]
\raisebox{5ex}{\parbox[t]{1.5cm}{\raggedright gen \&\\ retrieved}}
& {\includegraphics[width=0.1\textwidth, trim=0 80 0 80,clip]{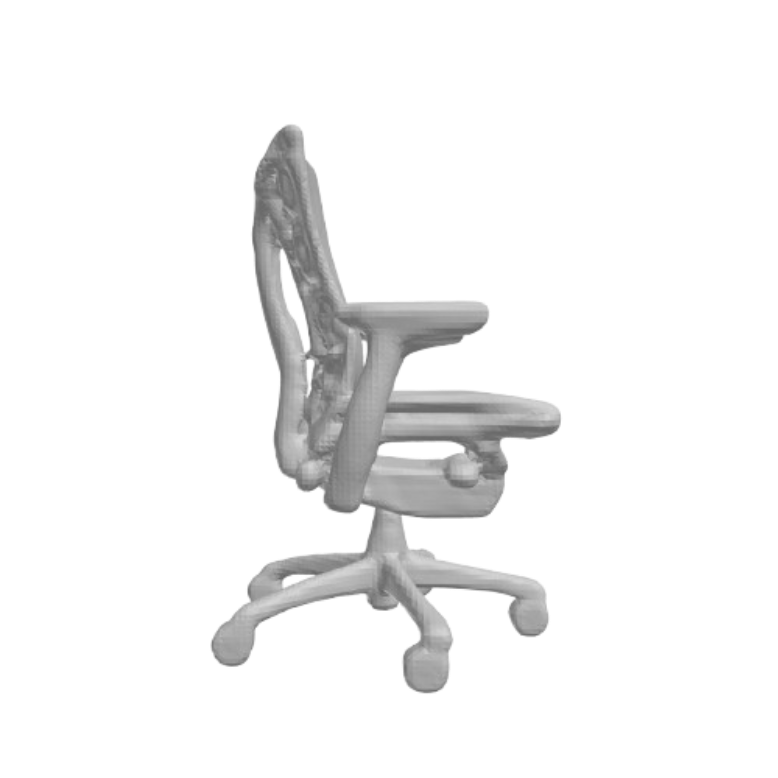}\hspace{-1.5em}
   \includegraphics[width=0.1\textwidth, trim=0 80 0 80,clip]{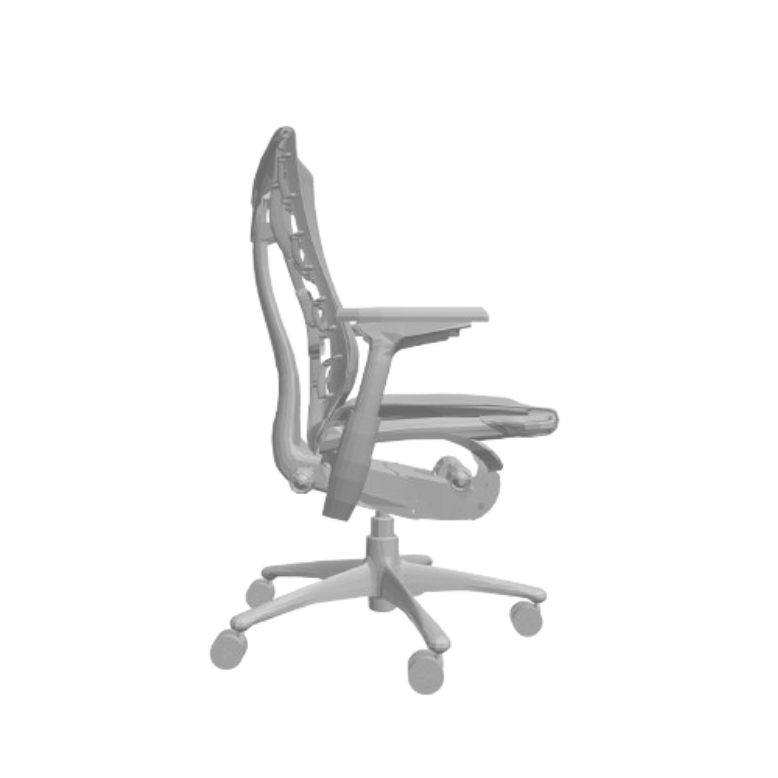}}
& {\includegraphics[width=0.1\textwidth, trim=0 80 0 80,clip]{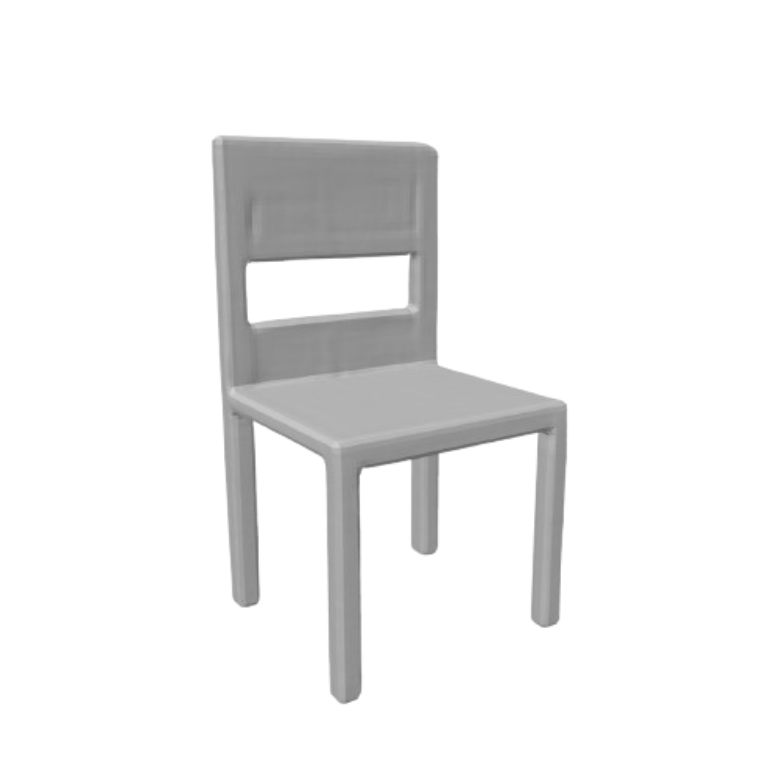}\hspace{-1.5em}
   \includegraphics[width=0.1\textwidth, trim=0 80 0 80,clip]{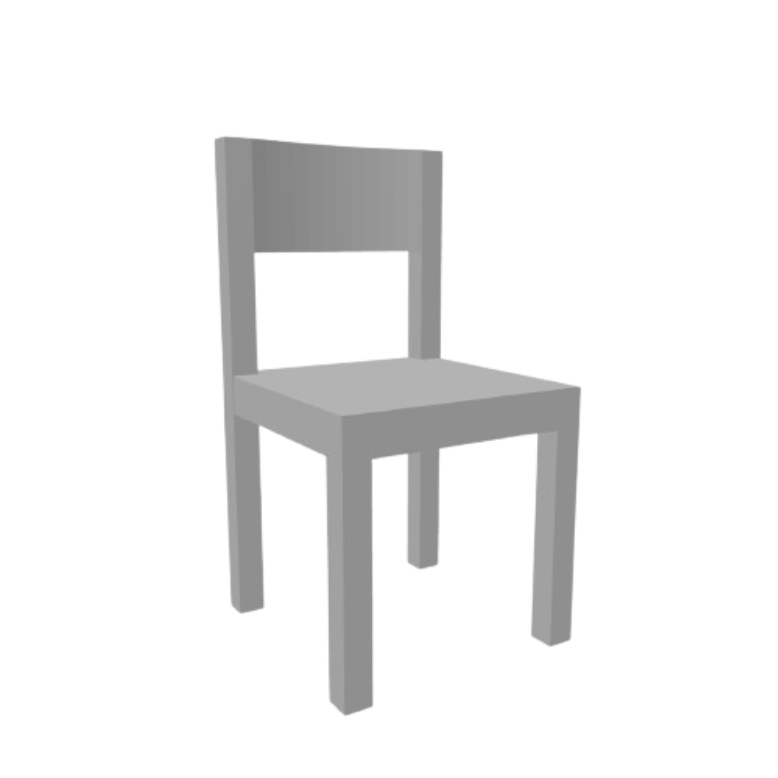}}
& {\includegraphics[width=0.1\textwidth, trim=0 80 0 80,clip]{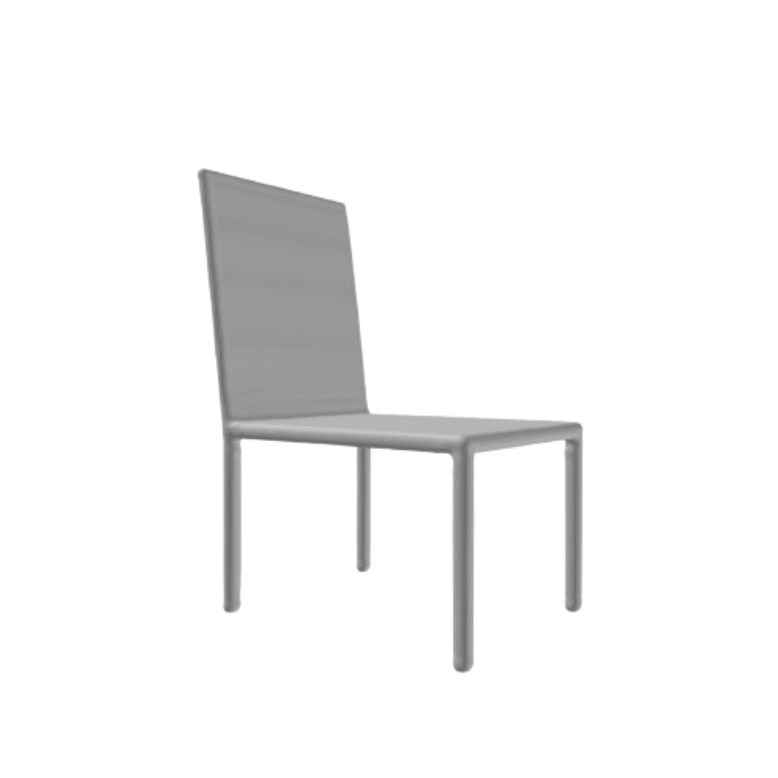}\hspace{-1.5em}
   \includegraphics[width=0.1\textwidth, trim=0 80 0 80,clip]{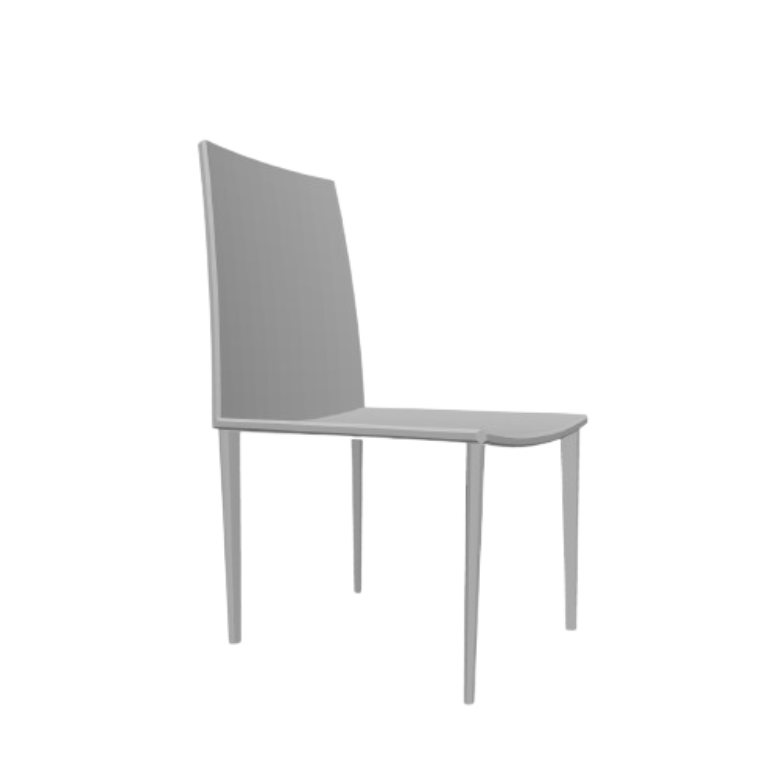}}
& {\includegraphics[width=0.1\textwidth, trim=0 80 0 80,clip]{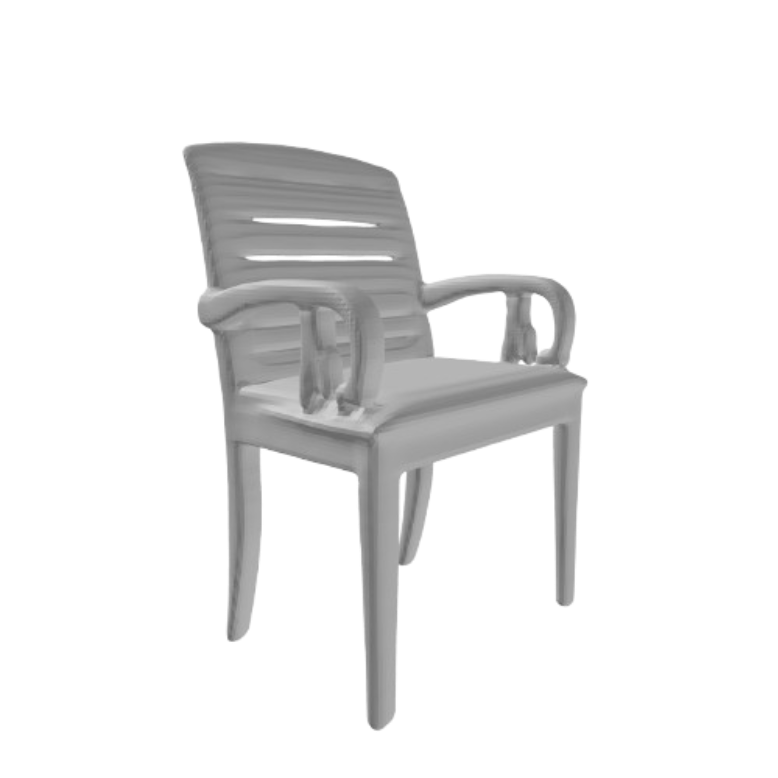}\hspace{-1.5em}
   \includegraphics[width=0.1\textwidth, trim=0 80 0 80,clip]{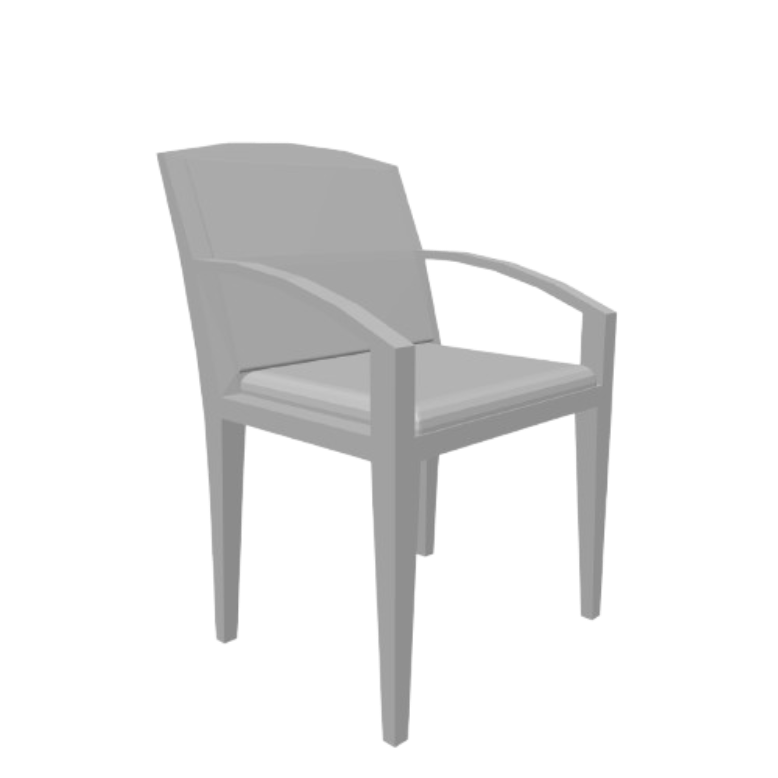}}
& {\includegraphics[width=0.1\textwidth, trim=0 80 0 80,clip]{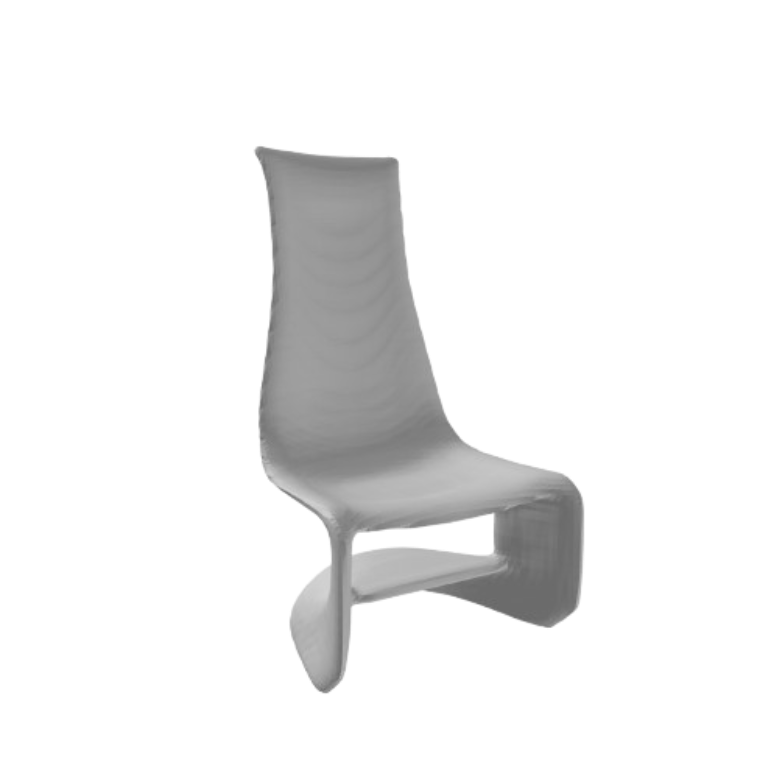}\hspace{-1.5em}
   \includegraphics[width=0.1\textwidth, trim=0 80 0 80,clip]{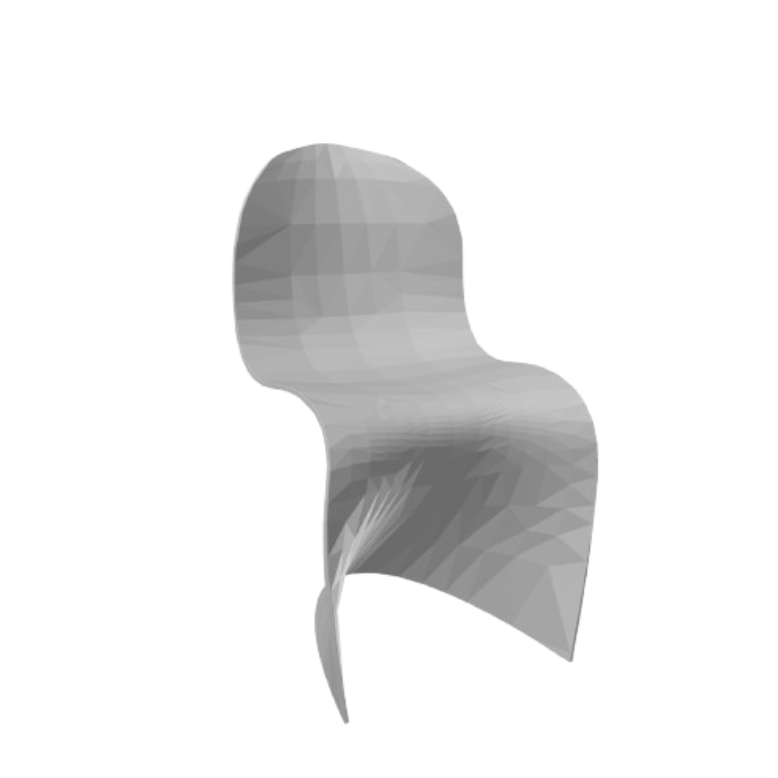}}
\\
\bottomrule
\end{tabular}
\caption{\textbf{Evaluating memorization on the \textit{chair} category in ShapeNet.}
We report $Z_U$ for each model with respect to the \textit{chair} subset of ShapeNet's training and test sets.
The last row shows, for each model, generated shapes at the $60^\text{th}$ percentile ranked by LFD distance to their nearest training shapes.
As $Z_U$ increases, generated samples transition from near-identical replicas of training shapes to clearly novel shapes.
For the text-conditional model Michelangelo, we use the prompt ``A 3D model of a chair'' to sample, following its training recipe.}
\label{tab:memorization-table}
\end{table*}

\begin{table*}[t]
    \centering
    \normalsize
    \setlength{\tabcolsep}{7.5pt}
    \begin{tabular}{@{}lcccccc@{}}
        \toprule
        method
        & LAS-Diffusion {\scriptsize (class)}
        & 3DShape2VecSet
        & Michelangelo
        & Trellis-small
        & Trellis-large
        & Trellis-xlarge
        \\
        \midrule
        split
        & IM-NET~\citep{chen2019learning}
        & 3DILG~\citep{zhang20223dilg}
        & 3DILG
        & Trellis500K~\citep{Trellis}
        & Trellis500K
        & Trellis500K
        \\
        $Z_U$
        & 0.46
        & 1.07
        & -0.33
        & -0.67
        & -1.57
        & -2.19
        \\
        \bottomrule
    \end{tabular}
    \caption{\textbf{Evaluating memorization on the entire training sets.} We report $Z_U$ for each model with respect to its entire training set and test set.
    For the text-conditional models (\ie, Michelangelo and Trellis), we use training prompts for generation. For Trellis, we randomly sample 100 shapes from the Trellis500K dataset as the test set in the $Z_U$ calculation.
    }
    \label{tab:existing-on-entire}
\end{table*}

\subsection{Memorization Metric} 
With LFD as our most accurate \textit{object-level} distance metric, we use it to define a \textit{model-level} memorization score based on the Mann-Whitney U test~\citep{mann1947test}.
Following prior work~\citep{meehan2020non}, this test takes the distances from generated and held-out samples to their nearest training shapes and outputs a single value that quantifies memorization.

Formally, let $T$ be the training set, $P_\text{test}$ the held-out test set, and $Q$ the generated set. Let $x \in P_\text{test}$ be a held-out test sample, and let $y \in Q$ be a generated sample. We define
\[
d_T(\,\cdot\,) = \min_{t \in T} \operatorname{dist}(\,\cdot\,,t),
\]
and construct two sets of distance values
\[
\mathcal{D}_{\text{test}} = \{d_T(x)\}_{x \in P_\text{test}}, \quad
\mathcal{D}_{\text{gen}} = \{d_T(y)\}_{y \in Q}.
\]
Following~\citep{meehan2020non}, we use the standardized z-score statistic $Z_U(\mathcal{D}_{\text{test}}, \mathcal{D}_{\text{gen}})$ ($Z_U$ for short) as our memorization score. A detailed explanation of $Z_U$ calculation is in Appendix~\ref{app:Metric_Def}.

In short, $Z_U$ quantifies memorization by ranking generated samples and held-out test samples based on their distances to their nearest training samples. When $Z_U < 0$, the generated set is, on average, closer to $T$ than $P_\text{test}$ is to $T$. Thus, we treat $Z_U < 0$ as evidence of memorization, with the strength of memorization increasing as $Z_U$ decreases.

Since $Z_U$ is based on the ranks of nearest-neighbor distances, it is directly comparable across models as long as $|P_\text{test}|$ and $|Q|$ are fixed. However, $Z_U \ge 0$ can come either from genuine generalization or from low-quality samples that are far from all training shapes. We therefore design an evaluation framework that controls for generation quality and makes memorization levels comparable across models.

\subsection{Evaluation Framework}
\label{subsec:eval_frame}
We use Fréchet Distance (FD)~\citep{FID}, a widely used metric for generative models, as our quality indicator to disentangle memorization from generation quality.
In Section~\ref{subsec:eval}, we find that Uni3D is the strongest point cloud encoder among the encoders we evaluate, so we use it to compute FD. Specifically, given a generated set $Q$ and a reference set $R$, we extract Uni3D features and fit Gaussian distributions to the features with means $\mu_R, \mu_Q$ and covariance matrices $\Sigma_R, \Sigma_Q$. The FD between these two distributions is:
\begin{equation}
\mathrm{FD} = \lVert \mu_R - \mu_Q \rVert_2^2
+ \mathrm{Tr}\Bigl(
\Sigma_R + \Sigma_Q- 2\bigl(\Sigma_R^{1/2}\,\Sigma_Q\,\Sigma_R^{1/2}\bigr)^{1/2}
\Bigr).
\end{equation}
We consider two variants of this metric: training FD and test FD. For training FD, we generate shapes from training prompts and compute FD with the training set; for test FD, we generate shapes from test prompts and compute FD with the test set. 
While training FD is more widely used~\citep{zheng2022sdf, Trellis}, test FD may be a fairer quality indicator, since training FD can be impacted by memorization: exactly reproducing the training data can drive training FD to zero.

With these components, we build an evaluation framework to decouple true generalization from low generation quality.
We use LFD to compute two sets of nearest-neighbor distances from generated shapes and test shapes to the training set, and use $Z_U$ to compare the two distance sets. To maintain concise notation, we hereafter refer to the memorization score calculated using LFD as $Z_U$ (formally $Z_U^{\text{LFD}}$), unless a different distance metric is specified.
We then interpret $Z_U$ only among models with similar test FD values, so that differences in $Z_U$ mainly reflect changes in memorization rather than changes in sample quality.

\section{Evaluating Existing Models}
\label{sec:exist}

We evaluate the memorization behavior of several representative models with fixed $|P_\text{test}|=|Q|=100$. The selected models span earlier models trained on a single ShapeNet category~\citep{hui2022wavelet, NFD}, models trained on the full ShapeNet dataset~\citep{3DShape2Vecset,zhao2023michelangelo}, and Trellis~\citep{Trellis} trained on 500K samples. We assume that these models already generate high-quality shapes, so we only evaluate their memorization level.

We first sample chair shapes from multiple models whose training data include the \textit{chair} category in ShapeNet, and compute $Z_U$ with respect to ShapeNet's \textit{chair} training and test sets. As shown in Table~\ref{tab:memorization-table} and Figure~\ref{fig:retrieval_chair} in Appendix~\ref{app:retrieval_res}, some models, such as LAS-Diffusion and Wavelet Generation, produce replicas of training shapes, whereas others, such as 3DShape2VecSet and Michelangelo, generate geometrically novel shapes. Models that qualitatively show stronger memorization also have lower $Z_U$, indicating that $Z_U$ correctly captures memorization.

We then evaluate recent state-of-the-art models using their full reported training and test datasets. For Trellis, which does not provide an official split, we construct a test set by randomly sampling and excluding data from the training set and calculate $Z_U$ accordingly. As shown in Table~\ref{tab:existing-on-entire}, recent 3D generative models have $Z_U$ values clustered around zero. Text-conditional models Michelangelo and Trellis have slightly negative $Z_U$ values. Consistent with the qualitative examples in Appendix~\ref{app:retrieval_res}, we find that Michelangelo and Trellis already demonstrate reasonable generalization capabilities.
In Appendix~\ref{app:retrieval_res}, we also provide additional qualitative results for recent models whose training and test sets are not available for computing $Z_U$.

In summary, we find that earlier models exhibit strong memorization, whereas recent models demonstrate effective generalization. This shift may result from various evolving design factors in the field, including the increasing data diversity and dataset size in 3D datasets, as well as modeling choices such as parameter scale and 3D representations.

\section{Impact of Data on Memorization}
\label{Sec:FactorExploration}

To understand how each design factor affects memorization in 3D models, we conduct controlled experiments based on our evaluation framework. In this section, we first investigate the impact of training data properties, including data modality, diversity, and conditioning granularity.

\begin{figure}[htbp]
    \centering
    \setlength{\tabcolsep}{1pt}
    \renewcommand{\arraystretch}{0}
    \begin{tabular}{ccc}
        \shortstack[c]{%
            \includegraphics[width=0.32\linewidth]{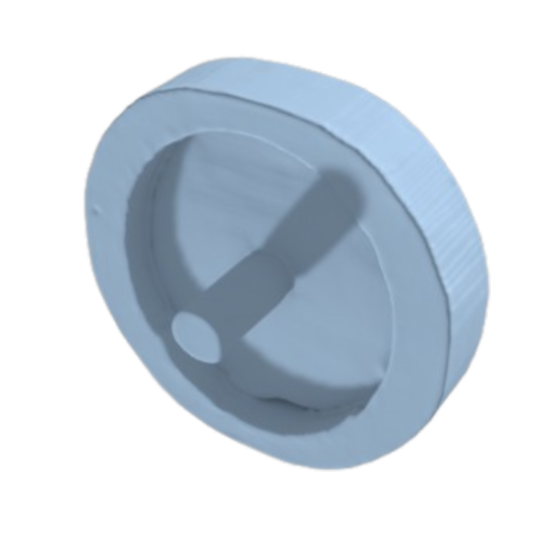}\\[-0.2ex]
            \scriptsize 10th (629)} &
        \shortstack[c]{%
            \includegraphics[width=0.32\linewidth]{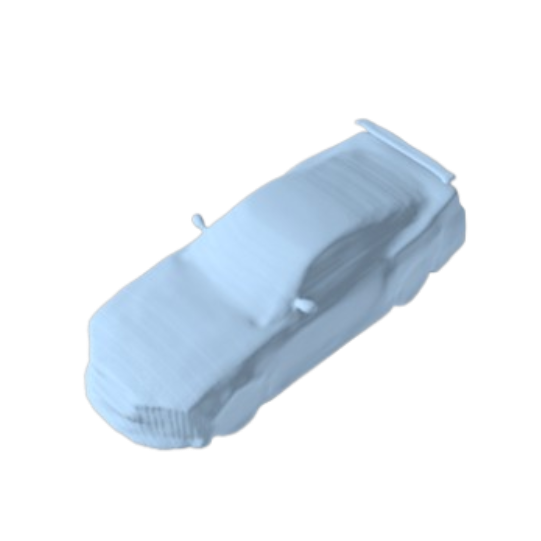}\\[-0.2ex]
            \scriptsize 20th (1031)} &
        \shortstack[c]{%
            \includegraphics[width=0.32\linewidth]{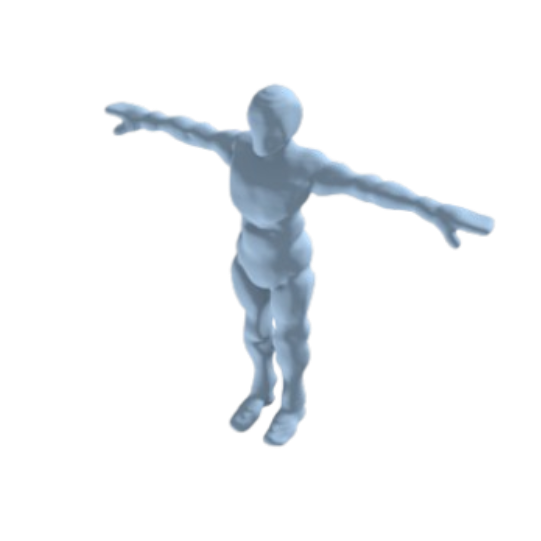}\\[-0.2ex]
            \scriptsize 30th (1302)} \\
        \shortstack[c]{%
            \includegraphics[width=0.32\linewidth]{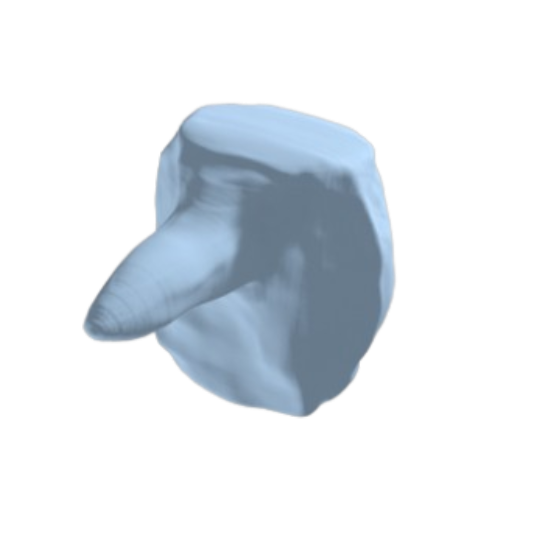}\\[-0.2ex]
            \scriptsize 40th (1753)} &
        \shortstack[c]{%
            \includegraphics[width=0.32\linewidth]{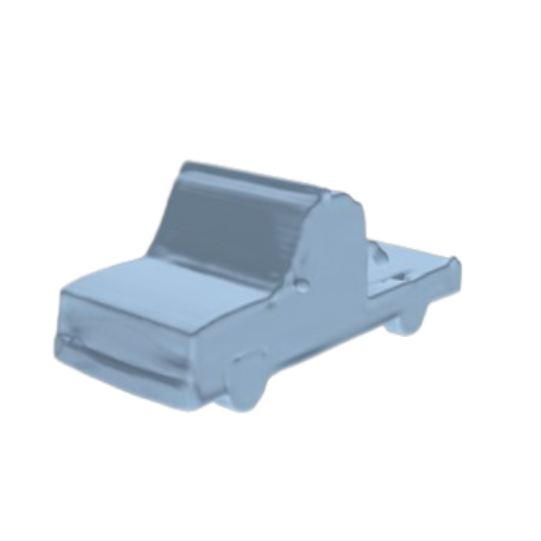}\\[-0.2ex]
            \scriptsize 50th (2146)} &
        \shortstack[c]{%
            \includegraphics[width=0.32\linewidth]{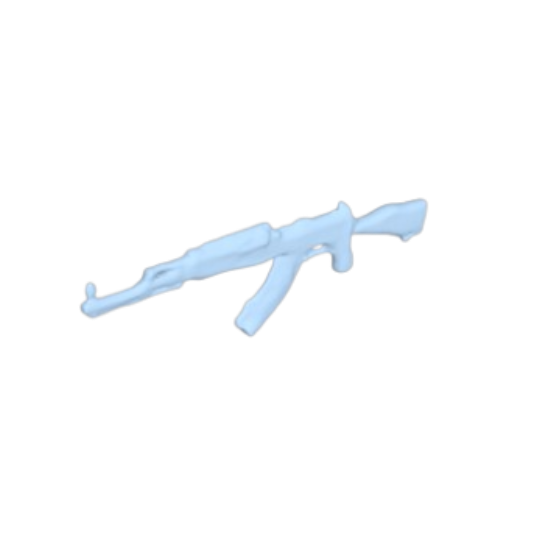}\\[-0.2ex]
            \scriptsize 60th (2528)} \\
        \shortstack[c]{%
            \includegraphics[width=0.32\linewidth]{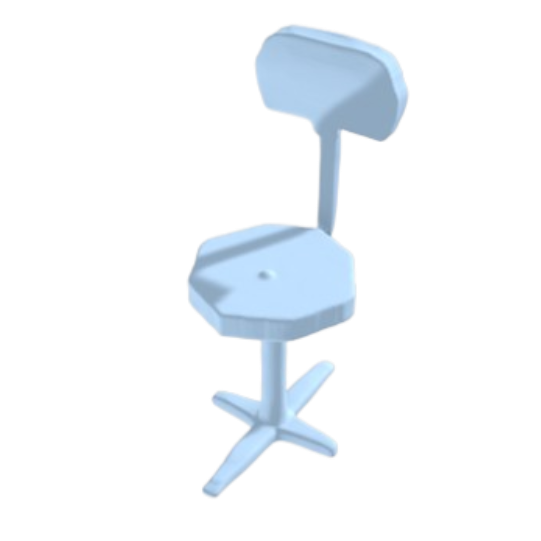}\\[-0.2ex]
            \scriptsize 70th (2936)} &
        \shortstack[c]{%
            \includegraphics[width=0.32\linewidth]{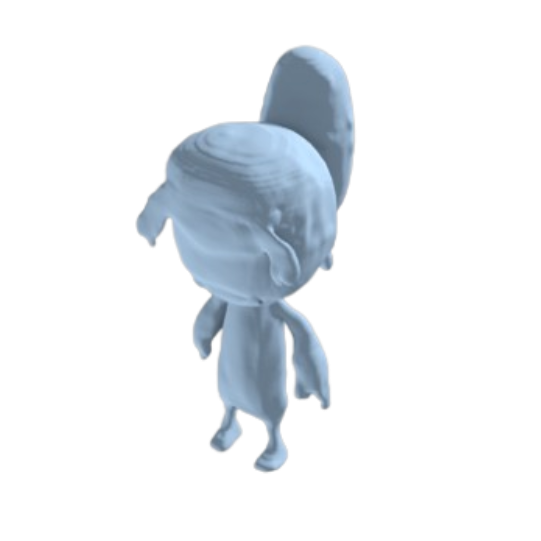}\\[-0.2ex]
            \scriptsize 80th (3417)} &
        \shortstack[c]{%
            \includegraphics[width=0.32\linewidth]{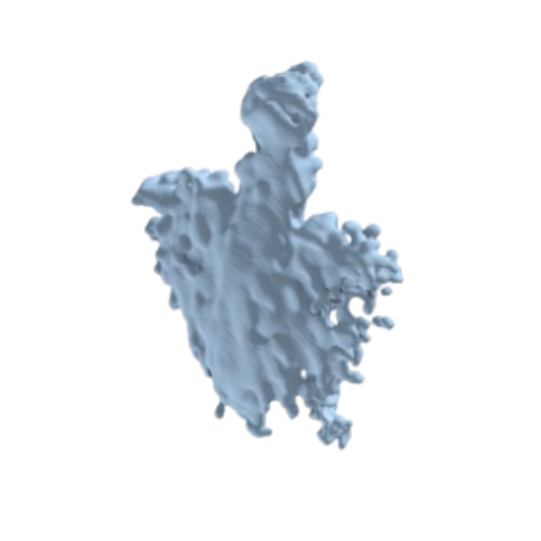}\\[-0.2ex]
            \scriptsize 90th (4225)} \\
    \end{tabular}
    \caption{\textbf{Generated samples from the baseline model} at each decile, ranked by LFD to the nearest training shape.}
    \label{fig:baseline_3x3}
    \vspace{-1em}
\end{figure}

\subsection{Experimental Setup}
\begin{table*}[t]
\centering
\vspace{-2.5em}

\tablestyle{10pt}{1.2}
\setlength{\tabcolsep}{6pt}  
\begin{tabularx}{\linewidth}{>{\raggedright\arraybackslash}c c c c c c c}

sampling prompt &
training data &
3D gen. &
top-1 NN {\scriptsize(SSCD)} &
top-1 NN {\scriptsize(LFD)} &
image gen. &
top-1 NN {\scriptsize(SSCD)} \\
\shline
\addlinespace[0.3em]

\makecell{plush toy character with\\ round head and small limbs} &
\adjustbox{valign=c}{\includegraphics[width=.09\linewidth]{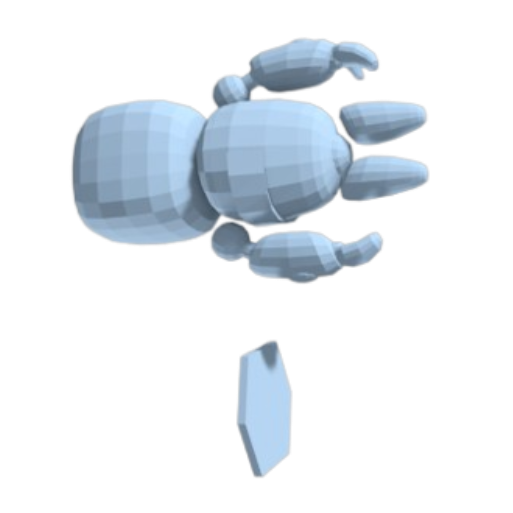}} &
\adjustbox{valign=c}{\includegraphics[width=.09\linewidth]{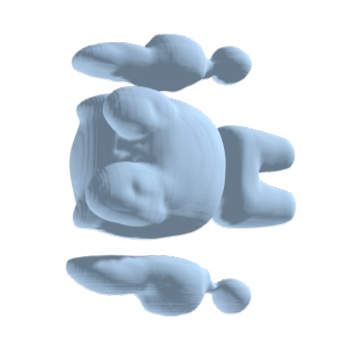}} &
\adjustbox{valign=c}{\includegraphics[width=.09\linewidth]{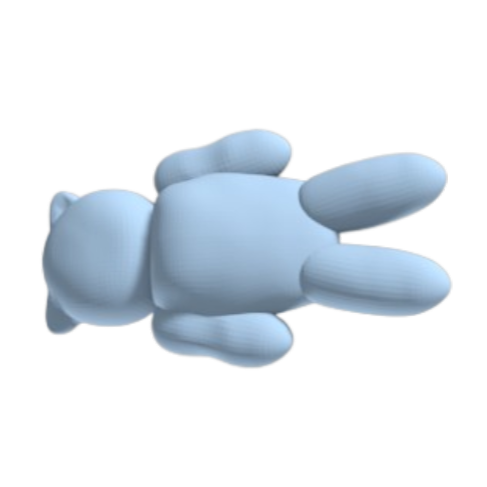}} &
\adjustbox{valign=c}{\includegraphics[width=.09\linewidth]{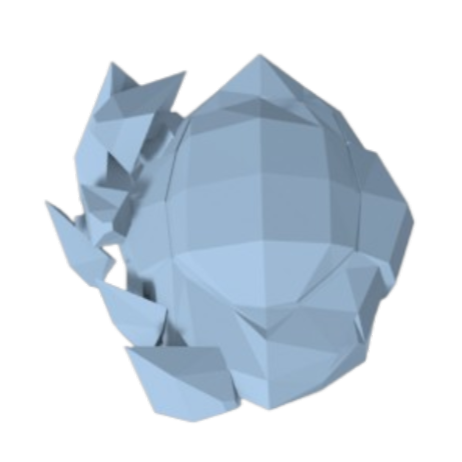}} &
\adjustbox{valign=c}{\includegraphics[width=.09\linewidth]{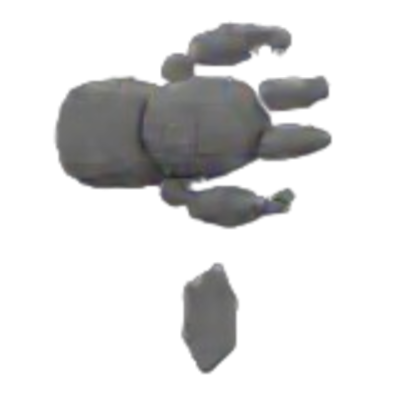}} &
\adjustbox{valign=c}{\includegraphics[width=.09\linewidth]{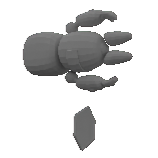}} \\
\addlinespace[0.3em]

\makecell{a Buddha statue sitting on a mat} &
\adjustbox{valign=c}{\includegraphics[width=.09\linewidth]{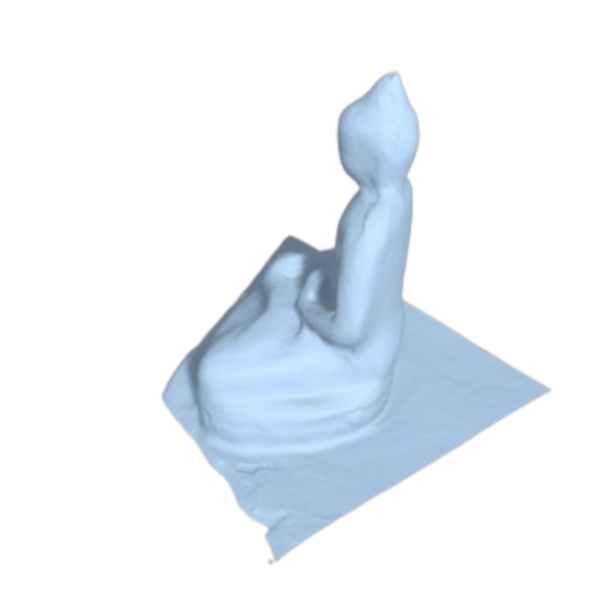}} &
\adjustbox{valign=c}{\includegraphics[width=.09\linewidth]{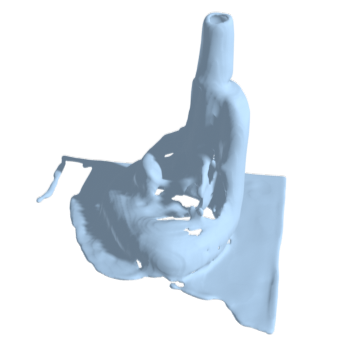}} &
\adjustbox{valign=c}{\includegraphics[width=.09\linewidth]{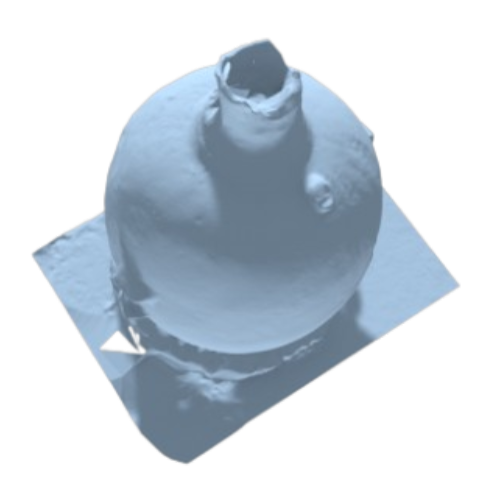}} &
\adjustbox{valign=c}{\includegraphics[width=.09\linewidth]{figs/Img_vs_3D/3D2Top1NN_LFD.png}} &
\adjustbox{valign=c}{\includegraphics[width=.09\linewidth]{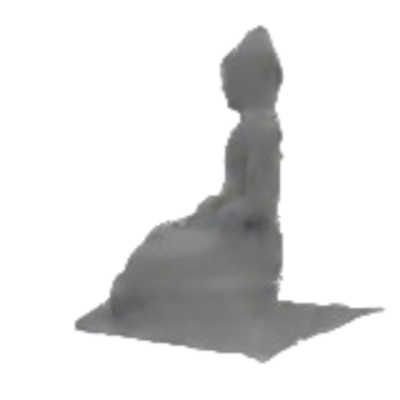}} &
\adjustbox{valign=c}{\includegraphics[width=.09\linewidth]{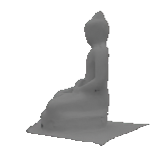}} \\

\addlinespace[0.1em]
\Xhline{0.3pt}
\addlinespace[0.1em]

\end{tabularx}
\caption{\textbf{Qualitative comparison between 3D and image models for memorization.} 
For each prompt, we show the training example, the generated 3D shape and its top-1 nearest training shapes under SSCD and LFD, together with the generated image and its top-1 nearest training image under SSCD.
Image generation tends to replicate training samples, whereas the 3D generator produces more novel shapes.}
\label{tab:3d_img_case}
\end{table*}

\paragraph{Dataset.} We conduct controlled experiments on a customized subset of Objaverse-XL~\citep{deitke2023objaversexl}. We obtain captions for shapes from Cap3D~\citep{luo2023scalable}.
To assign class labels to shapes, we first use Qwen3 Embedding~\citep{zhang2025qwen3emb} to select the most relevant class for each caption from the 100 most frequent categories in Objaverse-LVIS~\citep{deitke2023objaverse}. Then, we use Qwen3~\citep{yang2025qwen3} to remove mismatched caption-label pairs.
This process results in a dataset of about 140K 3D objects with captions and labels. In our experiments, we split the dataset into a 120K training set and a 20K test set.

\paragraph{Model.} While 3D representations are diverse, we use the vector-set (Vecset) representation in our controlled experiments, as it is widely used in 3D generation~\citep{zhao2023michelangelo,wu2024direct3d,li2025triposg,hunyuan3d2025hunyuan3d}. The autoencoder encodes the sampled point cloud into a compact, unordered latent Vecset and decodes it by predicting occupancy or signed distance function (SDF) values at randomly sampled spatial query points.
In our experiments, we use the pre-trained $1024 \times 32$ VecSetX autoencoder~\citep{3DShape2Vecset} and adopt Hunyuan3D 2.1~\citep{hunyuan3d2025hunyuan3d} as the diffusion backbone.

\paragraph{Memorization evaluation.} We follow standard practice in image diffusion models~\citep{somepalli2022diffusion} by evaluating memorization in samples generated from training prompts.
Throughout the experiments, we use our evaluation framework in Section~\ref{subsec:eval_frame} and fix $|P_\text{test}| = |Q| = 2500$. $P_\text{test}$ is downsampled from the full test set while preserving the class distribution.

Details of the data curation pipeline, autoencoder, and diffusion model backbone are provided in Appendix~\ref{app:dataset_model}.

\begin{figure}[h]
    \centering
    \includegraphics[width=\linewidth]{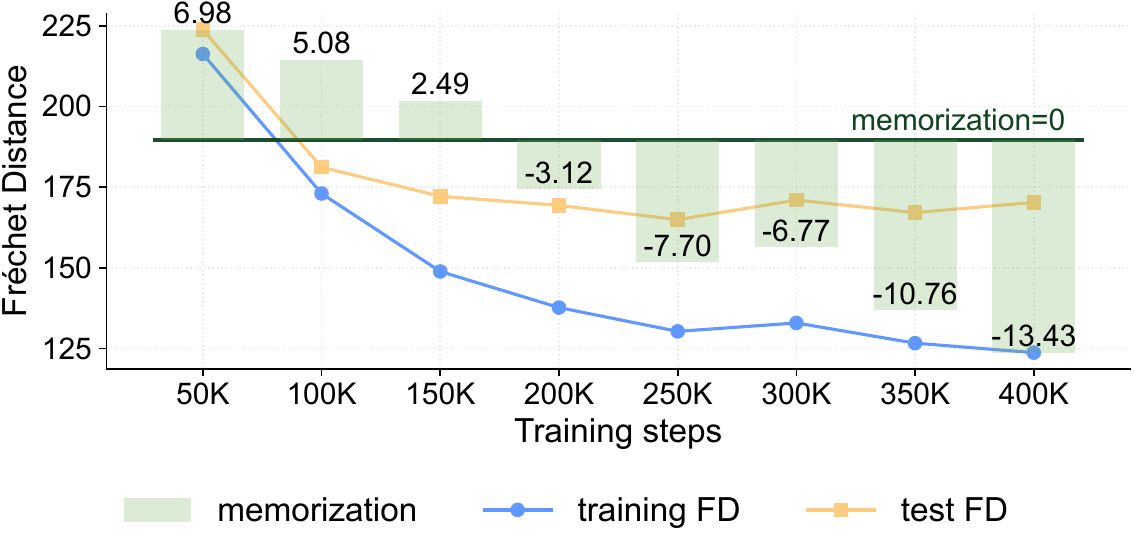}
    \caption{\textbf{Training dynamics of the baseline model.} As training progresses, $Z_U$ and training FD simultaneously decrease, whereas test FD decreases initially before plateauing at around 200K steps.}
    \label{fig:checkpoint-probing}
    \vspace{-1em}
\end{figure}
\subsection{Baseline Model}
\label{subsec:Baseline}

As the basis of our controlled experiments, we design a baseline with reasonable generation quality and low training cost. We focus on the 16 most frequent classes and limit the dataset to 50K objects to improve stability and avoid model collapse. We then train a 323M-parameter text-conditional 3D diffusion model on this subset for 200K steps.

Figure~\ref{fig:baseline_3x3} confirms that most shapes generated by the baseline model are visually plausible.
We show the trends of training FD, test FD, and $Z_U$ in Figure~\ref{fig:checkpoint-probing}. We observe that generation quality on test prompts (\ie, test FD) plateaus around 200K steps, and further training results in higher memorization, as indicated by $Z_U$. 
Meanwhile, the strong correlation between training FD and $Z_U$ suggests that training FD may indeed be influenced by memorization and, therefore, is not a reliable indicator of generation quality.

In Appendix~\ref{app:baseline}, we sanity-check our setup by confirming that increasing dataset size reduces memorization and that larger models memorize more. These observations are consistent with prior work~\citep{somepalli2022diffusion,carlini2023extracting_training,gu2023memorization,zeng2025generative}.

\begin{figure*}[t]
    \centering
    \vspace{-2.5em}
    \begin{subfigure}{0.48\textwidth}
        \centering
        \includegraphics[width=\linewidth]{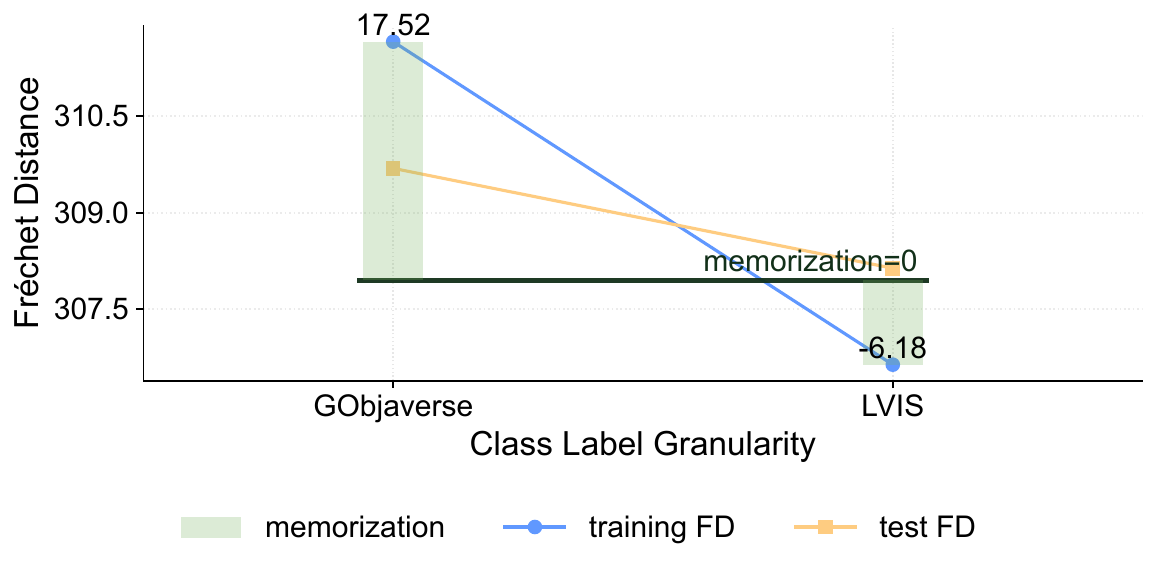}
        \caption{Class-conditional models.}
    \end{subfigure}
    \hfill
    \begin{subfigure}{0.48\textwidth}
        \centering
        \includegraphics[width=\linewidth]{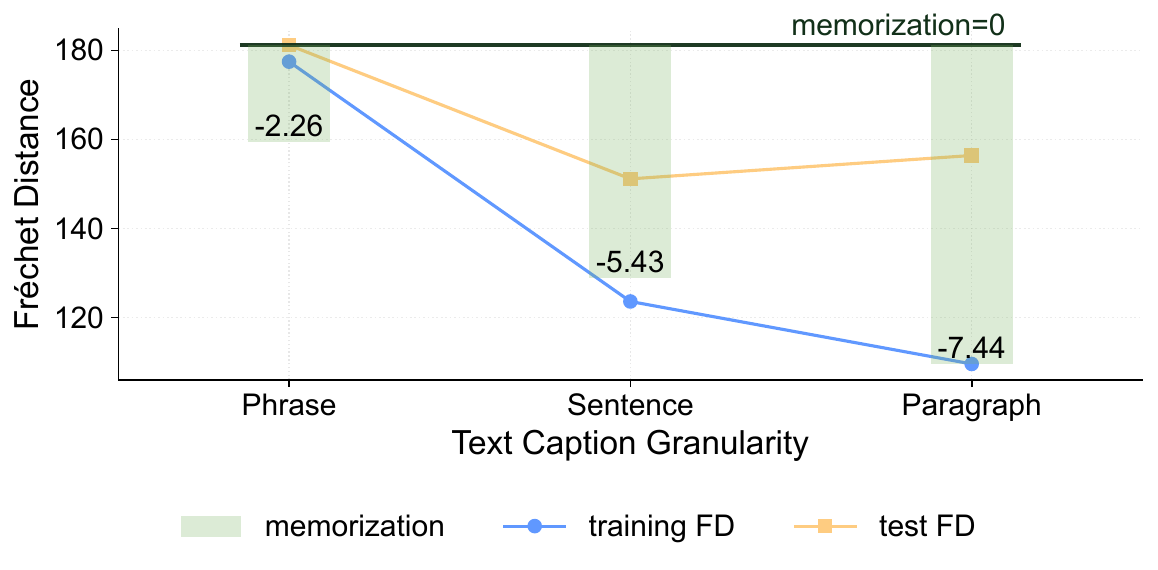}
        \caption{Text-conditional models.}
    \end{subfigure}
    \caption{
    \textbf{Fine-grained conditioning increases memorization.}
    For both class-conditional (left) and text-conditional (right) models, as the conditioning becomes finer-grained, the 3D generative models exhibit stronger memorization.
    }
    \label{fig:cond_granularity}
\end{figure*}

\subsection{Data Modality}
\begin{table}[htbp]
    \centering
    \normalsize
    \setlength{\tabcolsep}{6pt}
    \begin{tabular}{lccc}
        \toprule
        modality & $Z_U^\textit{DinoV2}$ & $Z_U^\textit{SSCD}$ & $Z_U^\text{LFD}$\\
        \midrule
        image & 8.58 & -8.71 & - \\
        3D & 15.70 & 2.49 & -5.60 \\
        \bottomrule
    \end{tabular}
    \caption{\textbf{Images are more prone to memorization than 3D shapes.} We compute $Z_U^\textit{DinoV2}$ and $Z_U^\textit{SSCD}$ for image generation models, and extend these metrics to 3D shapes as in Section~\ref{subsec:eval}. Images have lower $Z_U$, indicating stronger memorization.}
    \label{tab:Img_vs_3D}
\end{table}

Both image and 3D shape generation models operate in latent space rather than ambient space, so it is unclear whether data modality still affects memorization. We test this by comparing the two modalities in a controlled setting.

\paragraph{Setup.} We render a single fixed-view image for each object in the training dataset to obtain a 50K-image dataset at $256\times256$ resolution, in one-to-one correspondence with the 3D assets.
We train two models, one for images and one for shapes, using the same text prompts and the same Hunyuan3D backbone. The 3D model operates on the unordered Vecset representation, for which we omit positional embeddings, whereas the image model uses a 2D latent grid, to which we add 2D positional embeddings before flattening it into a sequence.
Empirically, adding positional embeddings to Vecset or removing them from the image latent grid significantly degrades generation quality, so this discrepancy is necessary. For images, we use Flux VAE~\citep{batifol2025flux}. It also provides high reconstruction quality and matches VecSetX's latent sequence length (1024), enabling a controlled comparison. Details are provided in Appendix~\ref{app:data_modality}.

\paragraph{Memorization evaluation.} To compare memorization across modalities, we follow prior work~\citep{somepalli2022diffusion} and use SSCD and DinoV2 as copy-detection models for images, and we extend them to 3D as described in Section~\ref{subsec:eval}. Table~\ref{tab:Img_vs_3D} shows that, under both $Z_U^\textit{DinoV2}$ and $Z_U^\textit{SSCD}$, the image generation model exhibits stronger memorization than the 3D generation model. As discussed in Section~\ref{subsec:eval}, SSCD and DinoV2 are not reliable retrieval metrics for 3D shapes. However, even if we compare $Z_U$ for shapes and images based on the most accurate distance metrics for each respective modality (\ie, LFD for shapes and SSCD for images), the $Z_U$ results still indicate that the image generation model memorizes more than the 3D generation model.

This aligns with the qualitative examples in Table~\ref{tab:3d_img_case}, where generated images are replications of training images, while generated shapes exhibit novel features (\eg, a toy with floating wings and a taller, sharper-headed Buddha).

\paragraph{Discussion.} Since each training image is a single view of the underlying 3D object, which collapses spatial structure and geometric detail into a near-grayscale projection, our image set naturally has less information. Thus, the image modality being more prone to memorization than the 3D modality could potentially be explained by the observation that diffusion models tend to memorize simpler data~\citep{somepalli2023understanding}.

\textit{Under matched configurations, diffusion models are more prone to memorizing images than 3D shapes.}

\begin{figure}[h]
    \centering
    \includegraphics[width=\linewidth]{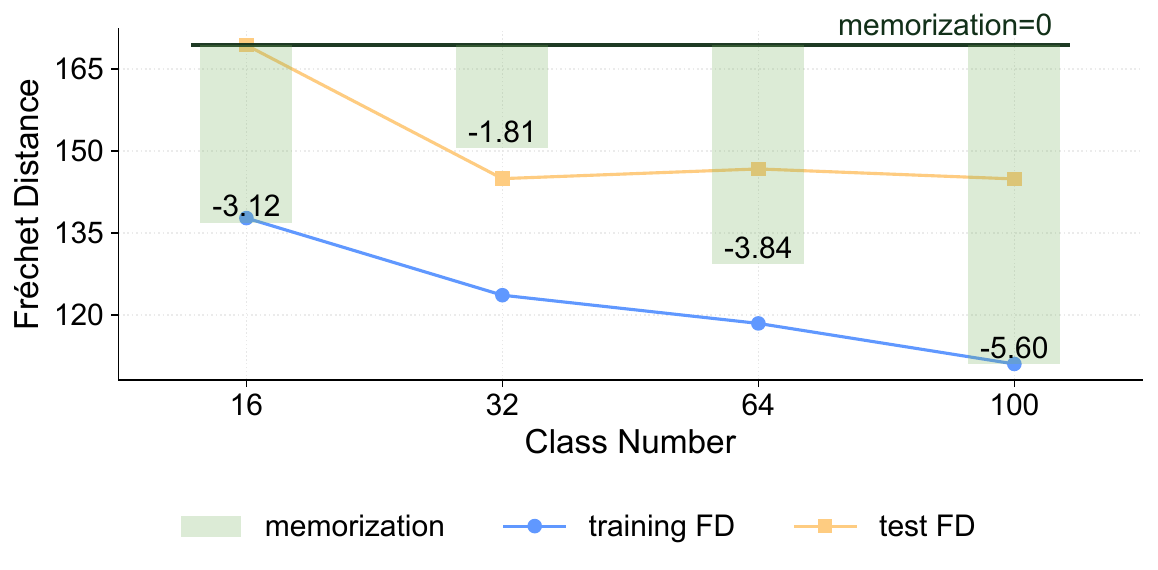}
    \caption{\textbf{Higher data diversity increases memorization.} With a fixed training set size, increasing the number of classes from 16 to 100 leads to a moderate increase in memorization.}
    \label{fig:Data-Diversity}
\end{figure}

\subsection{Data Diversity}

\paragraph{Setup.} We study how data diversity impacts memorization by creating sub-sampled datasets of equal size from the top 16, 32, 64, and 100 most frequent classes, while preserving the class distribution of the corresponding full datasets.

\paragraph{Memorization evaluation.} In Figure~\ref{fig:Data-Diversity}, we compare the four settings. Higher data diversity increases memorization (\ie, leads to lower $Z_U$) without reducing quality, as indicated by the nearly flat test FD for 32, 64, and 100 classes.

\paragraph{Discussion.} This observation stands in contrast to prior findings on image diffusion models~\citep{gu2023memorization}, which report that increasing dataset diversity improves generalization on CIFAR-10. However, CIFAR-10 is fundamentally limited in scale and semantic diversity, and the experiments in~\citep{gu2023memorization} only use up to 2000 images.
In comparison, our setting is more realistic: our data are sub-sampled from Objaverse-XL, one of the largest and most diverse 3D object datasets.

\textit{With a fixed data budget, higher semantic diversity leads 3D diffusion models to memorize more.}

\begin{figure*}[t]
    \centering
    \vspace{-2.5em}
    \begin{subfigure}{0.48\textwidth}
        \centering
        \includegraphics[width=\linewidth]{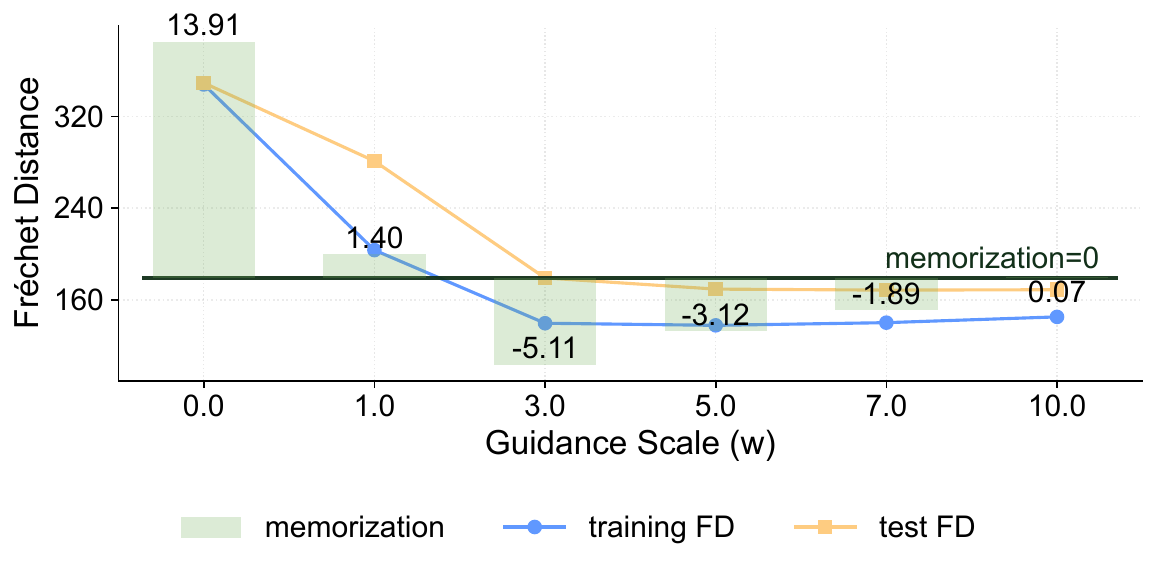}
        \caption{Base model}
    \end{subfigure}
    \hfill
    \begin{subfigure}{0.48\textwidth}
        \centering
        \includegraphics[width=\linewidth]{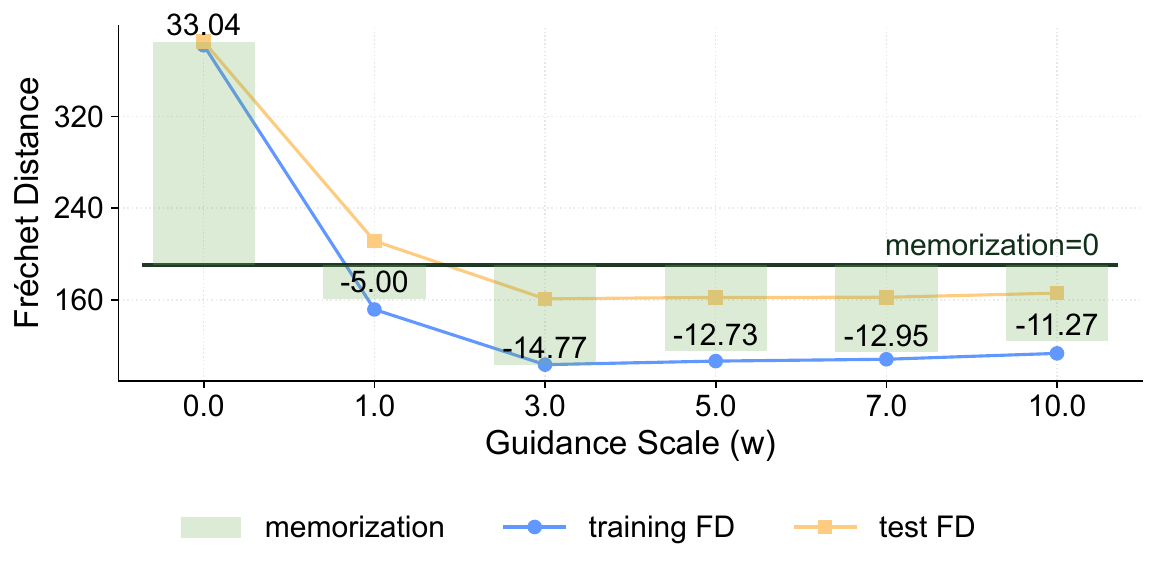}
        \caption{Large model}
    \end{subfigure}
    \caption{
    \textbf{Moderate guidance scales lead to the strongest memorization.}
    For both the base model (left) and the large model (right), unconditional generation ($w=0$) or generation with no guidance ($w=1$) yields low memorization but poor generation quality. A moderate guidance scale ($w=3$) leads to the strongest memorization; further increasing the guidance scale ($w>3$) reduces memorization.
    }
    \label{fig:gs}
\end{figure*}

\subsection{Conditioning Granularity}
\label{subsec:cond_gran}

We conduct experiments with both class-conditional and text-conditional models to study how conditioning granularity affects memorization in diffusion models. 

\paragraph{Setup.} For class conditioning, in addition to the original 16 fine-grained LVIS labels, we use Qwen3-VL~\citep{Qwen3-VL} to assign each training shape to one of six coarse categories from GObjaverse~\citep{zuo2024sparse3d} (animals, daily-use, electronics, furniture, human-shape, and transportation). For text conditioning, we use Qwen3-VL to generate training prompts at three increasing levels of granularity: phrase, sentence, and paragraph. Captioning prompts, implementation details, and text-length distributions are provided in Appendix~\ref{app:cond_granularity}.

\paragraph{Memorization evaluation.} Figure~\ref{fig:cond_granularity} shows that memorization consistently increases as we move from coarse-grained conditions (\ie, broad classes or phrases) to fine-grained conditions (\ie, specific classes or paragraphs).

\paragraph{Discussion.} As the conditions become finer-grained, it is natural that fewer training samples correspond to each condition. This reduced diversity makes it easier for the generative model to associate the specific conditions with their exact corresponding training samples~\citep{somepalli2023understanding,wen2024detecting}.
Fine-grained conditions have higher controllability, while coarser ones reduce memorization. Mixing prompts with different granularities may be a good way to balance the two.

\textit{Higher granularity in conditions increases memorization in 3D shape generation models.}

\section{Impact of Modeling on Memorization}
\label{sec:mitigation}

In the previous section, we explore how characteristics of training data can impact memorization. However, modifying training data may be unrealistic in practical settings. Thus, to identify effective ways to mitigate memorization, we further investigate the impact of modeling designs.

\subsection{Guidance Scale}
\label{subsec:GS}

The guidance scale $w$ is a core factor in diffusion models that apply classifier-free guidance (CFG). It represents how strongly the condition is applied, following the formula:
\begin{equation}
\tilde{\epsilon}(z, c)
= w \cdot \epsilon(z, c) + (1-w) \cdot \epsilon(z)
\end{equation}
where $z$ is the latent Vecset, $c$ is the conditioning embedding, and
$\epsilon$ is the noise prediction network. Here, $w$ is the guidance scale; $w = 0$ yields unconditional generation, and $w = 1$ yields conditional generation with no guidance. 

\paragraph{Setup.} We study how guidance scale impacts memorization by generating samples from the text-conditional baseline model with different guidance scales ($w \in \{0,1,3,5,7,10\}$). We then repeat this study on a larger (1.5B-parameter) text-conditional model (details in Appendix~\ref{app:baseline}) and a class-conditional model (details in Appendix~\ref{app:gs}) to confirm that the same trend holds.
\begin{table}[htbp]
\centering
\tablestyle{5pt}{1.2}
\begin{tabular}{lcccc}
$w$ & gen & top match & caption (top match) & LFD \\
\shline
0 & \adjustbox{valign=c}{\includegraphics[width=.12\linewidth]{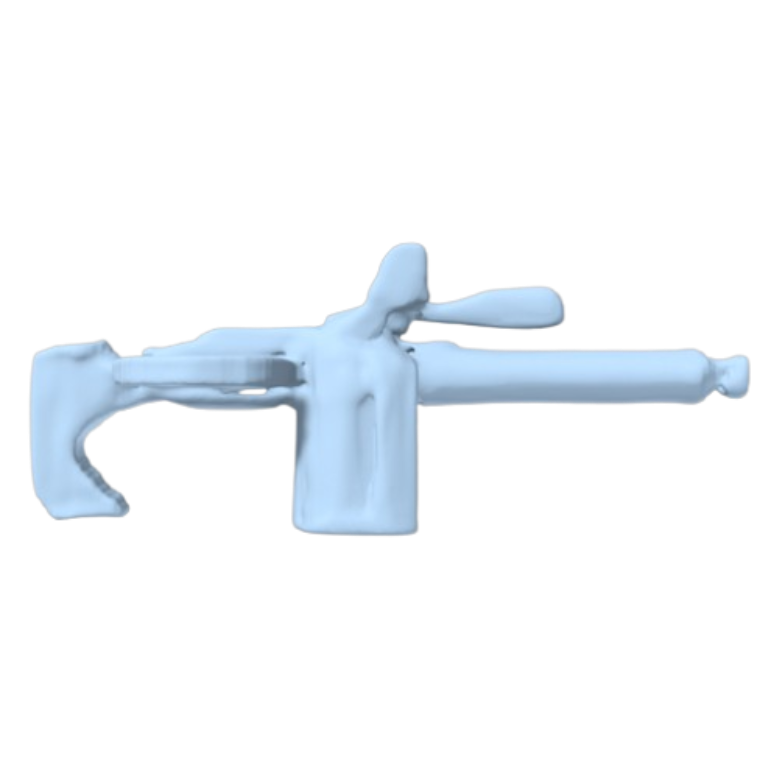}} & \adjustbox{valign=c}{\includegraphics[width=.12\linewidth]{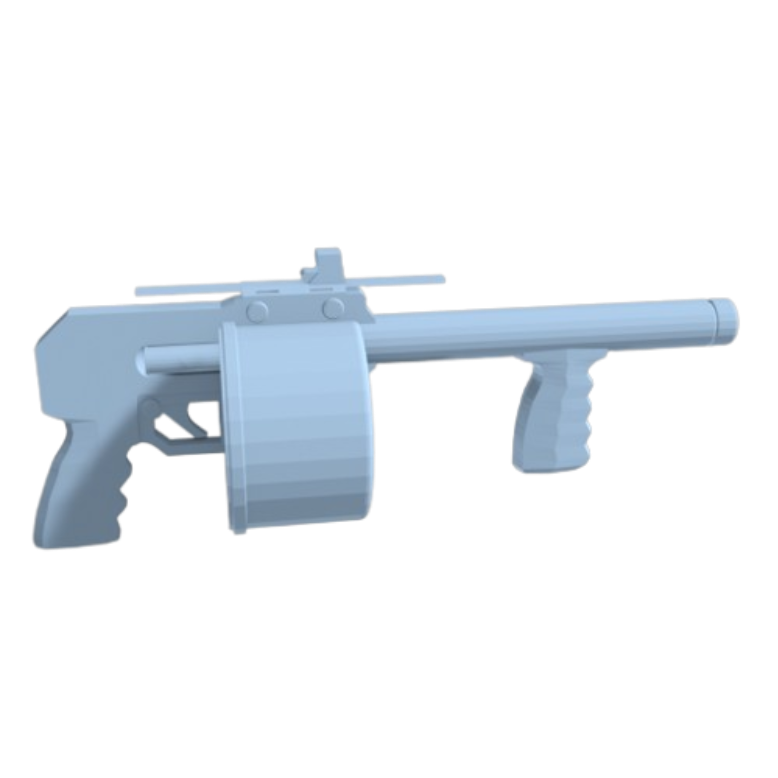}} & \adjustbox{valign=c}{\makecell{``stylized tommy gun\\with drum magazine and\\cartoonish eyes''}} & 5748 \\
1 & \adjustbox{valign=c}{\includegraphics[width=.12\linewidth]{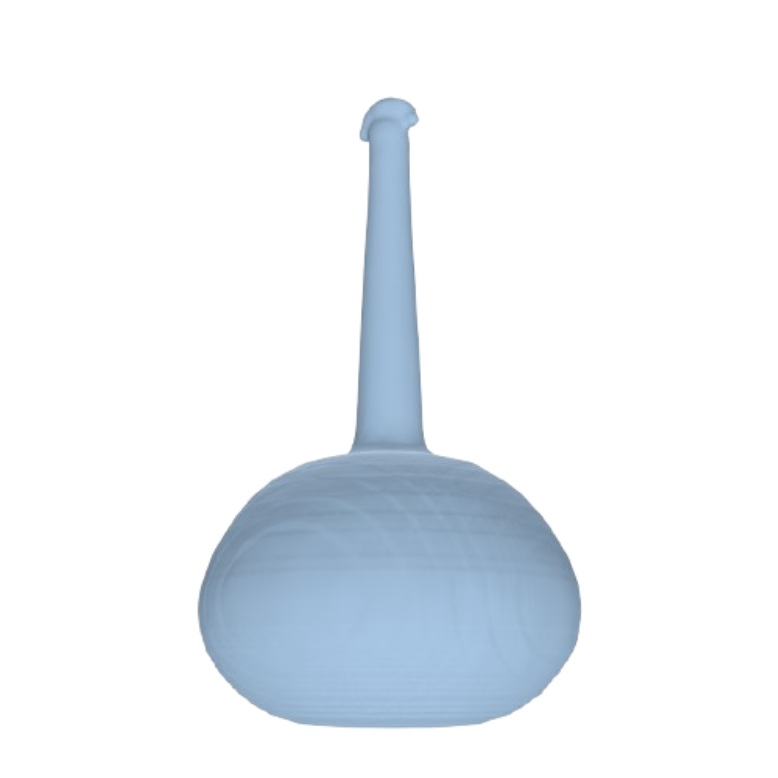}} & \adjustbox{valign=c}{\includegraphics[width=.12\linewidth]{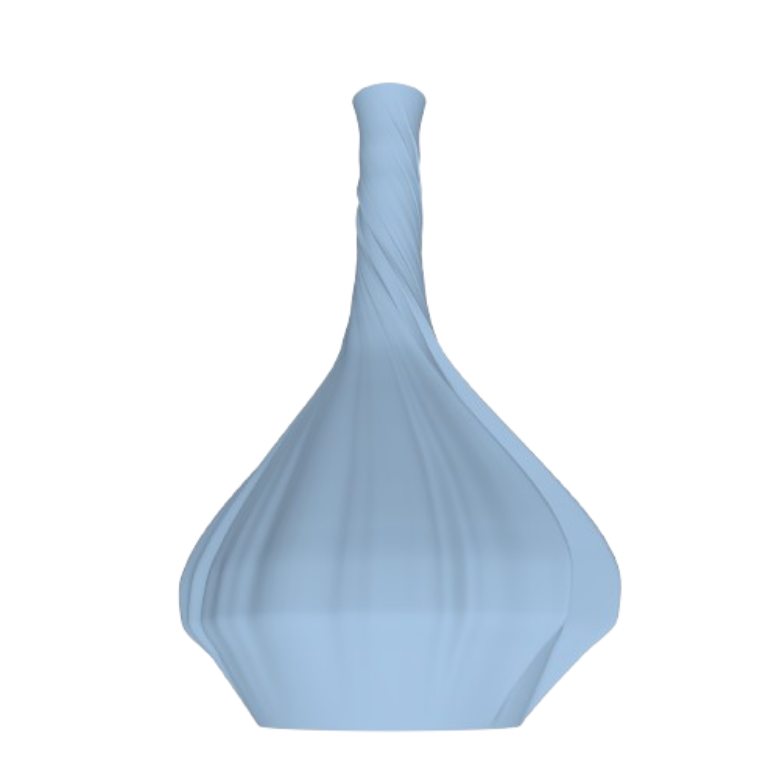}} & \adjustbox{valign=c}{\makecell{``vase''}} & 3482 \\
3 & \adjustbox{valign=c}{\includegraphics[width=.12\linewidth]{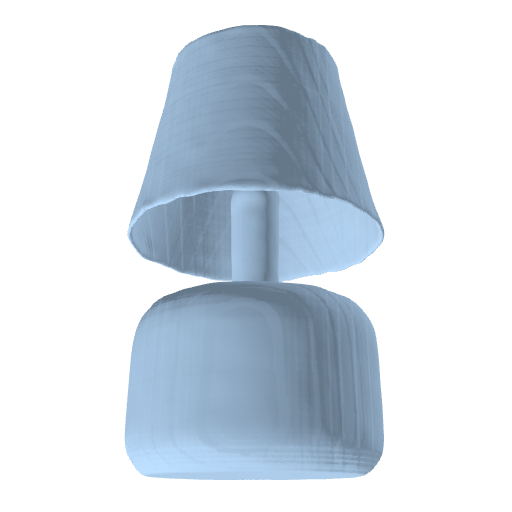}} & \adjustbox{valign=c}{\includegraphics[width=.12\linewidth]{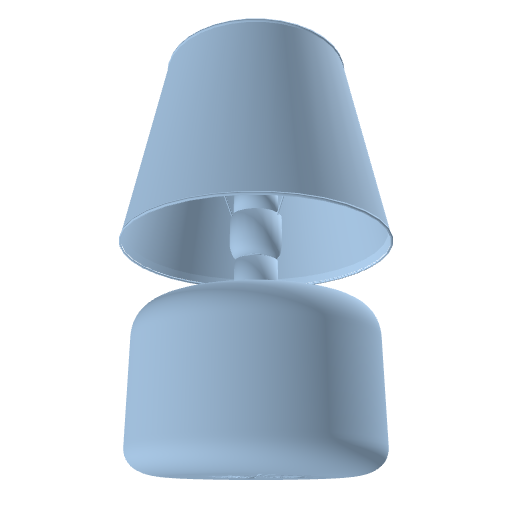}} & \adjustbox{valign=c}{\makecell{``table lamp with\\conical lampshade and\\cylindrical base''}} & 1076 \\
5 & \adjustbox{valign=c}{\includegraphics[width=.12\linewidth]{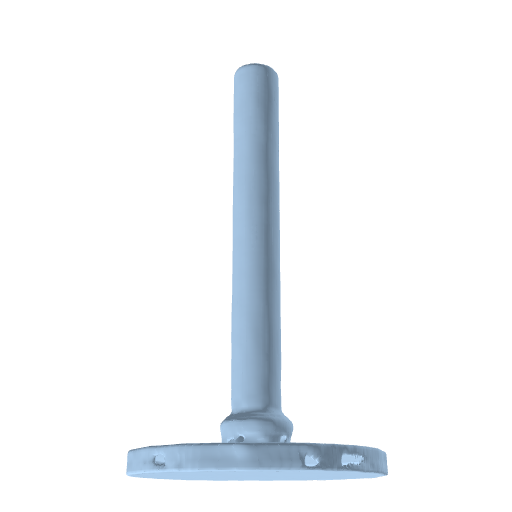}} & \adjustbox{valign=c}{\includegraphics[width=.12\linewidth]{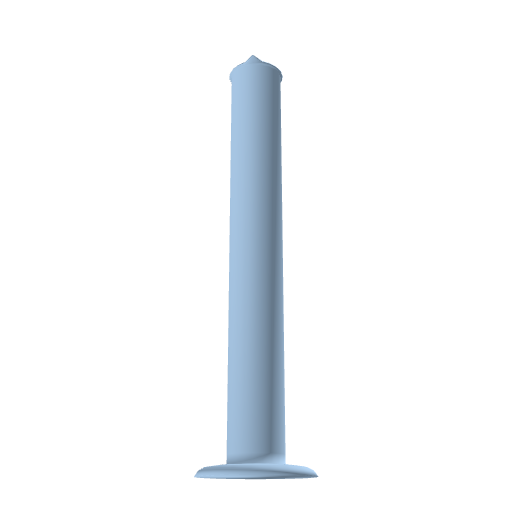}} & \adjustbox{valign=c}{\makecell{``slender cylinder with\\circular base''}} & 4202 \\
7 & \adjustbox{valign=c}{\includegraphics[width=.12\linewidth]{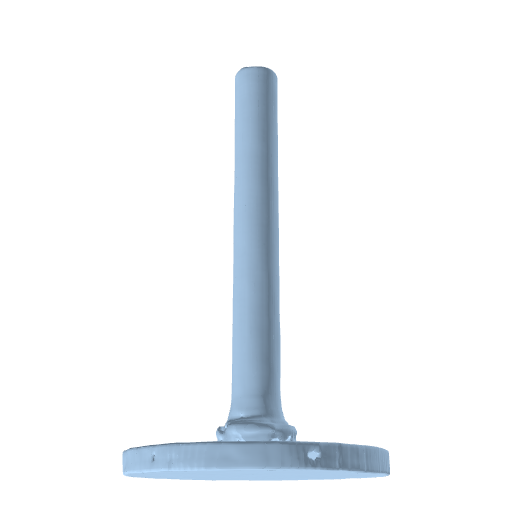}} & \adjustbox{valign=c}{\includegraphics[width=.12\linewidth]{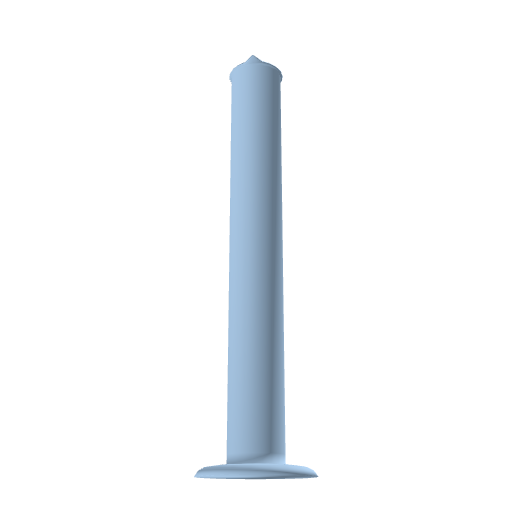}} & \adjustbox{valign=c}{\makecell{``slender cylinder with\\circular base''}} & 4134 \\
10 & \adjustbox{valign=c}{\includegraphics[width=.12\linewidth]{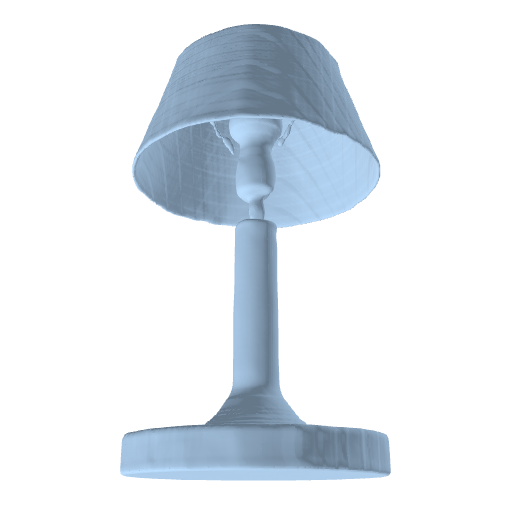}} & \adjustbox{valign=c}{\includegraphics[width=.12\linewidth]{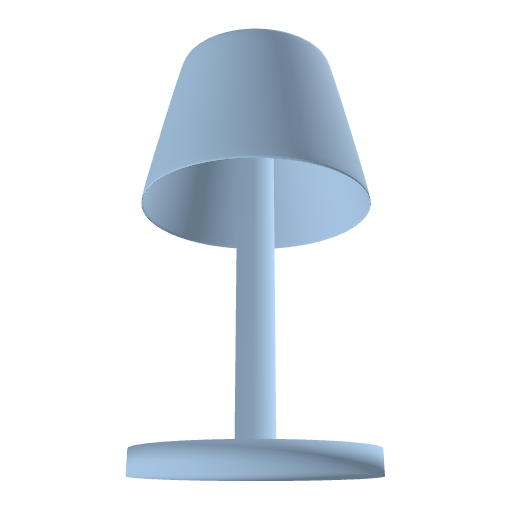}} & \adjustbox{valign=c}{\makecell{``table lamp with\\a shade''}} & 2610 \\
\end{tabular}
\caption{\textbf{Guidance scale case study for the prompt ``table lamp with conical lampshade and cylindrical base''.} Memorization is strongest at guidance scale $w=3$, where the model most closely reproduces the training lamp. In contrast, samples at $w=0$ or $w=1$ either ignore the prompt or only weakly follow it, and samples at larger $w$ emphasize sub-phrases such as the base or the shade, rather than reproducing the full shape.}
\vspace{-1em}
\label{tab:gs-check}
\end{table}

\paragraph{Memorization evaluation.} Figure~\ref{fig:gs} shows a non-monotonic trend between guidance scale and memorization: when generating unconditionally or with no guidance, the memorization level is low; increasing the guidance scale to 3 significantly increases memorization, yet further increasing the guidance scale turns out to reduce memorization.

\paragraph{Discussion.} In line with previous work~\citep{gu2023memorization,jain2025classifier}, we find that generating unconditionally or with no guidance leads to reduced memorization. However, this is accompanied by poor generation quality (as shown by the high training and test FD). Moreover, studies on guided diffusion models~\citep{dhariwal2021diffusion,ho2022classifier} show that larger guidance scales place more weight on the conditioning signal and trade diversity for higher fidelity, which has led to the common belief that, for training prompts, stronger guidance encourages memorization. Nevertheless, when the guidance scale exceeds 3, our results no longer align with this common belief.

To better understand this, we identify prompts for which $w=3$ yields low retrieval distance, while higher values of $w$ do not, and show an example in Table~\ref{tab:gs-check} (two more cases in Appendix~\ref{app:gs}).
The model generates irrelevant shapes (\eg, ``machine gun'') when $w=0$, generates only a similar cylindrical base when $w=1$, faithfully generates a shape highly similar to the training data when $w=3$, and produces shapes partially aligned with the prompts for even higher $w$ (\eg, ``cylindrical base'', ``table lamp with shade'').

\textit{Moderate guidance scales lead to the strongest memorization, whereas larger guidance scales may drive models to focus on parts of the prompt and thereby reduce copying.}

\subsection{Latent Space}
\label{subsec:latent}
\begin{figure}[htbp]
    \centering
    \includegraphics[width=\linewidth]{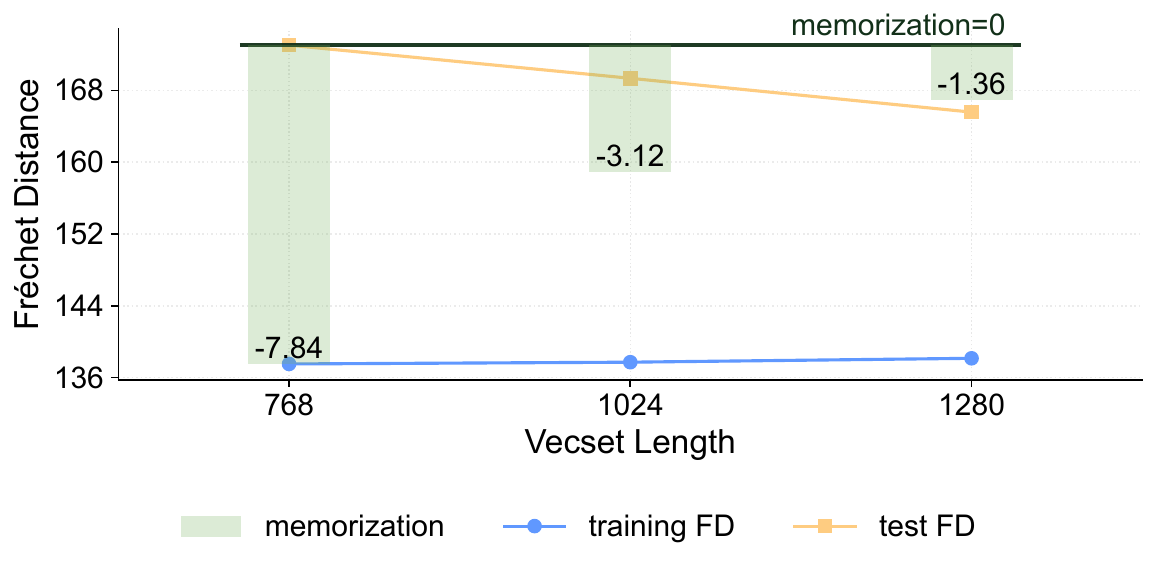}
    \caption{\textbf{Increasing Vecset sequence length helps generalization.} Generation quality improves, and memorization decreases.}
    \label{fig:Latent_Length}
\end{figure}

\paragraph{Setup.} We study the effect of latent Vecset sequence length on memorization by training diffusion models with sequence lengths of 768, 1024 (baseline), and 1280. This does not require retraining the autoencoder~\citep{zhang2024clay}. In Appendix~\ref{app:latent}, we verify that changing sequence length does not degrade the autoencoder's reconstruction performance.

\begin{figure}[h]
\centering
\setlength{\tabcolsep}{1.2pt}
\renewcommand{\arraystretch}{-0.5}
  \begin{tabular}{cccc}
    train & 768 & 1024 & 1280 \\
    \includegraphics[width=.25\linewidth]{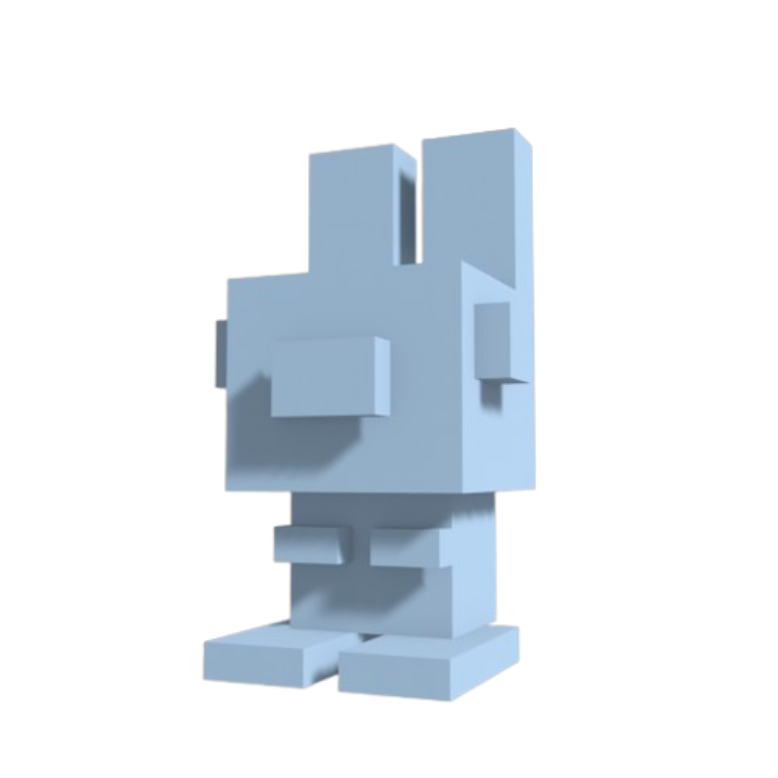} & 
    \includegraphics[width=.25\linewidth]{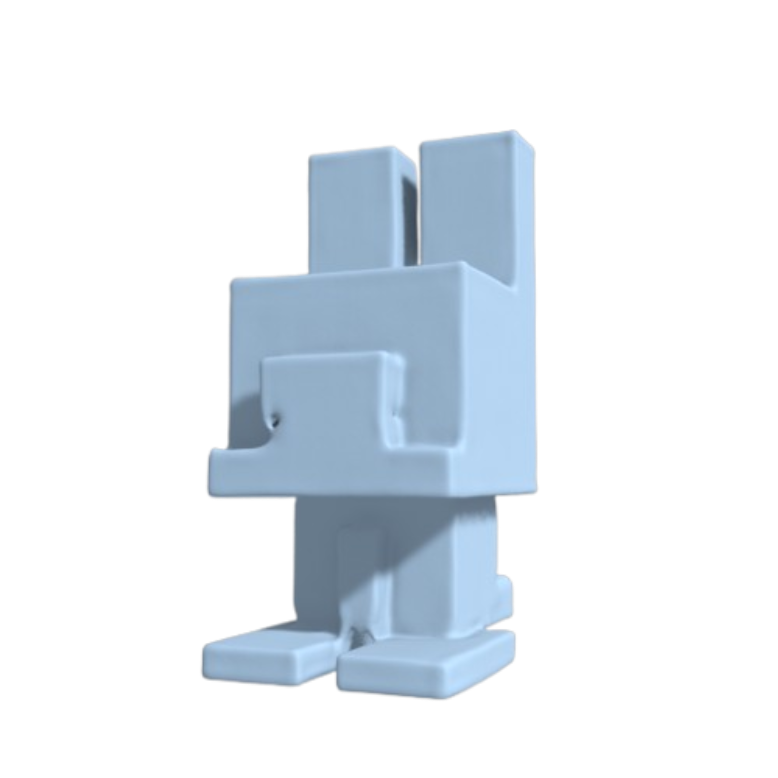} & 
    \includegraphics[width=.25\linewidth]{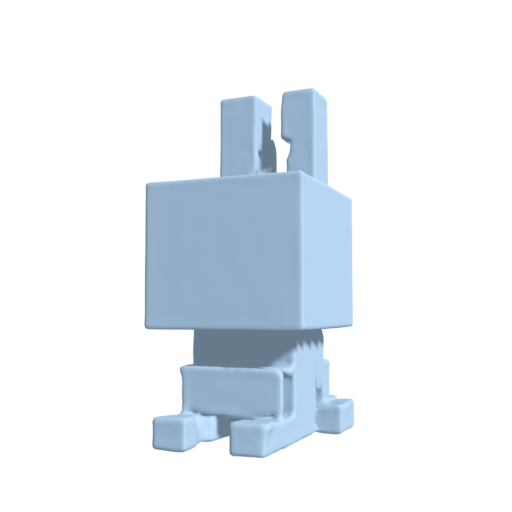} & 
    \includegraphics[width=.25\linewidth]{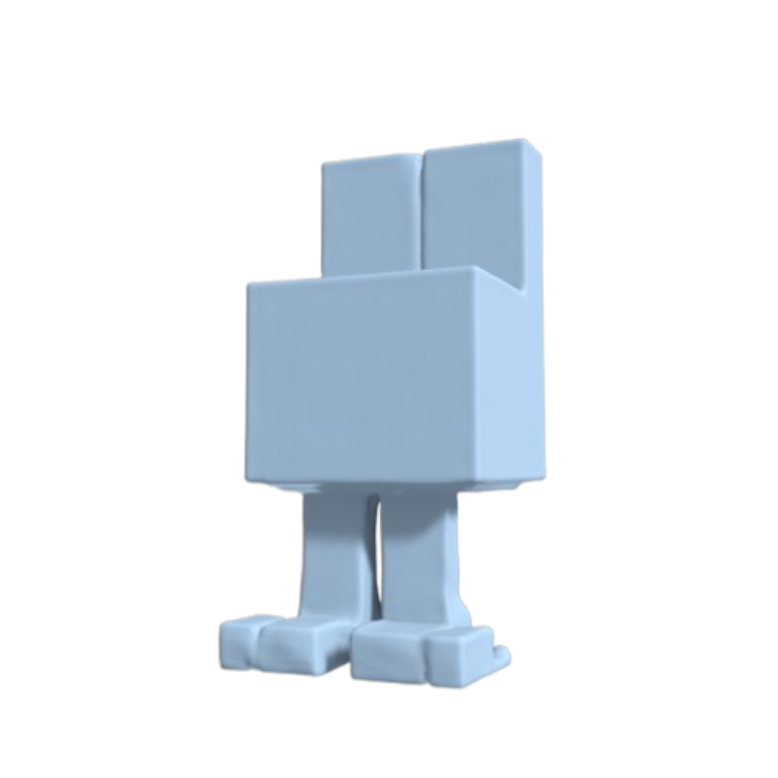} \\
  \end{tabular}
  \vspace{-0.25em}
  \caption{\textbf{Shapes generated from models trained with different Vecset lengths for the prompt ``blocky stylized rabbit figure''.} Models trained with longer Vecset lengths (1024 and 1280) generate high-quality shapes that are well aligned with the prompt, while exhibiting novel features not present in the training shape.
  }
  \label{fig:vecset-length}
\end{figure}

\paragraph{Memorization evaluation.} As shown in Figure~\ref{fig:Latent_Length}, increasing the latent sequence length improves generation quality and reduces memorization. To further inspect this, we identify prompts for which the 768-Vecset model achieves low retrieval distance, while the longer-Vecset models do not. As illustrated in Figure~\ref{fig:vecset-length} (additional examples in Appendix~\ref{app:latent}), all models produce shapes that are visually aligned with the prompt. The model with the shortest Vecset length generates shapes that closely resemble the training shape. As the latent length increases, the model still generates high-quality shapes aligned with the text prompt, but it no longer fully reproduces a specific training example.

\paragraph{Discussion.} These observations indicate that increasing the Vecset length helps models generalize better. 
Using the same training recipe and autoencoder, models with longer Vecsets generate shapes that are better aligned with the prompts and less overfitted to specific training examples.

\textit{Longer Vecsets mitigate memorization while maintaining high shape fidelity and prompt alignment.}

\subsection{Rotation Augmentation}

\begin{figure}[htbp]
    \centering
    \includegraphics[width=\linewidth]{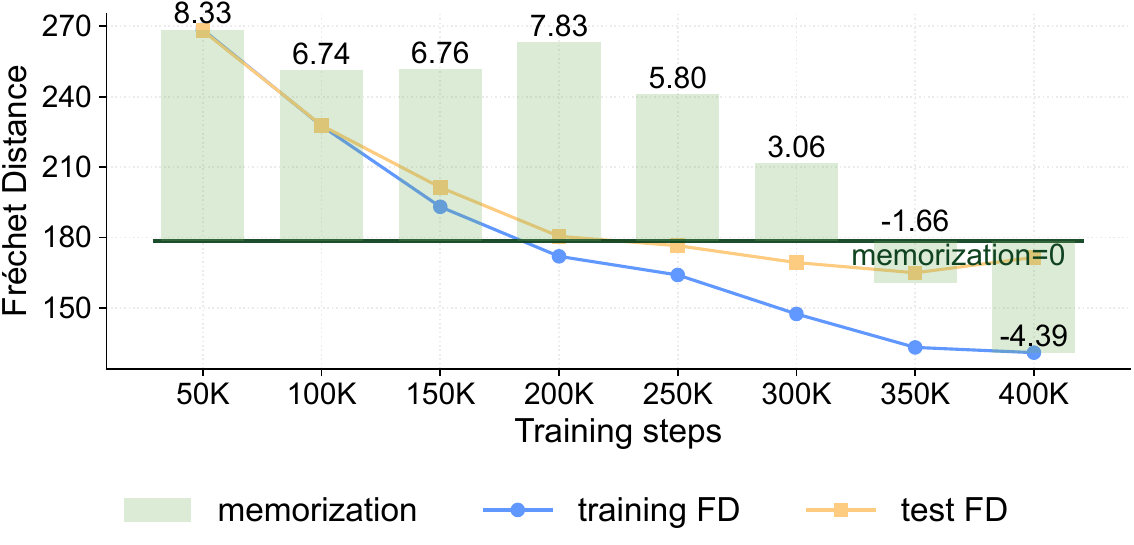}
    \caption{\textbf{Training dynamics with yaw rotation augmentation.}
    Compared to the baseline, test FD converges more slowly, yet after convergence, memorization (as reflected by $Z_U$) is much lower.}
    \label{fig:aug_curve}
\end{figure}

Data augmentation is widely used to improve generalization in deep learning~\citep{zhang2016understanding,hernandez2018data}. In 3D, however, augmentation is less straightforward. Common choices~\citep{ioannidou2017deep, zhu2024advancements}, such as jittering or downsampling point clouds, can distort geometry and harm quality. Among shape-preserving augmentations, rotation is a natural choice.

\paragraph{Setup.}
We study how rotation affects memorization in 3D generative models. For simplicity, we only consider yaw rotations at four discrete angles: $0^\circ$, $90^\circ$, $180^\circ$, and $270^\circ$ (details are in Appendix~\ref{app:rot_aug}). Each training shape is encoded only once, after applying a single random yaw rotation chosen from these angles. For each generated shape, we calculate the LFD for all four poses and take the minimum to prevent rotation from affecting retrieval performance.

\paragraph{Memorization evaluation.} Figure~\ref{fig:aug_curve} shows the training dynamics under rotation augmentation. Compared to the baseline curves in Figure~\ref{fig:checkpoint-probing}, the generation quality (\ie, test FD) requires substantially more training steps to converge with augmentation (350K \vs 200K). Nonetheless, once both runs converge and reach similar test FD, the model trained with rotation augmentation memorizes less, as indicated by a $Z_U$ closer to 0 (\eg, -1.66 at 350K \vs\ -3.12 at 200K).

\paragraph{Discussion.}
Rotation augmentation slows the convergence of generation quality, which can be a drawback under tight training budgets. However, at a similar generation quality (\ie, similar test FD), the rotation-augmented model memorizes less than the baseline. This suggests that rotation improves generalization without sacrificing performance. In 3D, where data is limited and expensive, rotation offers a simple and practical way to mitigate memorization.

\textit{Rotation augmentation slows convergence of generation quality but reduces memorization at similar quality levels.}

\section{Limitations}
Since our controlled experiments rely on the retrieval metric, LFD, our findings may be affected by its inaccuracy.
In addition, using training prompts for text-conditional generation may bias $Z_U$ downward even for models that do generalize. Using test prompts could potentially yield more accurate $Z_U$ estimates for high-performance 3D generators.

Our study is based on Vecset diffusion models, which we treat as representative of current 3D generative models.
Although we expect most of our findings to generalize across architectures, some experiments, such as varying Vecset length, remain specific to this model family. Further, due to computational constraints, we conduct our controlled experiments on a relatively small-scale model. Therefore, while the generation quality of our model is acceptable, it is not fully comparable with cutting-edge models.

\section{Conclusion}
In this work, we take an initial step toward understanding memorization in 3D generative models. We design a memorization evaluation framework for 3D generative models, and quantitatively evaluate the memorization level of existing 3D generators.
Furthermore, through carefully controlled experiments, we find that generative models tend to memorize image data more than 3D data. Increasing data diversity or conditioning granularity, or using a moderate guidance scale, results in more memorization. We propose several empirical recommendations to reduce memorization without degrading performance, such as using a longer latent Vecset and applying simple rotations during training. 
We hope our study can provide insights for designing future 3D generative models with improved generalization.

\paragraph{Acknowledgments.} We thank Koven Yu and Yinbo Chen for helpful discussions and feedback. We are pleased to acknowledge that our experiments are primarily conducted using the Princeton Research Computing resources at Princeton University, which are managed by a consortium of groups led by the Princeton Institute for Computational Science and Engineering (PICSciE) and Research Computing. We also gratefully acknowledge the support of Lambda, Inc. for providing additional computing resources.

{
    \small
    \bibliographystyle{ieeenat_fullname}
    \bibliography{main}
}

\appendix
\clearpage
\maketitlesupplementary
\setcounter{equation}{0}

\section
{Metrics Definition and Implementation}
\label{app:Metric_Def}

In this section, we detail the definitions and implementation of the metrics and the evaluation framework from Section~\ref{sec:retrieval_methods}.

\subsection{Distance Metrics}
\label{app:Dist_metric}
\paragraph{Chamfer distance (CD)} measures the average squared nearest-neighbor distance between two point clouds. We randomly sample 4096 points to compute CD. Given two point clouds $S_1, S_2 \subset \mathbb{R}^3$, their CD is defined as:
\begin{equation}
\begin{split}
d_{\text{CD}}(S_1, S_2) &= \frac{1}{|S_1|} \sum_{x \in S_1} \min_{y \in S_2} \|x - y\|_2^2 \\
&\quad + \frac{1}{|S_2|} \sum_{y \in S_2} \min_{x \in S_1} \|y - x\|_2^2
\end{split}
\end{equation}

\paragraph{Point cloud encoders} encode a point cloud $S$ into an embedding
$f(S) \in \mathbb{R}^D$. We consider the following encoders
and embedding dimensions:
\begin{align}
    D_{\text{PointNet++}} &= 512,\nonumber\\
    D_{\text{Uni3D}}      &= 1024,\nonumber\\
    D_{\text{ULIP-2}}     &= 1280\nonumber
\end{align}
Given two point clouds $S_1$ and $S_2$, we define their L2-normalized embedding distance as:
\begin{equation}
    d_{\text{emb}}(S_1, S_2)
    = 1 - \big\langle f(S_1), f(S_2) \big\rangle.
\end{equation}

\paragraph{Light Field Distance (LFD)} is computed by first rendering $256\times256$-resolution silhouettes of an object from 10 canonical viewpoints. Then, a feature vector is extracted from each silhouette by concatenating its Zernike moments and Fourier descriptors. The LFD between two shapes is computed by summing the $L_1$ distances between the corresponding view descriptors across the 10 viewpoints. We use the original implementation~\citep{chen2003LFD} to compute LFD.

\paragraph{Image encoders.}
For each mesh $M$, we render 12 RGB images $\{I_i(M)\}_{i=1}^{12}$ at resolution $256 \times 256$ using the same pipeline in Appendix~\ref{app:rendering}. We encode each view with an image
encoder and use the [CLS] token as the per-view
representation. The final embedding $f(M) \in \mathbb{R}^D$ for the mesh $M$ is obtained by averaging the
[CLS] tokens across the 12 views and L2-normalizing the
result. Given two meshes $M_1$ and $M_2$, the embedding distance induced by an image encoder is defined analogously to the point cloud encoders as:
\begin{equation}
    d_{\text{emb}}(M_1, M_2)
    = 1 - \big\langle f(M_1), f(M_2) \big\rangle.
\end{equation}

\subsection{Memorization Metrics}
\label{app:Mem_metric}
\paragraph{Mann-Whitney U Test\@} is a well-established non-parametric test. 
Following \citep{meehan2020non}, we use it to detect data-copying behavior, and we only consider its global version.

Let $P$ denote the underlying data distribution. The training set $T$ and the 
held-out test set $P_\text{test}$ are both drawn
i.i.d.\ from $P$. Let $Q$ denote the set generated by a generative model.
Let $d(x, T)$ be the distance from a point $x$ to the training set $T$
(\eg, the distance to its nearest neighbor in $T$ under a chosen metric).
For each $x \in P_\text{test}$, define $A = d(x, T)$, 
and for each $y \in Q$, define $B = d(y, T)$.
When the model generalizes, the distance distributions induced by $P_\text{test}$ and $Q$ should match.
The null hypothesis, therefore, states that a generated distance
$B$ is larger than a test distance $A$ with probability $1/2$:
\begin{equation}
    H_0: \Delta_T(P_\text{test}, Q) = \frac{1}{2},
\end{equation}
where
\begin{equation}
    \Delta_T(P_\text{test}, Q)
    = \Pr(B > A).
\end{equation}

In practice, let $n = |P_\text{test}|$ and $m = |Q|$, and collect the sets
of distances:
\[
\{A_i\}_{i=1}^n = \{d(x_i, T)\}_{i=1}^n, 
\qquad
\{B_j\}_{j=1}^m = \{d(y_j, T)\}_{j=1}^m.
\]
The Mann-Whitney U statistic for $Q$ is:
\begin{equation}
    U_Q = \sum_{j=1}^m \sum_{i=1}^n \mathbb{I}\bigl[B_j > A_i\bigr],
\end{equation}
which is an unbiased estimator of $mn \, \Delta_T(P_\text{test}, Q)$.
Equivalently, if we rank all $n + m$ values $\{A_i\}_{i=1}^n \cup \{B_j\}_{j=1}^m$
from smallest to largest, and let $R(B_j)$ denote the rank of $B_j$, then:
\begin{equation}
    R_Q = \sum_{j=1}^m R(B_j),
    \qquad
    U_Q = R_Q - \frac{m (m+1)}{2}.
\end{equation}

\paragraph{Mann-Whitney z-score.}
Under the null hypothesis $H_0$ and for $m, n \gtrsim 20$, the statistic
$U_Q$ is approximately normally distributed with mean $\mu_U$ and standard deviation $\sigma_U$:
\begin{equation}
    \mu_U = \frac{mn}{2},
    \qquad
    \sigma_U = \sqrt{\frac{mn(m + n + 1)}{12}}.
\end{equation}
We therefore define the normalized Mann-Whitney z-score as:
\begin{equation}
    Z_U(P_\text{test}, Q; T)
    = \frac{U_Q - \mu_U}{\sigma_U}.
\end{equation}
Intuitively, $Z_U \ll 0$ indicates data-copying (generated samples are
systematically \emph{closer} to the training set than test samples). $Z_U \approx 0$ indicates generalization.

\subsection{Evaluation Framework}
\label{app:eval_frame}
\begin{figure}[htbp]
    \centering
    \begin{subfigure}[b]{0.49\linewidth}
        \centering
        \includegraphics[width=\linewidth]{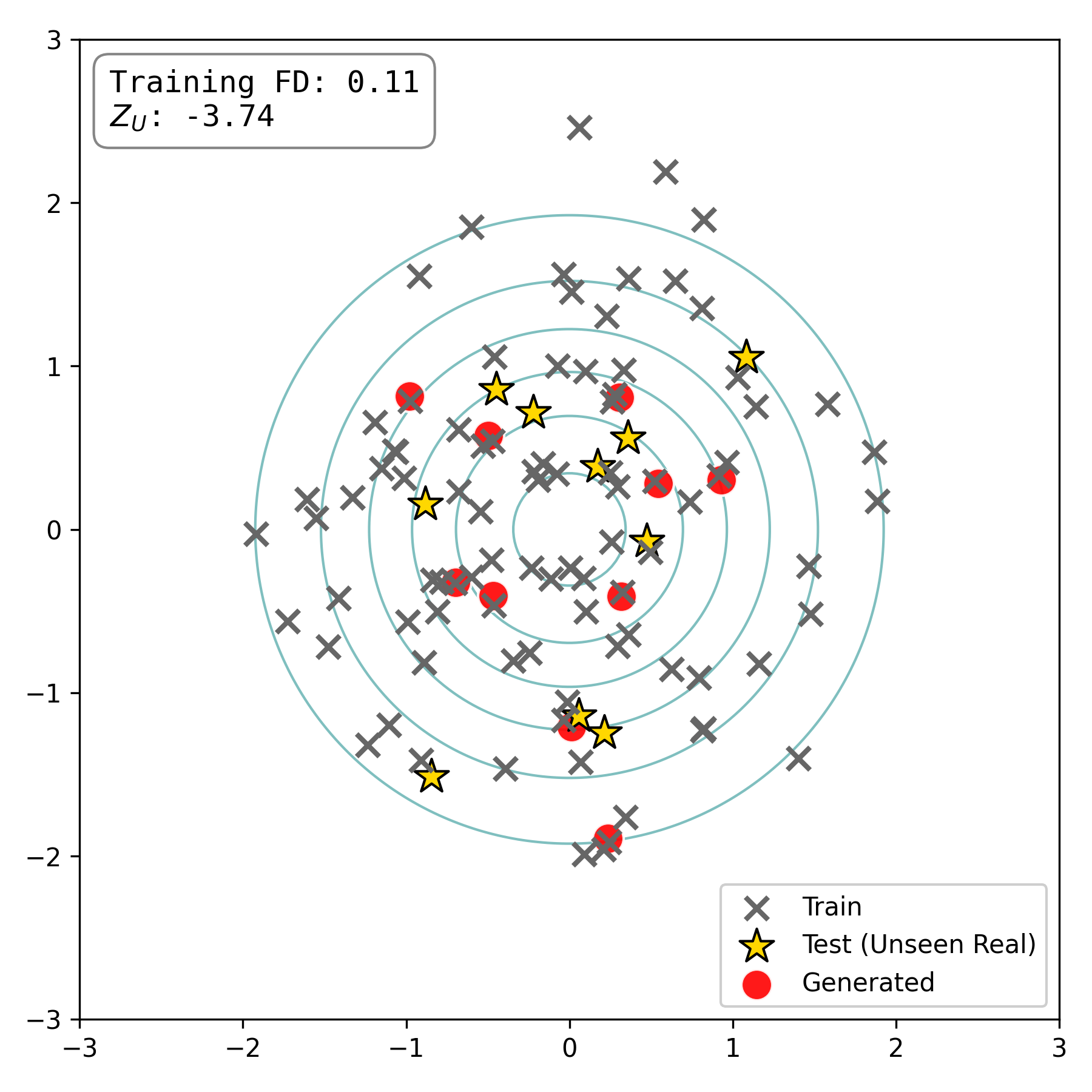}
        \caption{Memorization}
        \label{fig:toy_mem}
    \end{subfigure}
    \hfill
    \begin{subfigure}[b]{0.49\linewidth}
        \centering
        \includegraphics[width=\linewidth]{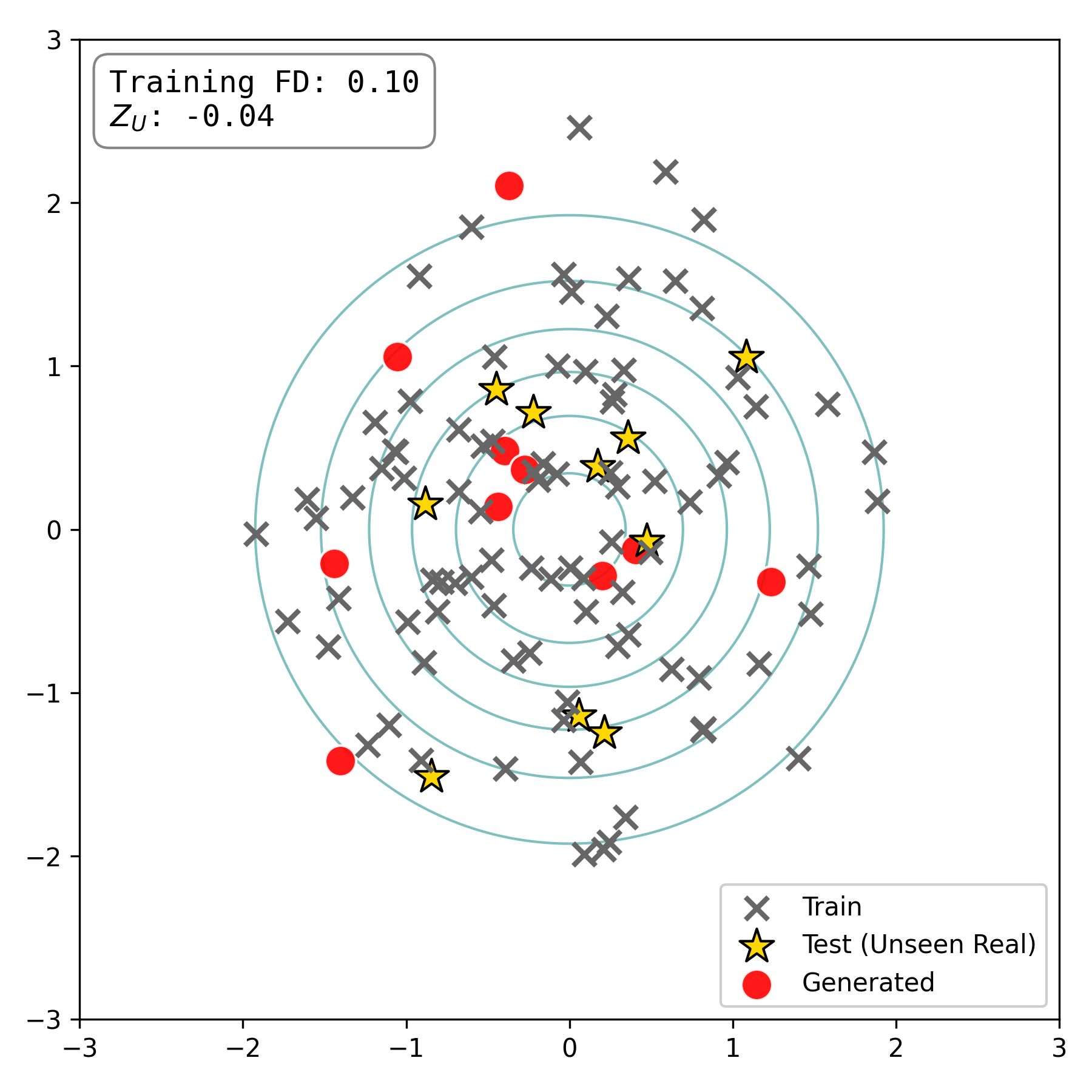}
        \caption{Generalization}
        \label{fig:toy_gen}
    \end{subfigure}%
    
    \caption{\textbf{Toy example illustrating the relationship between $Z_U$ and training FD.} 
    Grey crosses, gold stars, and red circles denote training, test, and generated data, respectively. Training and test data are sampled from a 2D Gaussian distribution. In the left plot, the generated samples memorize the training data, whereas in the right plot, the generated samples generalize. $Z_U$ (-3.74 \vs\ -0.04) captures the difference in memorization between the two scenarios, while training FD does not (0.11 \vs\ 0.10).}
    \label{fig:toy_example}
\end{figure}

\paragraph{Encoder.} We employ training FD and test FD as quality metrics. Based on Section~\ref{subsec:eval}, we select Uni3D as the best semantic encoder, and we use it in our FD calculation. 

\paragraph{Relation between $Z_U$ and training FD.} While training FD often correlates with memorization (\eg, a decrease in $Z_U$ generally accompanies a decrease in training FD), here, we use a toy example in Figure~\ref{fig:toy_example} to show that a low training FD may not always indicate memorization. In Figure~\ref{fig:toy_mem}, we draw training, test, and generated samples from a 2D Gaussian distribution, where the generated samples are created by sampling near the training points (\ie, memorization). In Figure~\ref{fig:toy_gen}, the generated samples follow the same distribution, but are not particularly close to the training data (\ie, generalization). $Z_U$ (-3.74 \vs\ -0.04) captures the difference in memorization between the two scenarios, whereas training FD is almost identical.

\begin{figure}[htbp] \centering \includegraphics[width=\linewidth]{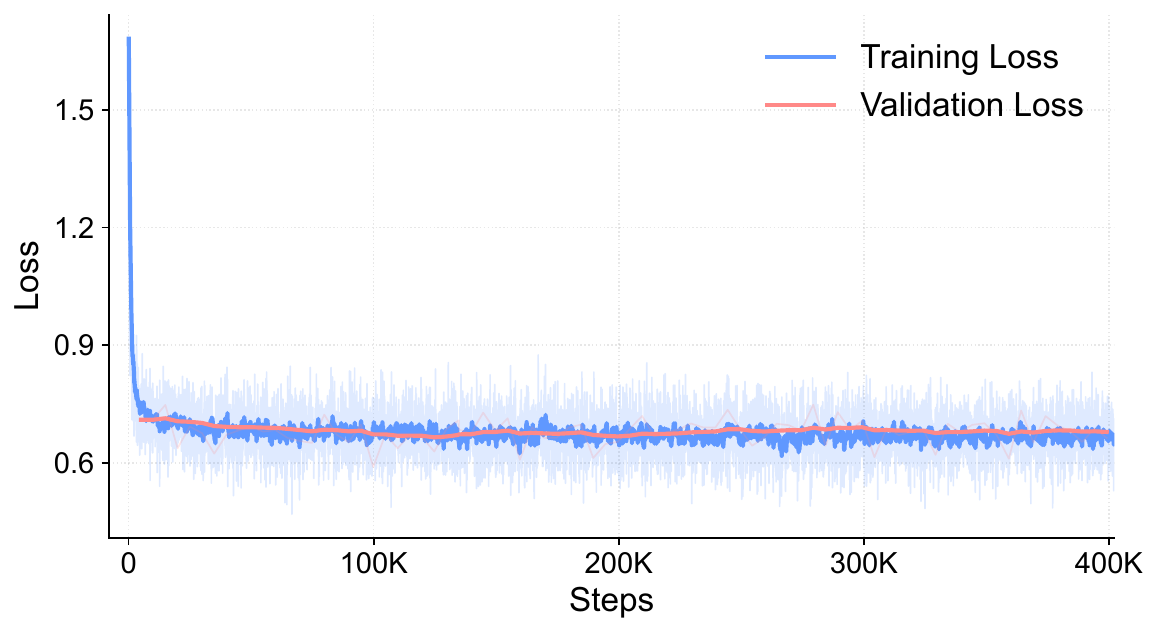} \caption{\textbf{Baseline model loss curves.} The training and validation losses of the baseline model throughout the training process.} \label{fig:base_loss} 
\end{figure}

\paragraph{Alternative metrics.} We observe that although the baseline model progresses from low-quality to high-quality generation throughout training (see Figure~\ref{fig:checkpoint-probing}), the training and validation losses fail to reflect this evolution. In Figure~\ref{fig:base_loss}, both losses converge at an early stage (approximately 10K steps) and then oscillate around 0.6. Consequently, we do not consider loss to be a reliable metric for quality evaluation.

\section{Existing Models' Retrieval Visualization}
\label{app:retrieval_res}
In this section, we visualize retrieval results for several existing models. Table~\ref{tab:model_training_data} summarizes the conditioning type, training data, and data split for all models considered.

\begin{table}[h]
    \centering
    \footnotesize
    \begin{tabularx}{\linewidth}{l l l l}
        \toprule
        model & cond. & training dataset & split \\
        \midrule
        NFD & Uncond & ShapeNet \textit{chair} & - \\
        Wavelet Generation & Uncond & ShapeNet \textit{chair} & IM-NET \\
        LAS-Diffusion & Uncond & ShapeNet \textit{chair} & IM-NET \\
        LAS-Diffusion & Class-cond & Five ShapeNet classes & IM-NET \\
        3DShape2VecSet & Class-cond & ShapeNet & 3DILG \\
        Michelangelo & Text-cond & ShapeNet & 3DILG \\
        3DTopia-XL & Text-cond & 256K Objaverse subset & - \\
        Trellis & Text-cond & Trellis500K & - \\
        \bottomrule
    \end{tabularx}
    \caption{\textbf{Conditioning type, training dataset, and dataset split for all evaluated models.} 
    A dash (-) indicates that the official split is not publicly released.
    }
    \vspace{-1em}
    \label{tab:model_training_data}
\end{table}

\begin{figure*}[t]
    \centering
    \includegraphics[width=\textwidth]{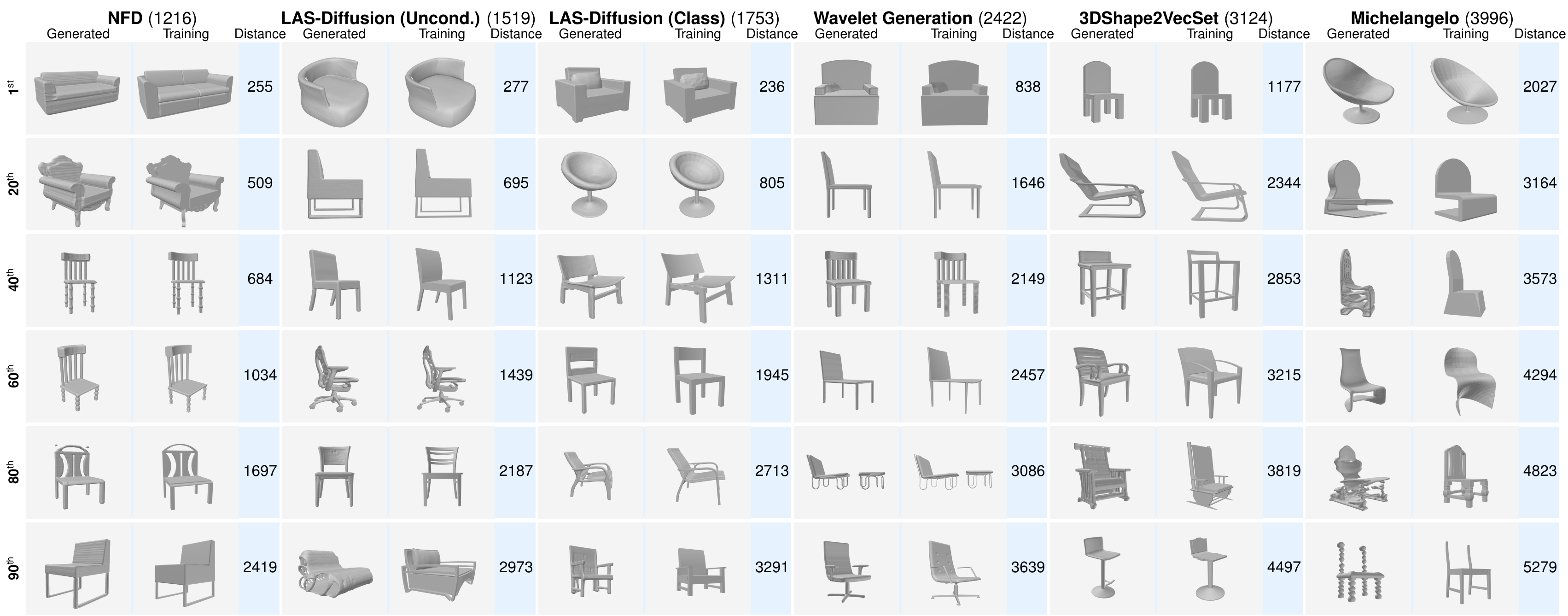}
    \caption{\textbf{Qualitative retrieval results on ShapeNet's \textit{chair} category.} We generate 100 chairs for each model and visualize the nearest training shapes for generated samples at the $1^{st}$, $20^{th}$, $40^{th}$, $60^{th}$, $80^{th}$, and $90^{th}$ percentiles of the LFD distance distribution for each model.
    NFD, unconditional LAS-Diffusion, and Wavelet Generation exhibit strong memorization: even generated shapes at the $60^{th}$-$80^{th}$ percentiles remain very close to their nearest training shapes.
    In contrast, conditional LAS-Diffusion, 3DShape2VecSet, and Michelangelo show novel geometric features even for generated samples with lower nearest-neighbor distances, indicating stronger generalization.}
    \label{fig:retrieval_chair}
    \vspace{-1em}
\end{figure*}

\begin{figure*}[b]
    \centering
    \includegraphics[width=\textwidth]{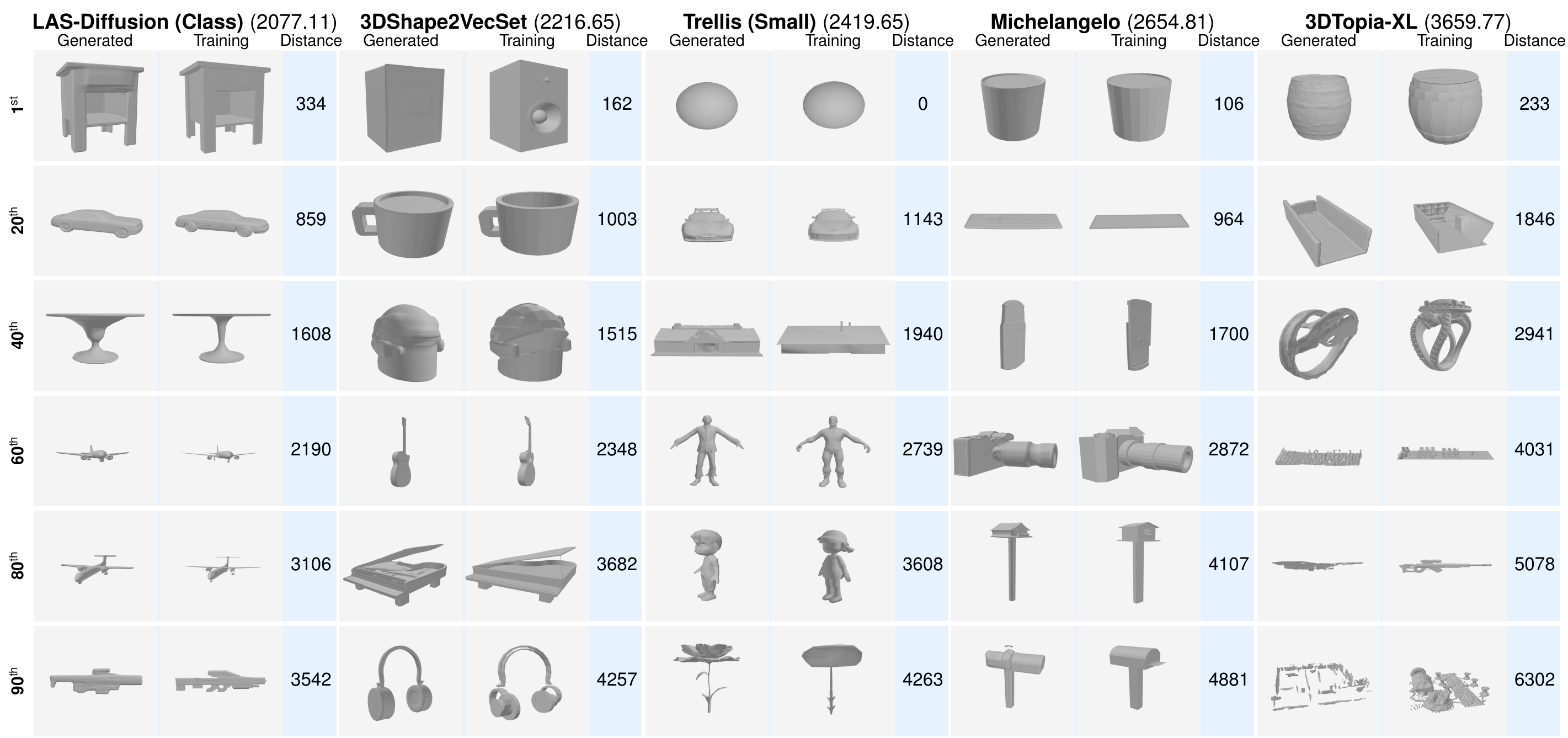}
    \caption{\textbf{Qualitative retrieval results on the entire training sets.}
    We generate 100 samples for each model and visualize the nearest training shapes for generated samples at the $1^{st}$, $20^{th}$, $40^{th}$, $60^{th}$, $80^{th}$, and $90^{th}$ percentiles of the nearest-neighbor LFD distance distribution for each model trained on large datasets.
    Across all models, the retrieved training shapes already become novel at moderate percentiles (\eg, $20^{th}$-$60^{th}$).
    Although LFD is category-sensitive and may not always retrieve visually near-identical shapes, the overall trends suggest that these models trained on large datasets primarily generalize rather than copy individual training examples. We note that 3DTopia-XL does not generalize well and often degenerates, producing low-quality shapes even when using training prompts.}
    \label{fig:retrieval_full}
    \vspace{-1em}
\end{figure*}

Figure~\ref{fig:retrieval_chair} shows retrieval results for six models on the \textit{chair} category. Figure~\ref{fig:retrieval_full} shows retrieval results for five models on their respective full training sets.

The trends are consistent with the quantitative results in Section~\ref{sec:exist}: models trained on smaller datasets (NFD, LAS-Diffusion, and Wavelet Generation) exhibit strong memorization; even among generated chairs in the $60^{\text{th}}$--$80^{\text{th}}$ percentiles (ranked by distance to the nearest training shape), many closely match training shapes.
In contrast, models trained on larger and more diverse datasets show strong generalization, with even generated shapes at lower percentiles being noticeably more diverse and exhibiting novel features.

\section{Detailed Experimental Setups}
\label{app:controlled_exp}

This section details the model architectures, datasets, and training configurations employed in the controlled experiments of Section~\ref{Sec:FactorExploration} and~\ref{sec:mitigation}. Furthermore, we include supplementary experiments and qualitative visualizations to support the findings presented in the main text.

\subsection{Dataset and Model Setup}
\label{app:dataset_model}
\paragraph{Dataset pipeline.} Our dataset is sourced from Objaverse-XL~\citep{deitke2023objaversexl}. We rank classes by frequency within the Objaverse-LVIS subset and select the 100 most common single-object classes. Rather than re-captioning the entire Objaverse-XL, we use the existing Cap3D captions~\citep{luo2023scalable}.

\begin{figure}[h]
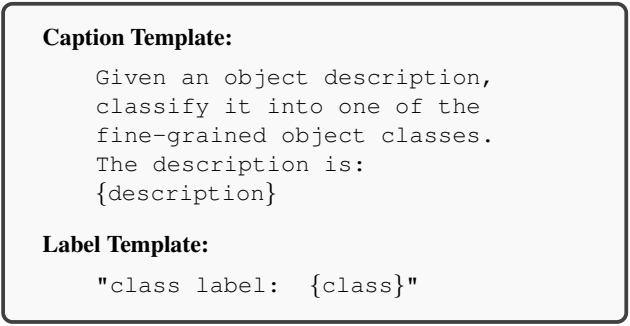

\centering
\begin{tcolorbox}[colback=gray!5!white]
\small

\textbf{Caption Template:}
\begin{quote}
\texttt{Given an object description, classify it into one of the fine-grained object classes.\\
The description is: \{description\}}
\end{quote}

\vspace{0.5em}
\textbf{Label Template:}
\begin{quote}
\texttt{"class label: \{class\}"}
\end{quote}

\end{tcolorbox}
\vspace{-10pt}
\caption{\textbf{Instruction template} used to encode Cap3D captions and object labels with Qwen3-8B Embedding.}
\label{fig:embed_template}
\end{figure}

To ensure data quality, we filter caption-label pairs for semantic alignment using Qwen3-8B Embedding~\citep{zhang2025qwen3emb}.
We compute caption and class label embeddings using the instruction templates shown in Figure~\ref{fig:embed_template} and measure their cosine similarity.
We retain approximately 229K caption-label pairs by applying a similarity threshold of $\tau = 0.7$.

\begin{figure}[h]
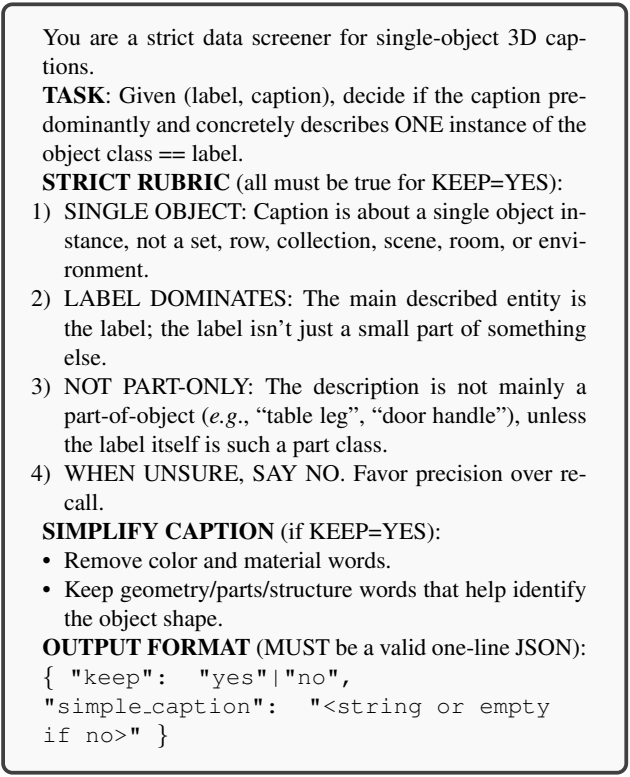

\centering

\begin{tcolorbox}[colback=gray!5!white, colframe=gray!50!black, colbacktitle=gray!75!black]
\small

You are a strict data screener for single-object 3D captions.

\textbf{TASK}:
Given (label, caption), decide if the caption predominantly and concretely describes ONE instance of the object class == label.

\textbf{STRICT RUBRIC} (all must be true for KEEP=YES):
\begin{itemize}
    \item[1)] SINGLE OBJECT: Caption is about a single object instance, not a set, row, collection, scene, room, or environment.
    \item[2)] LABEL DOMINATES: The main described entity is the label; the label isn't just a small part of something else.
    \item[3)] NOT PART-ONLY: The description is not mainly a part-of-object (\eg, ``table leg'', ``door handle''), unless the label itself is such a part class.
    \item[4)] WHEN UNSURE, SAY NO. Favor precision over recall.
\end{itemize}

\textbf{SIMPLIFY CAPTION} (if KEEP=YES):
\begin{itemize}
    \item Remove color and material words.
    \item Keep geometry/parts/structure words that help identify the object shape.
\end{itemize}

\textbf{OUTPUT FORMAT} (MUST be a valid one-line JSON):\\
\texttt{\{ "keep": "yes"|"no", "simple\_caption": "<string or empty if no>" \}}

\end{tcolorbox}
\vspace{-7pt}
\caption{Prompt - Filtering out misaligned caption-label pairs.}
\label{fig:prompt_mislabel}
\vspace{-5pt}
\end{figure}

\begin{figure}[t]
    \centering
    \includegraphics[width=\linewidth]{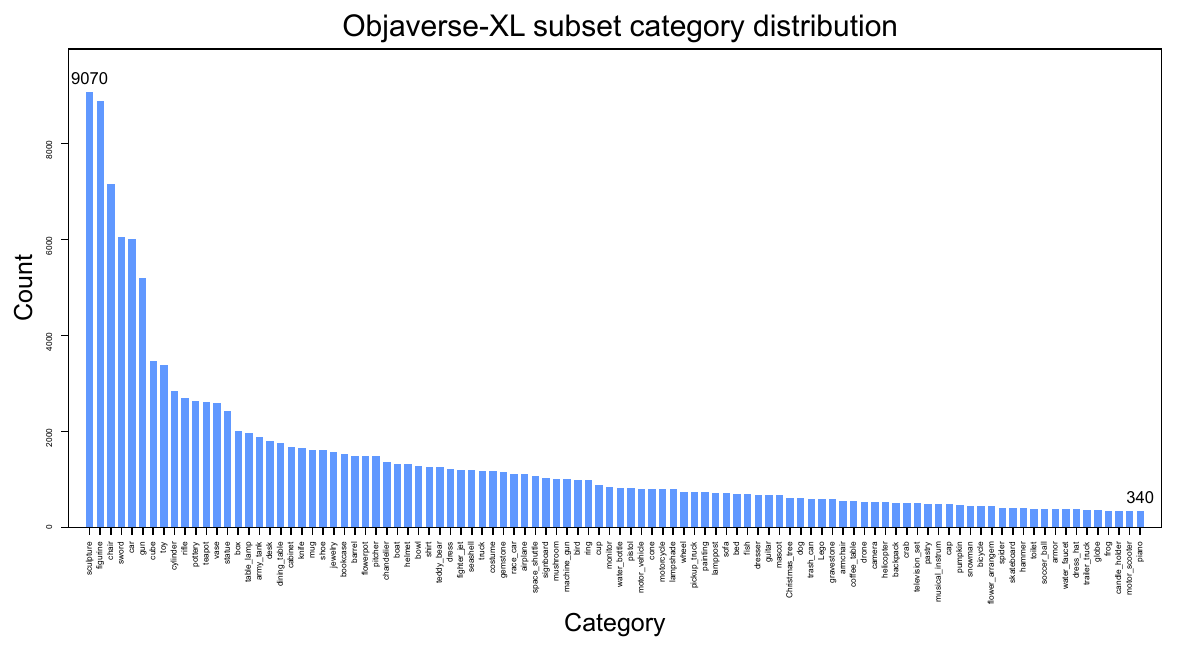}
    \vspace{-1.5em}
    \caption{\textbf{Category distribution} of our customized 100-category subset of Objaverse-XL.}
    \label{fig:class_dist}
    \vspace{-0.5em}
\end{figure}

Then, we apply Qwen3-30B~\citep{yang2025qwen3} to aggressively remove mislabeled pairs.
As shown in Figure~\ref{fig:prompt_mislabel}, we keep only pairs for which the LLM is confident and discard low-quality captions.
To reduce class imbalance, we cap the most frequent object class at 10K examples.
As a result, we obtain around 140K caption-label pairs, which we randomly split into 120K for training and 20K for testing. The class distribution of our customized dataset is shown in Figure~\ref{fig:class_dist}.

\paragraph{Vecset autoencoder.}
We use the latest pre-trained Vecset autoencoder, VecSetX, released by 3DShape2VecSet~\citep{3DShape2Vecset}.
The model takes sub-sampled input point embeddings as queries and is trained with an SDF regression objective~\citep{li2025triposg} and an Eikonal regularizer~\citep{gropp2020implicit}.
The bottleneck follows the normalized bottleneck autoencoder design~\citep{zhang2024lagem}.
The network consists of 24 layers of self-attention blocks with a hidden dimension of 1024, and the resulting latent code has shape $1024 \times 32$ by default.

Formally, let $D(x, f)$ denote the predicted SDF value at any query spatial location $x$ given latent code $f$, and let $s(x)$ denote the ground-truth signed distance.
The autoencoder is trained using a combination of an SDF regression loss and an Eikonal loss:
\begin{equation}
    \mathcal{L}
    = \mathcal{L}_{\mathrm{SDF}}
    + \lambda_{\mathrm{eik}} \,\mathcal{L}_{\mathrm{eik}}.
\end{equation}
The SDF regression term is:
\begin{equation}
    \mathcal{L}_{\mathrm{SDF}}
    = \mathbb{E}_x\bigl[\lvert D(x,f)-s(x)\rvert\bigr],
\end{equation}
and the Eikonal regularization term is:
\begin{equation}
\mathcal{L}_{\mathrm{eik}}
= \mathbb{E}_{x}\Bigl[ \bigl(\lVert \nabla_{x} D(x,f)\rVert_2 - 1\bigr)^2 \Bigr].
\end{equation}

\paragraph{Diffusion backbone.}
We build upon the Hunyuan3D 2.1 codebase~\citep{hunyuan3d2025hunyuan3d}. 
The architecture is a flow-based diffusion model composed of multiple Hunyuan-DiT blocks~\citep{li2024hunyuan}. 
It uses cross-attention layers to inject conditioning signals and replaces the feed-forward layers in the last few Transformer blocks with mixture-of-experts (MoE) layers.

\subsection{Baseline and Sanity Check}
\label{app:baseline}
In this section, we introduce three model sizes and describe controlled experiments on dataset size and model size.

\paragraph{Baseline model configuration.}
Our baseline model comprises approximately 323M parameters, consists of 12 Transformer blocks with a hidden dimension of 1024, and 16 attention heads.
Following the official Hunyuan3D 2.1 implementation, we disable positional embeddings and attention pooling.
The model is text-conditional: we use CLIP-B/16~\citep{radford2021learning} as the text encoder, giving a conditioning dimension of 512.
We use three MoE layers in the backbone, each with four experts and top-2 gating.

\begin{table}[htbp]
    \centering
    \resizebox{\linewidth}{!}{
        \begin{tabular}{lcccc}
            \toprule
            model & \#params & blocks & hidden dim & attn heads \\
            \midrule
            small    & 80M   & 12 & 512  & 8  \\
            baseline & 323M  & 12 & 1024 & 16 \\
            large    & 1.5B  & 16 & 2048 & 16 \\
            \bottomrule
        \end{tabular}
    }
    \caption{\textbf{Hyperparameter specifications for model variants.} All models share the same text encoder and MoE settings.}
    \label{tab:model_configs}
\end{table}

\paragraph{Model scaling variants.}
To investigate the impact of model capacity, we introduce a lightweight small model and a scaled-up large model. 
These variants retain the same architectural components as the baseline, including the CLIP-B/16 text encoder, MoE configuration, and the lack of positional embeddings, but vary in network depth and width. 
We summarize the specific hyperparameter settings for all three model configurations in Table~\ref{tab:model_configs}.

\begin{figure}[htbp]
    \centering
    \includegraphics[width=\linewidth]{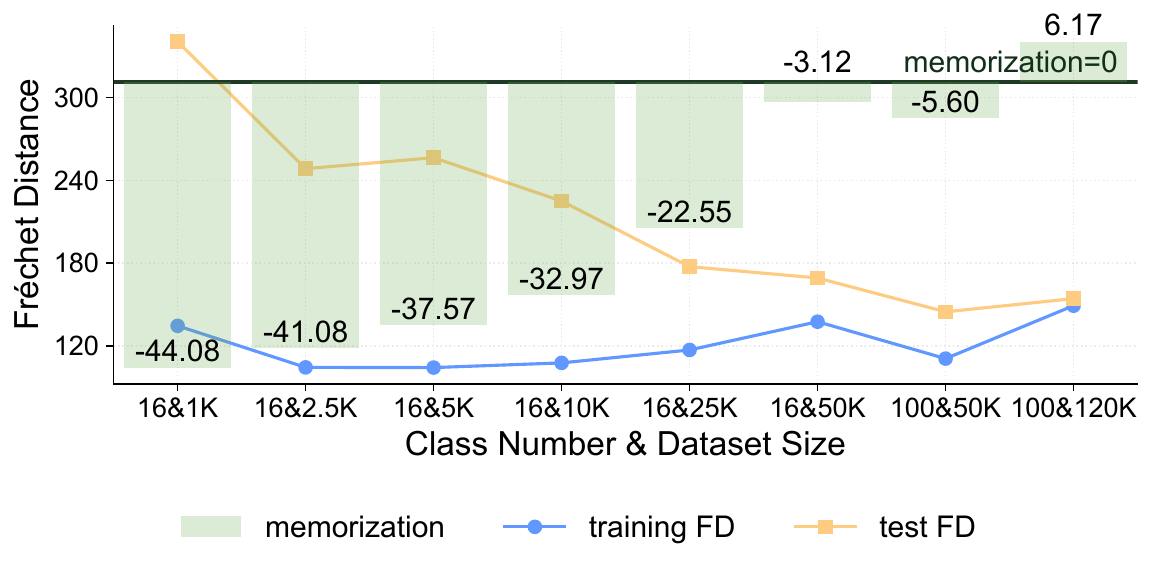}
    \caption{\textbf{Larger datasets reduce memorization.} With the same diffusion model, increasing the dataset size decreases $Z_U$.}
    \label{fig:dataset_probing}
\end{figure}

\paragraph{Dataset size sanity check.}
We train our baseline model with several class numbers and dataset size settings. Figure~\ref{fig:dataset_probing} shows that as the dataset size scales up, the memorization behavior is substantially reduced. Since the 16-class subset contains only about 58K training shapes, for the 16-class experiments, we only use up to 50K samples.

\begin{figure}[htbp]
    \centering
    \includegraphics[width=\linewidth]{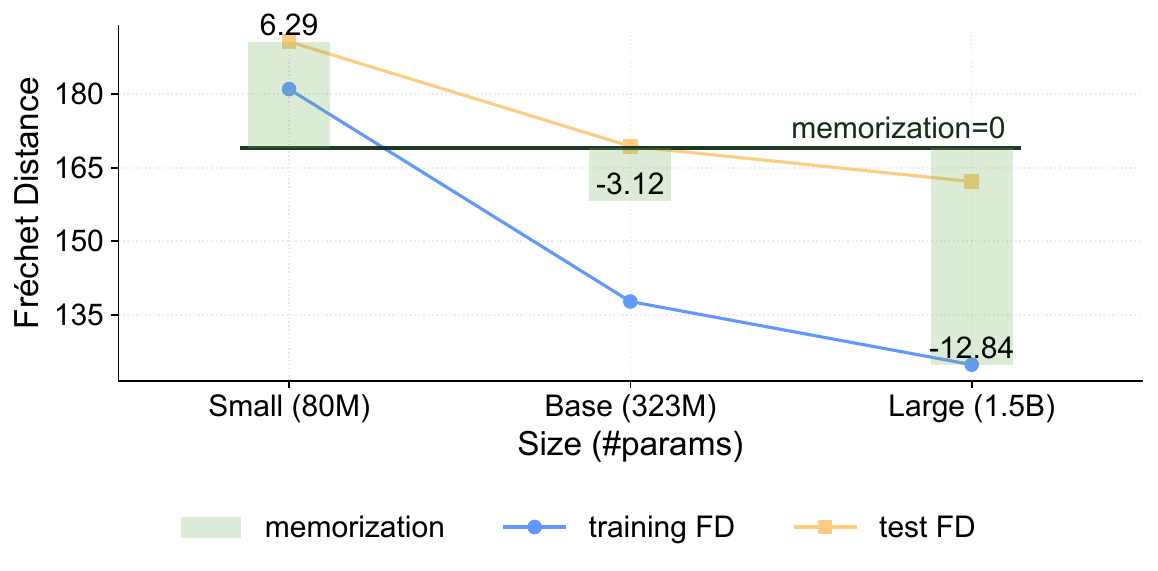}
    \caption{\textbf{Larger models strengthen memorization.} With the same training dataset, increasing the model size decreases $Z_U$.}
    \label{fig:size_probing}
\end{figure}

\paragraph{Model size sanity check.}
We also evaluate how model size affects memorization.
We train the small, baseline, and large models on a 16-class subset of 50K shapes for 200K steps.
Figure~\ref{fig:size_probing} shows that larger models exhibit stronger memorization.
It is worth noting that the small model fails to generate high-quality shapes and has high FD on both the training and test sets.

Our experimental results are consistent with the existing literature~\citep{somepalli2022diffusion,carlini2023extracting_training,zeng2025generative} on the impact of dataset size and model size on memorization.

\subsection{Data Modality}
\label{app:data_modality}
In this section, we provide additional details on the implementation and training of the image generation model, along with supplementary experimental results.

\paragraph{Dataset.} Instead of using the baseline model's 16-class training data, we use the 100-class dataset. We ensure this dataset is identical to the 50K dataset used for the 100-class model presented in Figure~\ref{fig:dataset_probing}.

\paragraph{Rendering pipeline.} We render the $256\times256$ image dataset following the procedures detailed in Appendix~\ref{app:rendering}. To constrain the dataset size to 50K, we exclusively use the sixth view from the 12 renderings available for each object.

\paragraph{Image autoencoder.} We use the Flux VAE to encode input images of resolution $3 \times 256 \times 256$ into latent grids of shape $16 \times 32 \times 32$. After injecting 2D positional embeddings, we flatten the spatial dimensions to yield a latent sequence of size $1024 \times 16$, which serves as the input to the diffusion model. We select the Flux VAE not only for its superior reconstruction capabilities but also to ensure the latent sequence length aligns with our Vecset configuration.

\begin{table}[htbp]
    \centering
    \small
    \setlength{\tabcolsep}{6pt}
    \begin{tabular}{lccc}
        \toprule
        modality & DinoV2 {\scriptsize (\%)} & SSCD {\scriptsize (\%)} & LFD {\scriptsize (\%)}\\
        \midrule
        image & \textbf{52.71} & \textbf{68.05} & - \\
        3D & 22.04 & 32.56 & 46.44\\
        \bottomrule
    \end{tabular}
    \caption{\textbf{Source data retrieval rate using the same 2500 training prompts for both image and 3D generative models.} \textbf{Higher} retrieval rate means higher probability of reproducing the source data. Images show a stronger tendency to replicate than 3D shapes.}
    \label{tab:prompt_img_3d}
\end{table}

\paragraph{Source data retrieval rate} measures how often the nearest training sample to a generated sample is exactly the ground truth sample for the input prompt. While previous work~\citep{somepalli2022diffusion} on large-scale text-to-image models reports that training prompts rarely regenerate their exact source images, Table~\ref{tab:prompt_img_3d} shows that our image model has a higher probability of retrieving training data than the 3D generative model when conditioned on training captions. This further demonstrates that images are more susceptible to memorization.

\subsection{Data Diversity}
\label{app:data_diversity}
Our experiments use the 16, 32, 64, and 100 most frequent classes in the original dataset. For each class subset, we sample 50K examples by taking samples from each class in proportion to its frequency within the selected class subset.

\subsection{Conditioning Granularity}
\label{app:cond_granularity}
In this section, we provide further information on how we construct class and text conditions of different granularities.

\begin{figure}[htbp]
    \centering
    \includegraphics[width=\linewidth]{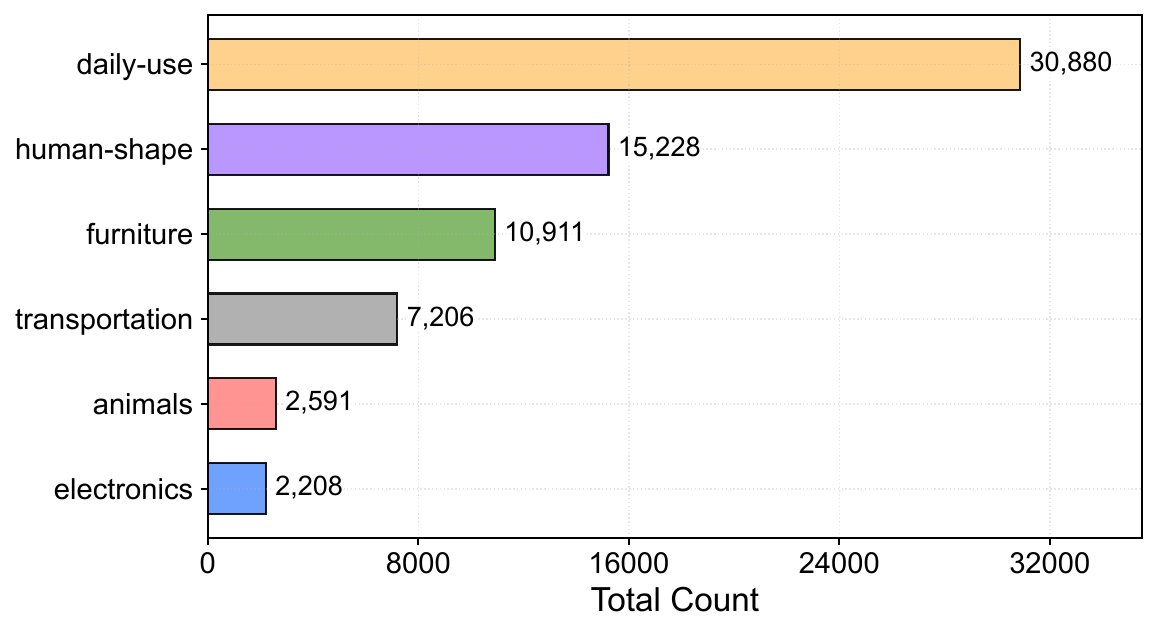}
    \caption{\textbf{Coarse-grained category distribution.} We use six categories (animals, daily-use, electronics, furniture, human-shape, and transportation) from the GObjaverse taxonomy.}
    \label{fig:gcat_dist}
    \vspace{-1.0em}
\end{figure}

\paragraph{Coarse-grained class labels.} We use the same dataset as the baseline but introduce an additional set of coarse-grained class labels.
We adopt the GObjaverse taxonomy, which originally comprises 10 categories: animals, daily-use, electronics, furniture, human-shape, transportation, buildings\&outdoor, plants, food, and poor-quality. 
However, since the distribution of our top-16 classes (Figure~\ref{fig:class_dist}) shows minimal representation of buildings\&outdoor, plants, and food, we exclude these categories from our experiments. 
Additionally, we remove the poor-quality category and re-classify any abstract shape descriptions as daily-use. 
Consequently, each shape in our dataset is assigned to one of the remaining 6 coarse-grained categories (animals, daily-use, electronics, furniture, human-shape, and transportation) using the prompt shown in Figure~\ref{fig:prompt_classification}. The coarse-grained category distribution is shown in Figure~\ref{fig:gcat_dist}.

\begin{figure}[h]
\centering
\begin{tcolorbox}[colback=gray!5!white, colframe=gray!50!black, colbacktitle=gray!75!black]
\small

You are a precise data labeler.

\textbf{TASK}:
Given a label and a caption, choose exactly ONE super-category from the provided CLOSED SET.

\textbf{ALLOWED CATEGORIES}:
\begin{itemize}[nosep]
    \item animals, daily-use, electronics, furniture, human-shape, transportation
\end{itemize}

\textbf{GUIDANCE}:
\begin{itemize}
    \item Use the caption semantics to determine the super-category.
    \item \textbf{Semantic Rule}: If the description suggests a toy, LEGO, or miniature but depicts a real-world concept (\eg, a toy car), choose the semantic category (\eg, Transportation).
    \item \textbf{Closed Set}: Never invent categories; pick ONLY from the allowed list.
    \item \textbf{Fallback}: If classification is truly impossible, select ``daily-use''.
\end{itemize}

\textbf{OUTPUT FORMAT} (Strict one-line JSON):\\
\texttt{\{ "category": "<Category Name>" \}}

\end{tcolorbox}
\vspace{-7pt}
\caption{Prompt - GObjaverse category classification.}
\label{fig:prompt_classification}
\vspace{-1em}
\end{figure}

\begin{figure}[htbp]
    \centering
    \includegraphics[width=\linewidth]{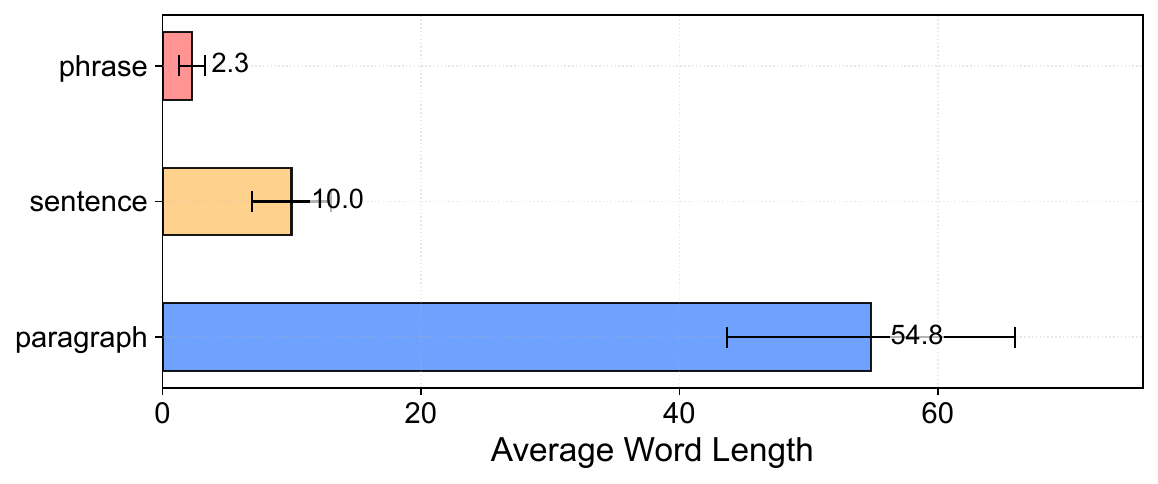}
    \caption{\textbf{Caption length distribution across granularities.} The distribution of caption length for phrase, sentence, and paragraph captions demonstrates distinct levels of detail.}
    \label{fig:caption_dist}
    \vspace{-1.0em}
\end{figure}

\paragraph{Multi-granularity captioning.}
We generate captions using Qwen3-VL~\citep{Qwen3-VL} across three distinct levels of granularity: phrase, sentence, and paragraph. To prompt Qwen3-VL, we randomly select four views from the 12 rendered images detailed in Appendix~\ref{app:rendering}. We show the lengths of different text granularities in Figure~\ref{fig:caption_dist}. The prompt used for the multi-granularity captioning is shown in Figure~\ref{fig:prompt_multiview}.

\begin{figure}[h]
\centering
\begin{tcolorbox}[colback=gray!5!white, colframe=gray!50!black, colbacktitle=gray!75!black]
\small

You are given multiple views of the SAME object captured from different angles.

\textbf{TASK}:
Combine evidence across views and describe the object at three distinct granularities.

\textbf{GRANULARITY DEFINITIONS}:
\begin{itemize}
    \item \textbf{Phrase}: A concise 3-10 word noun phrase (no punctuation at the end).
    \item \textbf{Sentence}: 1-2 sentences summarizing main parts, colors, and overall shape.
    \item \textbf{Paragraph}: 3-6 sentences covering fine details like materials, textures, distinctive features, geometry, and any context seen in the background.
\end{itemize}

\textbf{CONSTRAINTS}:
\begin{itemize}
    \item Return \textbf{ONLY} a compact JSON object.
    \item No extra text. No markdown. No newlines.
    \item Avoid speculation; if something is unknown, say 'unknown'.
\end{itemize}

\textbf{OUTPUT FORMAT} (Must use these exact keys):\\
\texttt{\{ "phrase": "...", "sentence": "...", "paragraph": "..." \}}

\end{tcolorbox}
\vspace{-7pt}
\caption{Prompt - Multi-granularity caption generation.}
\label{fig:prompt_multiview}
\end{figure}

\paragraph{Class-conditional model configuration.} Our class-conditional model shares nearly the same architecture as our baseline model; the only difference lies in the conditioning embedder. Instead of using a text encoder, we construct a trainable 512-dimensional embedding matrix. For each class label, we learn a corresponding class embedding vector, which is then injected into the model via cross-attention.

\subsection{Guidance Scale}
\label{app:gs}

\begin{figure}[htbp]
    \centering
    \includegraphics[width=\linewidth]{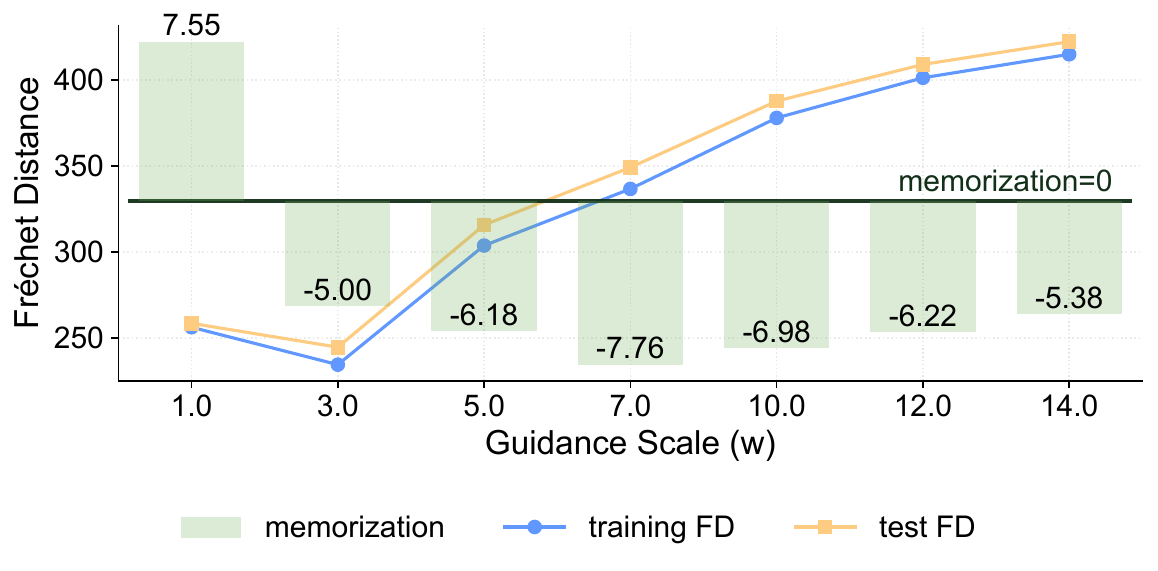}
    \caption{
    \textbf{Moderate guidance scales lead to the strongest memorization (class-conditional checkpoint)}, while further increasing the guidance scale reduces memorization, consistent with Section~\ref{subsec:GS}.
    Weak guidance also yields low memorization.}
    \label{fig:gs-app}
    \vspace{-1.0em}
\end{figure}

In Section~\ref{subsec:GS}, we evaluate six distinct guidance scales, $w \in \{0,1,3,5,7,10\}$, across both our baseline and large models (details in Appendix~\ref{app:baseline}). Here, we provide an additional guidance scale experiment in Figure~\ref{fig:gs-app}. Specifically, we run the same evaluation on a class-conditional checkpoint (details in Appendix~\ref{app:cond_granularity}).
This additional result is consistent with the conclusion in Section~\ref{subsec:GS}: memorization is strongest at moderate guidance scales, while stronger guidance reduces memorization. 

\begin{table}[htbp]
\centering
\tablestyle{5pt}{1.2}
\begin{tabular}{lcccc}
$w$ & gen & top match & caption (top match) & LFD \\
\shline
1 & \adjustbox{valign=c}{\includegraphics[width=.12\linewidth]{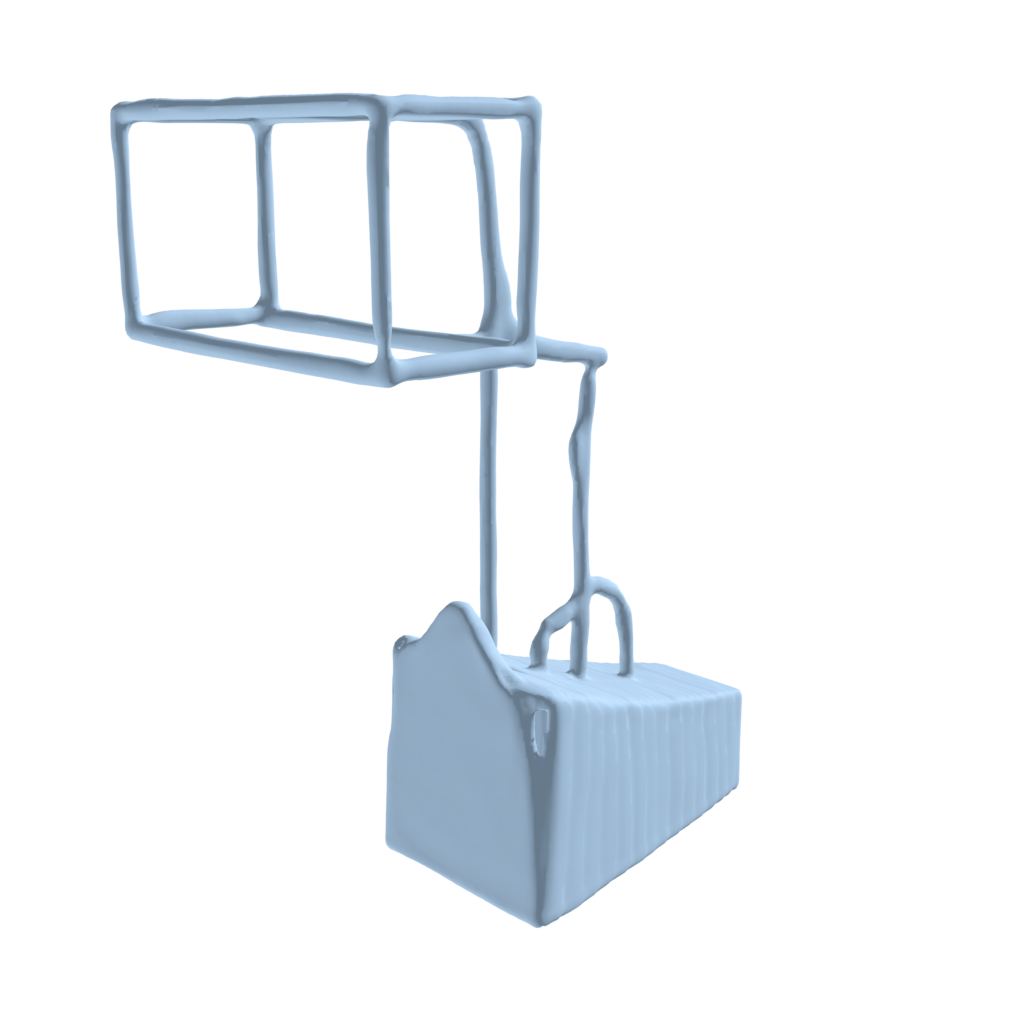}} & \adjustbox{valign=c}{\includegraphics[width=.12\linewidth]{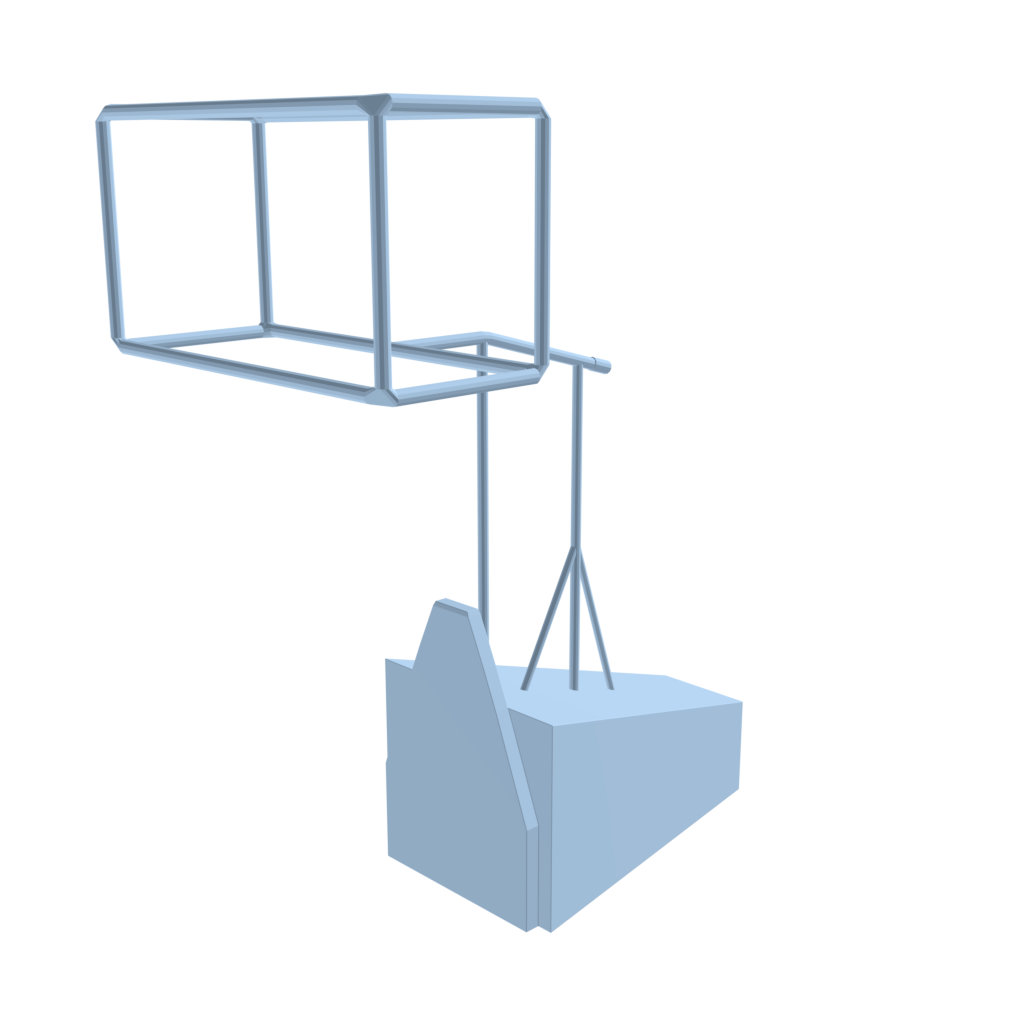}} & \adjustbox{valign=c}{\makecell{``a geometric sculpture\\ with a trapezoidal base\\ and a rectangular prism\\ suspended above it by rods''}} & 2427 \\
3 & \adjustbox{valign=c}{\includegraphics[width=.12\linewidth]{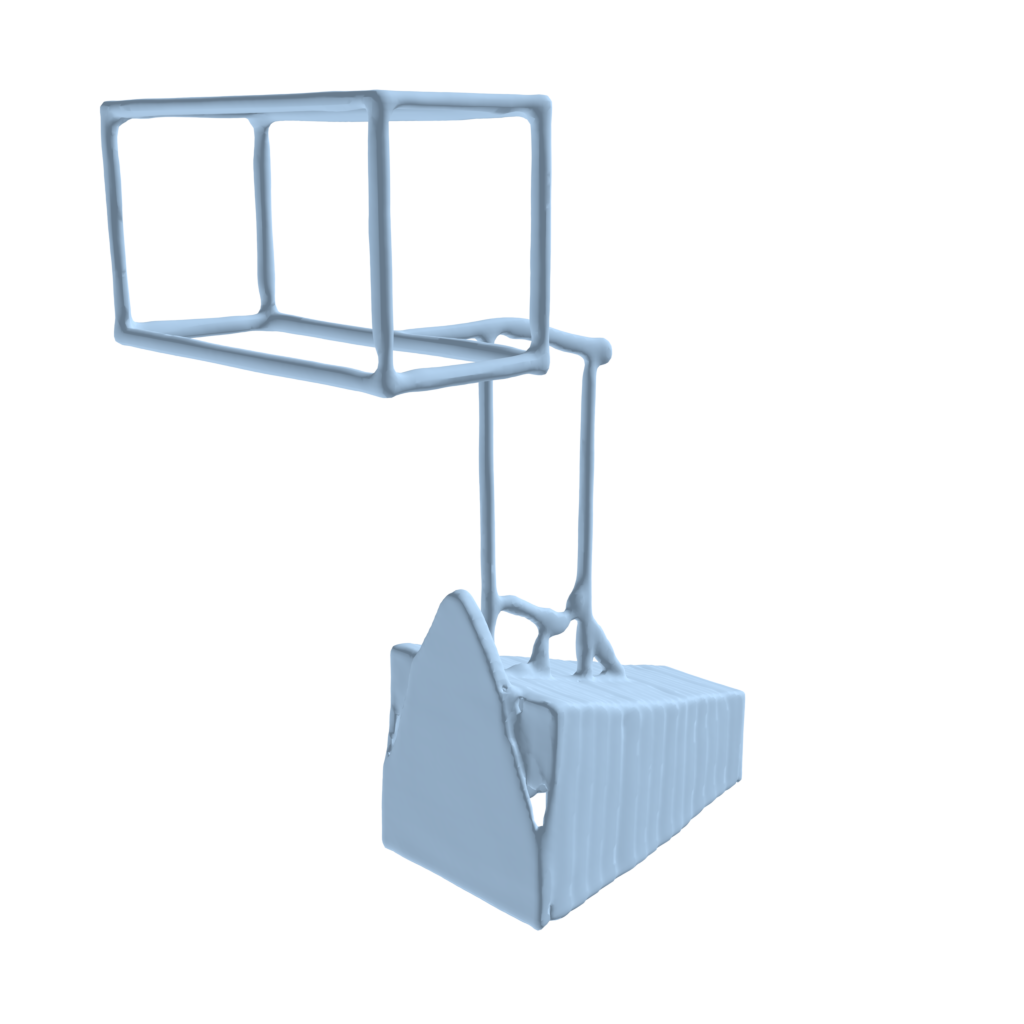}} & \adjustbox{valign=c}{\includegraphics[width=.12\linewidth]{figs/Appendix/GS-Cases/Basket/Ref.png}} & \adjustbox{valign=c}{\makecell{Same as $w=1$}} & 1282 \\
5 & \adjustbox{valign=c}{\includegraphics[width=.12\linewidth]{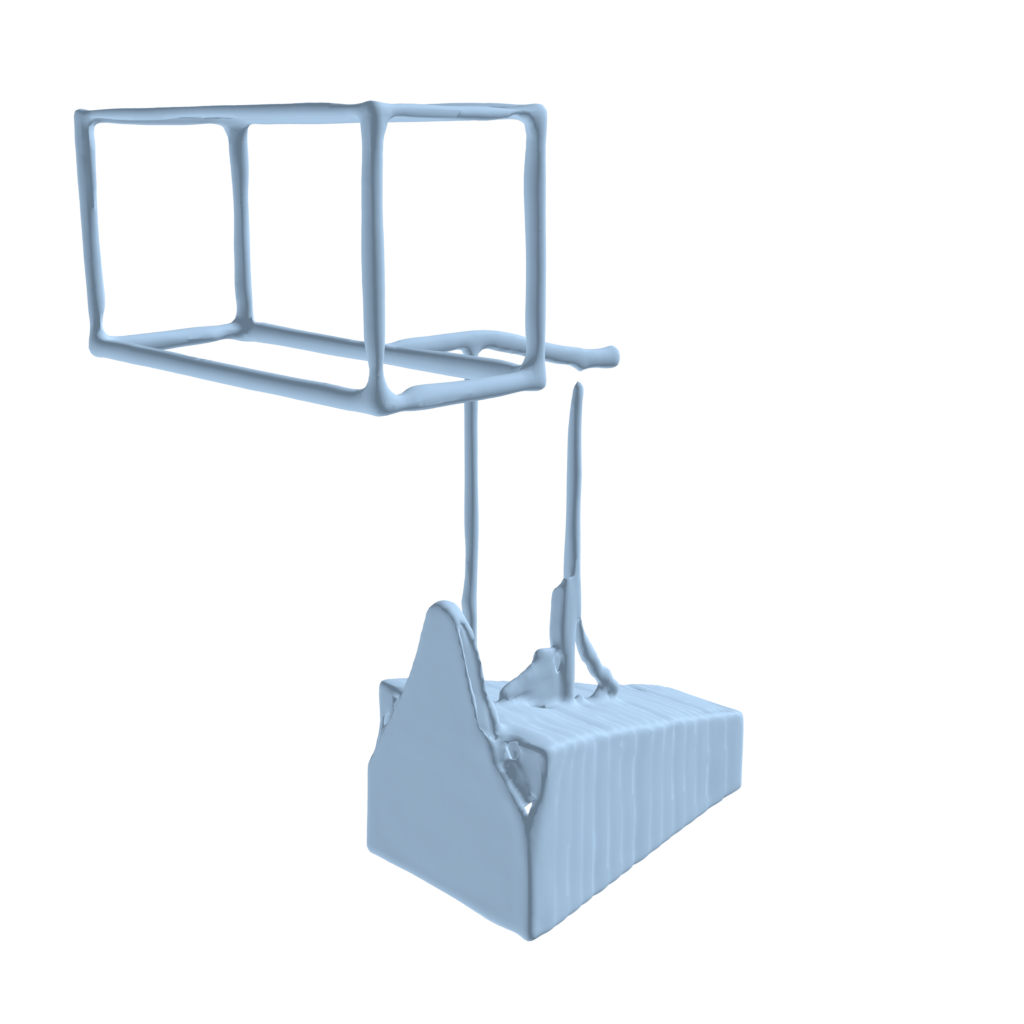}} & \adjustbox{valign=c}{\includegraphics[width=.12\linewidth]{figs/Appendix/GS-Cases/Basket/Ref.png}} & \adjustbox{valign=c}{\makecell{Same as $w=1$}} & 2467 \\
7 & \adjustbox{valign=c}{\includegraphics[width=.12\linewidth]{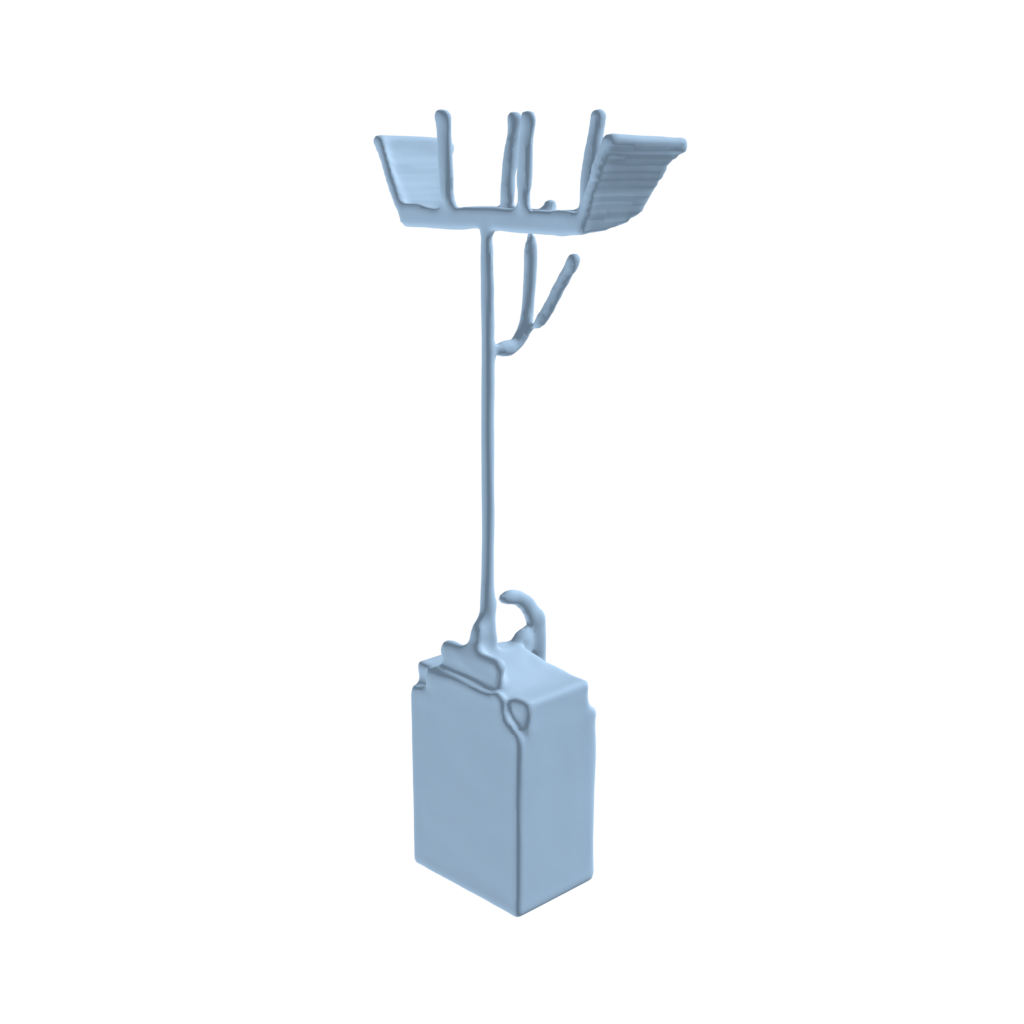}} & \adjustbox{valign=c}{\includegraphics[width=.12\linewidth]{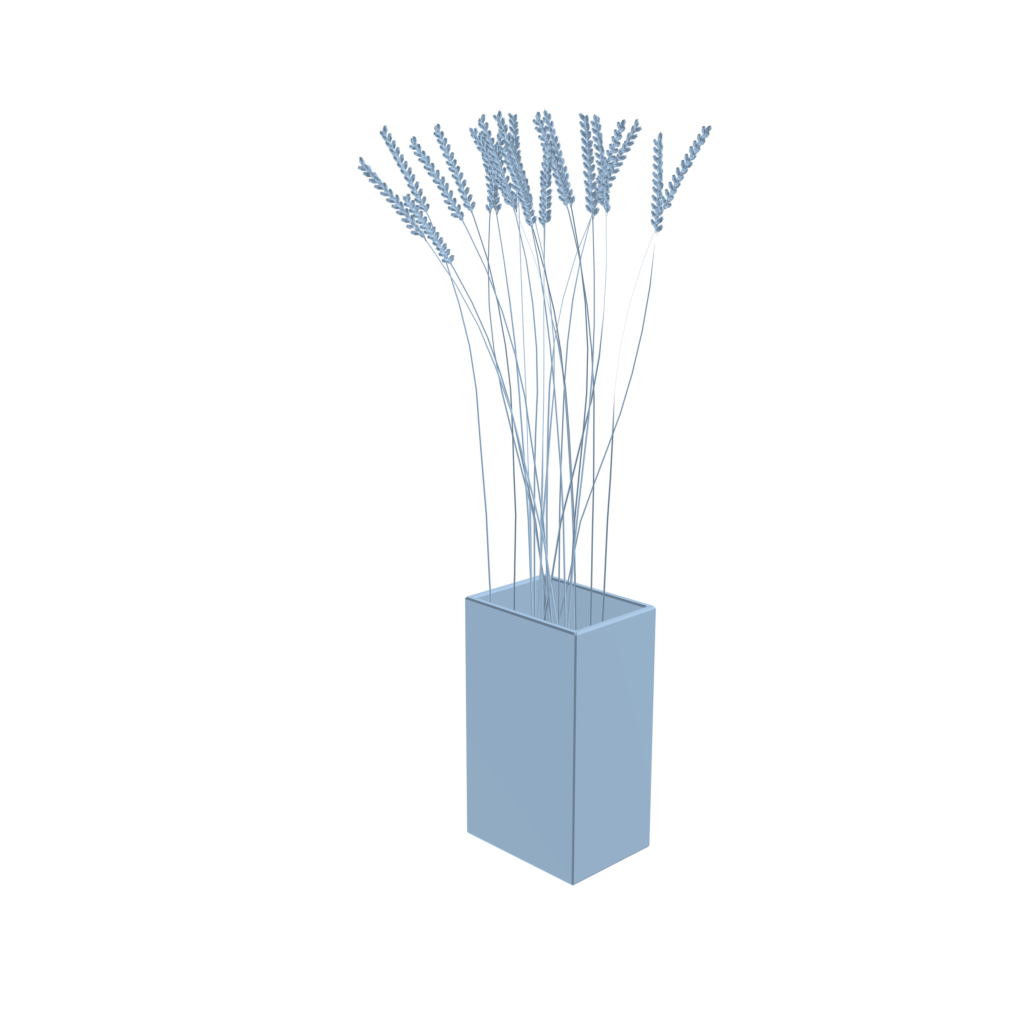}} & \adjustbox{valign=c}{\makecell{``round base, vertical stem,\\ domed shade''}} & 7555 \\
10 & \adjustbox{valign=c}{\includegraphics[width=.12\linewidth]{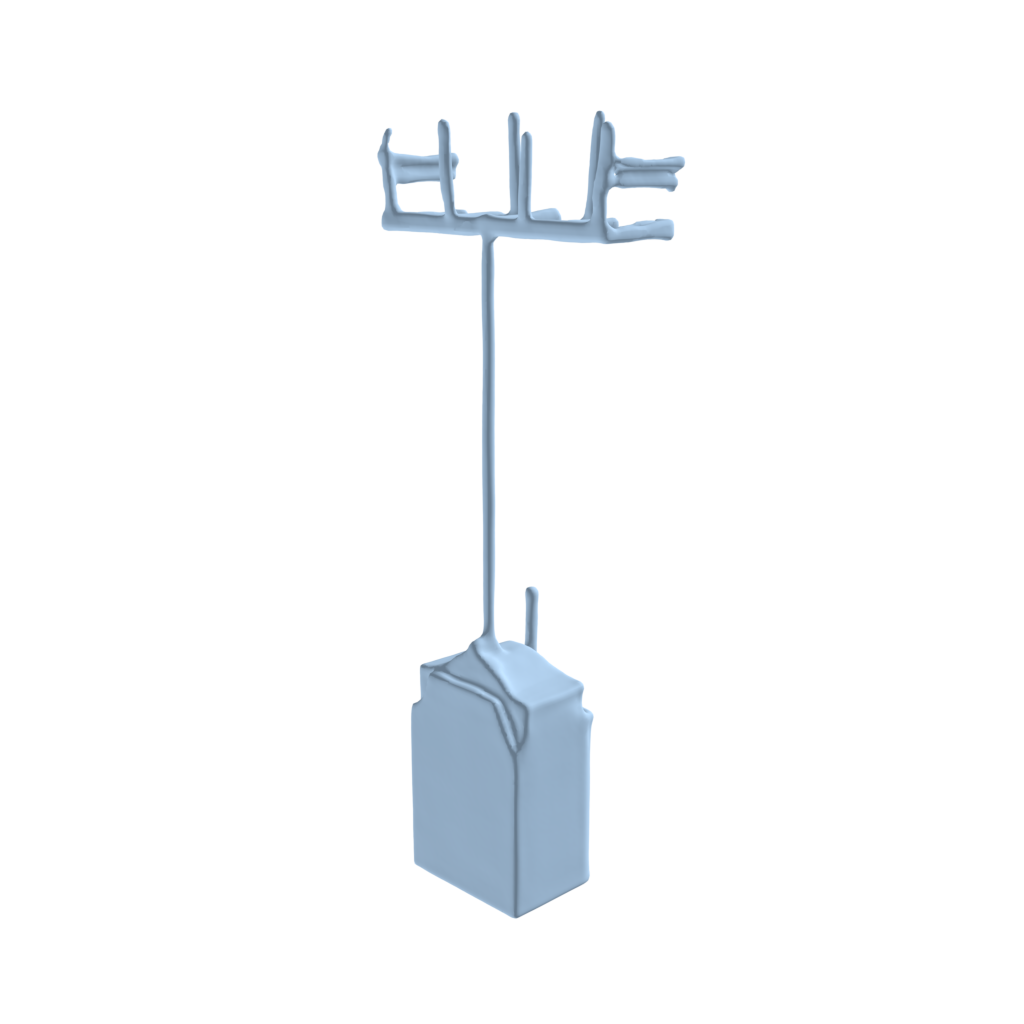}} & \adjustbox{valign=c}{\includegraphics[width=.12\linewidth]{figs/Appendix/GS-Cases/Basket/GS-10-Ref.png}} & \adjustbox{valign=c}{\makecell{Same as $w=7$}} & 7999 \\
\end{tabular}
\caption{\textbf{Guidance scale case study for the prompt ``a geometric sculpture with a trapezoidal base and a rectangular prism suspended above it by rods''.} 
The model reproduces the training shape at lower scales ($w\in\{1,3,5\}$), and emphasizes sub-phrases such as ``trapezoidal base'' and ``rods'', but fails to generate the rectangular prism at larger $w$.}
\vspace{-1em}
\label{tab:gs-basket}
\end{table}
\begin{table}[htbp]
\centering
\tablestyle{5pt}{1.2}
\begin{tabular}{lcccc}
$w$ & gen & top match & caption (top match) & LFD \\
\shline
1 & \adjustbox{valign=c}{\includegraphics[width=.12\linewidth]{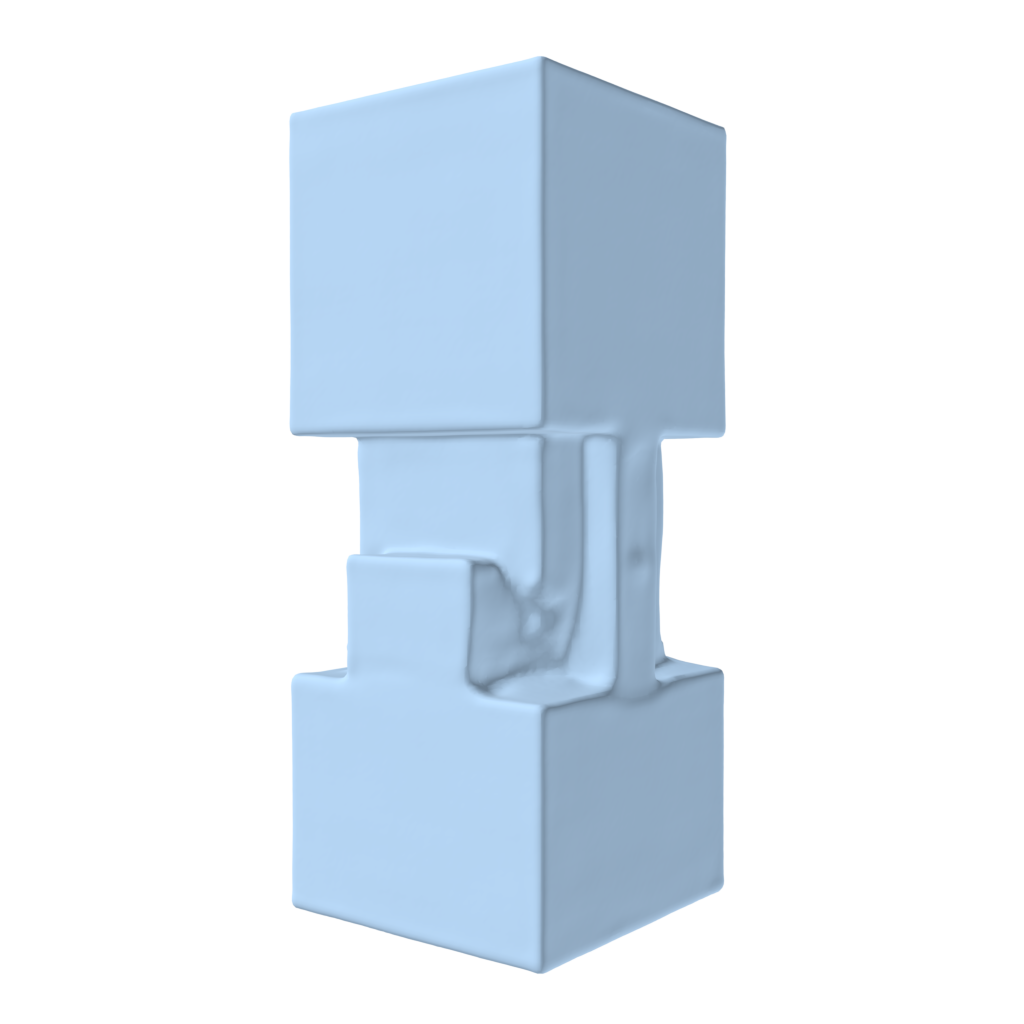}} & \adjustbox{valign=c}{\includegraphics[width=.12\linewidth]{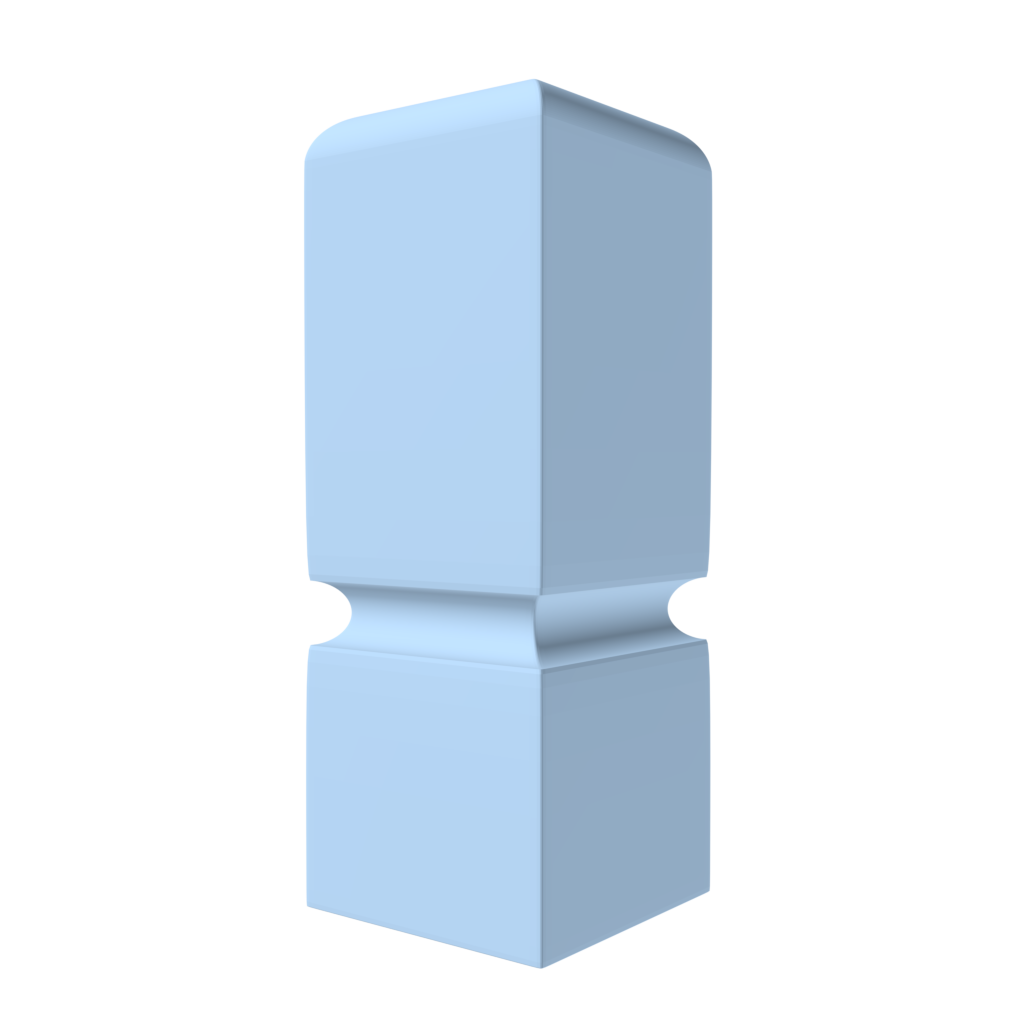}} & \adjustbox{valign=c}{\makecell{``a cube''}} & 1396 \\
3 & \adjustbox{valign=c}{\includegraphics[width=.12\linewidth]{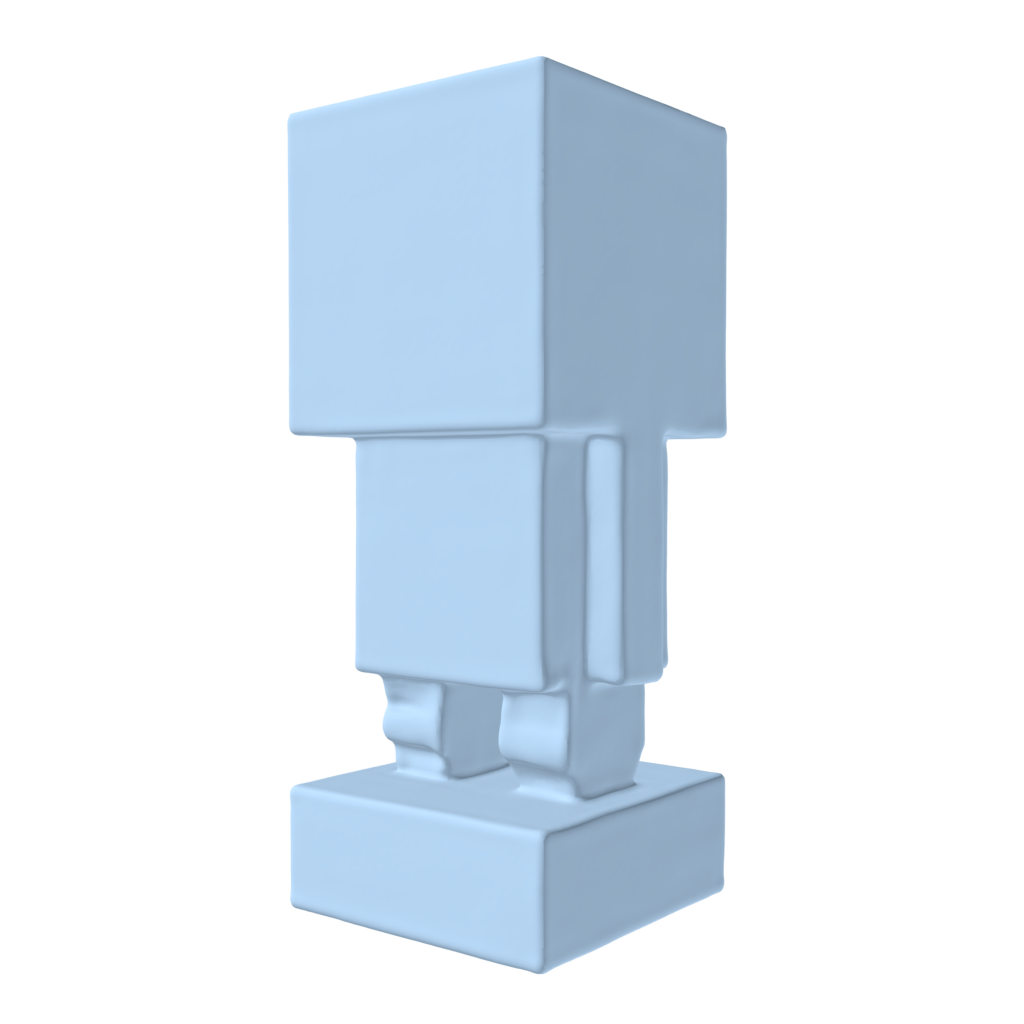}} & \adjustbox{valign=c}{\includegraphics[width=.12\linewidth]{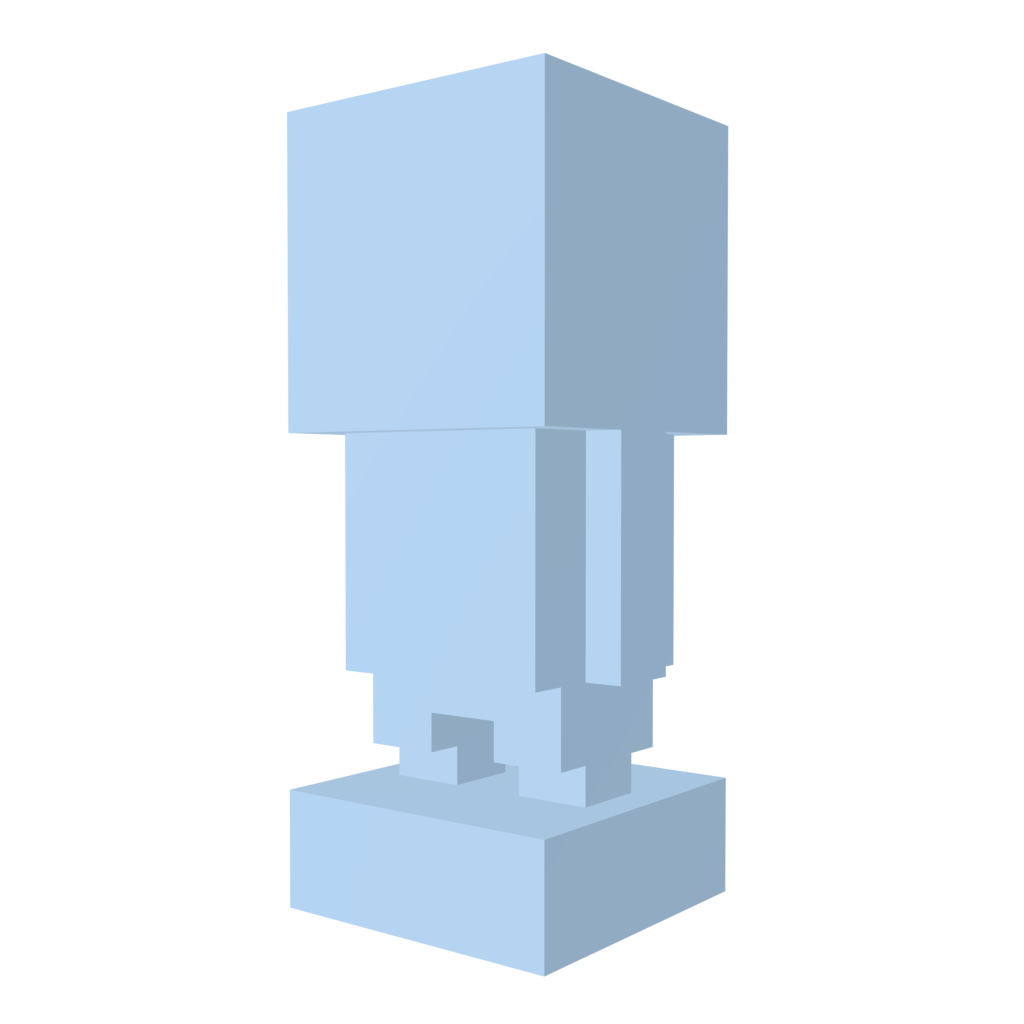}} & \adjustbox{valign=c}{\makecell{``blocky humanoid figure\\ standing on a base''}} & 390 \\
5 & \adjustbox{valign=c}{\includegraphics[width=.12\linewidth]{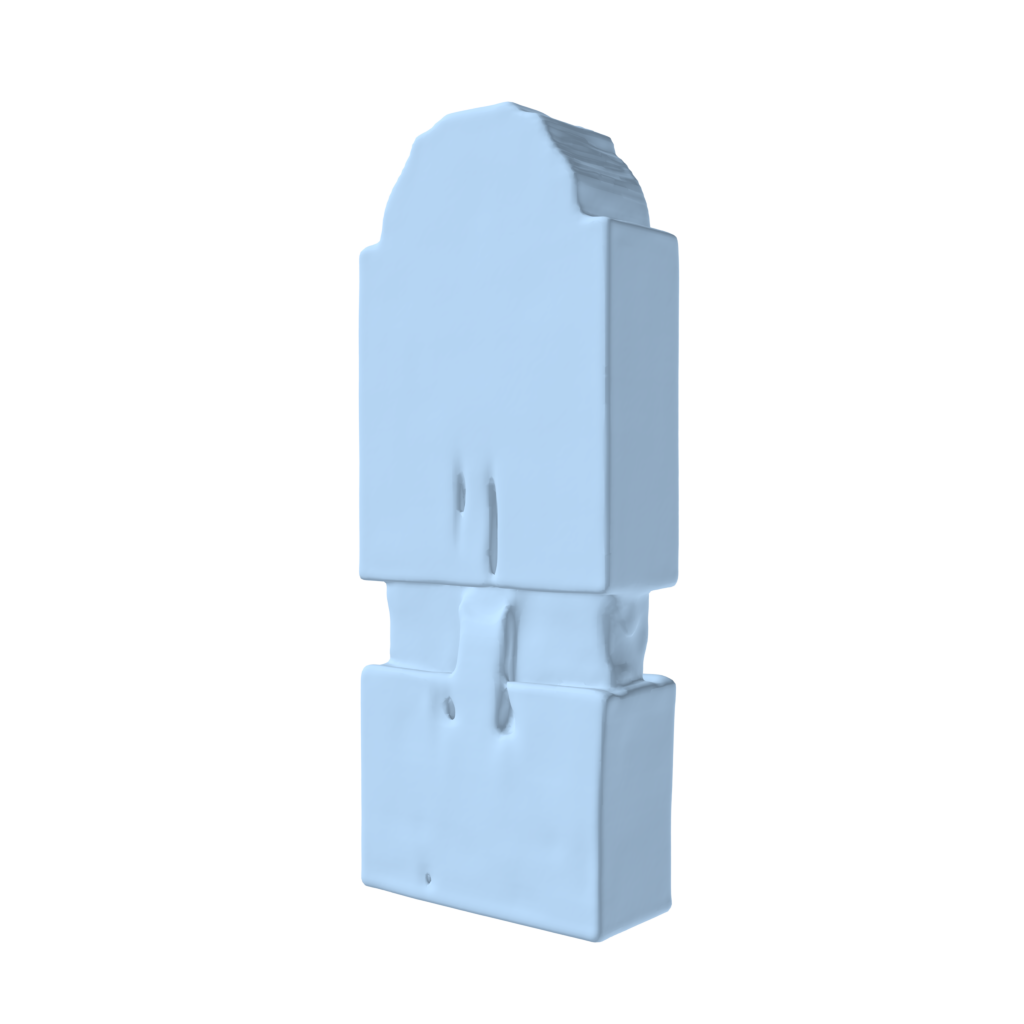}} & \adjustbox{valign=c}{\includegraphics[width=.12\linewidth]{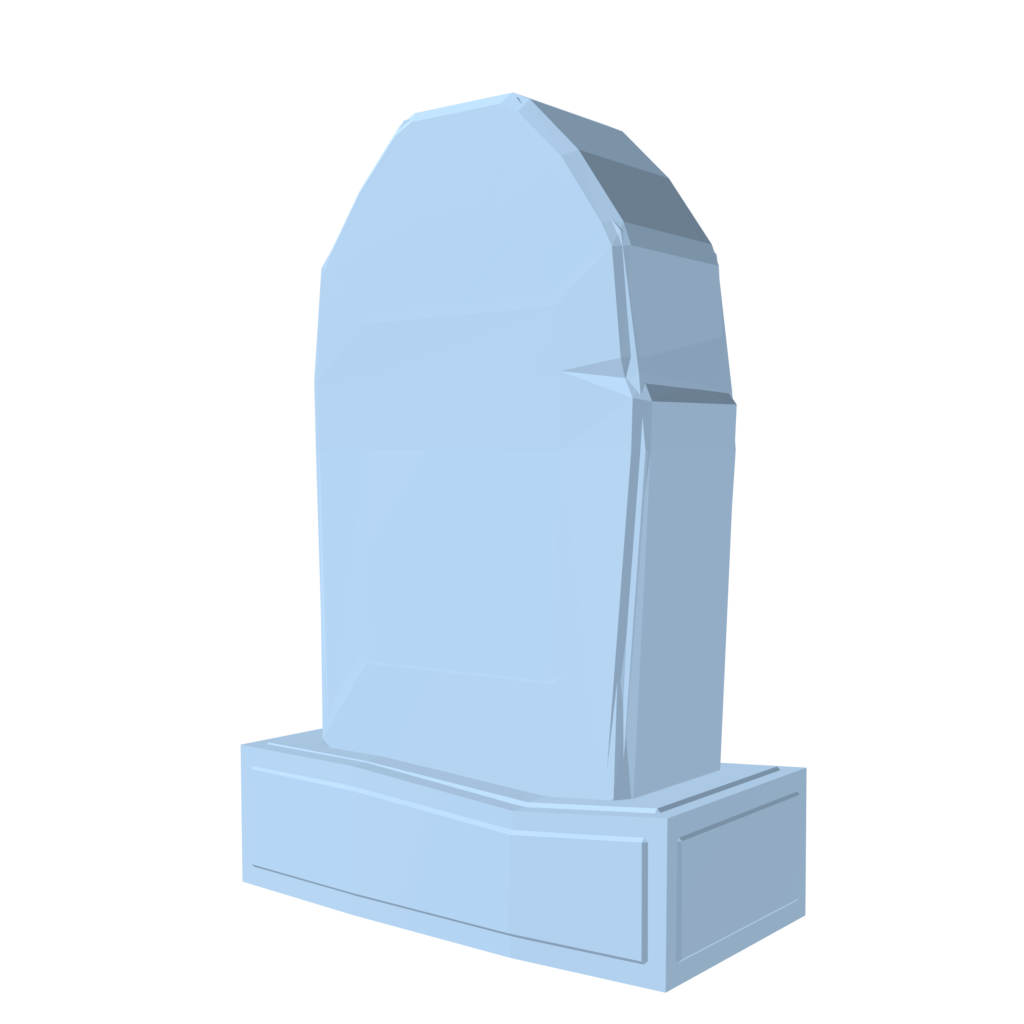}} & \adjustbox{valign=c}{\makecell{``chair with high curved\\ backrest and rectangular base''}} & 1991 \\
7 & \adjustbox{valign=c}{\includegraphics[width=.12\linewidth]{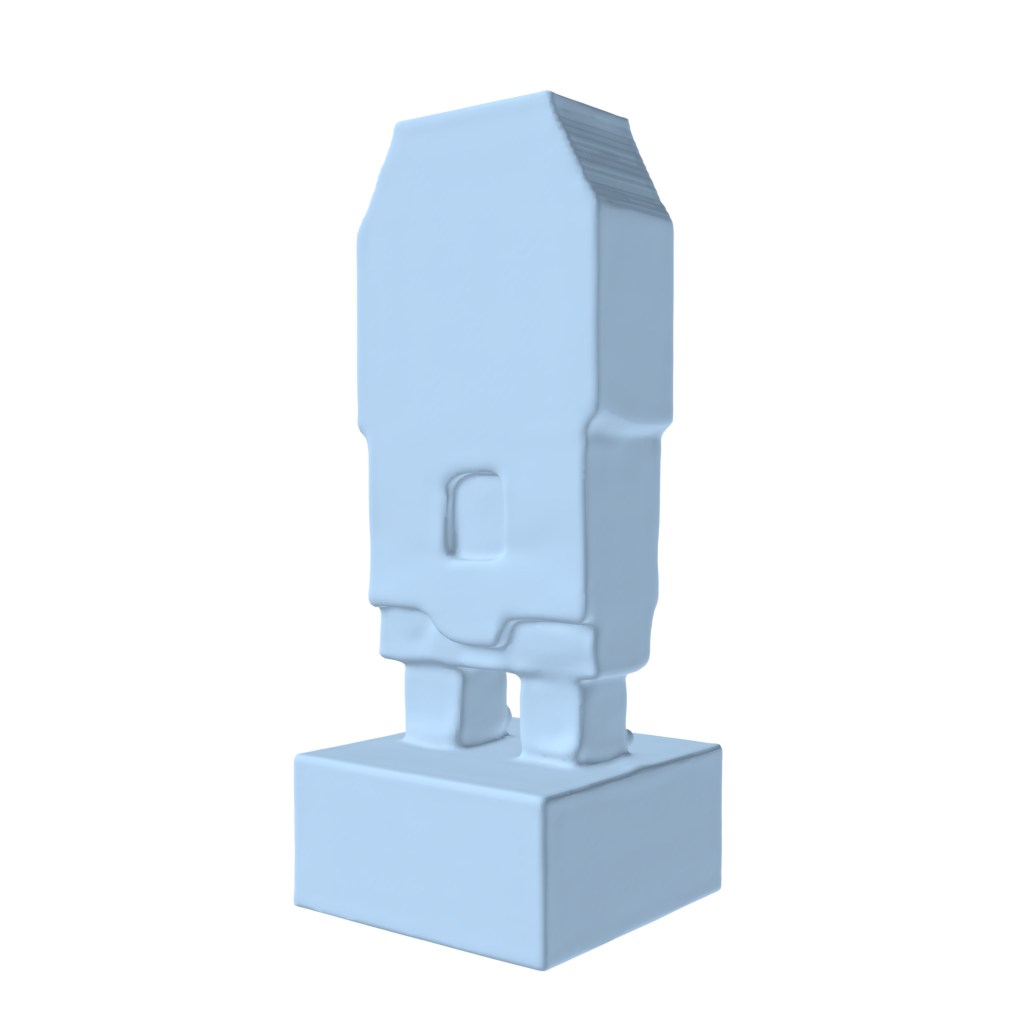}} & \adjustbox{valign=c}{\includegraphics[width=.12\linewidth]{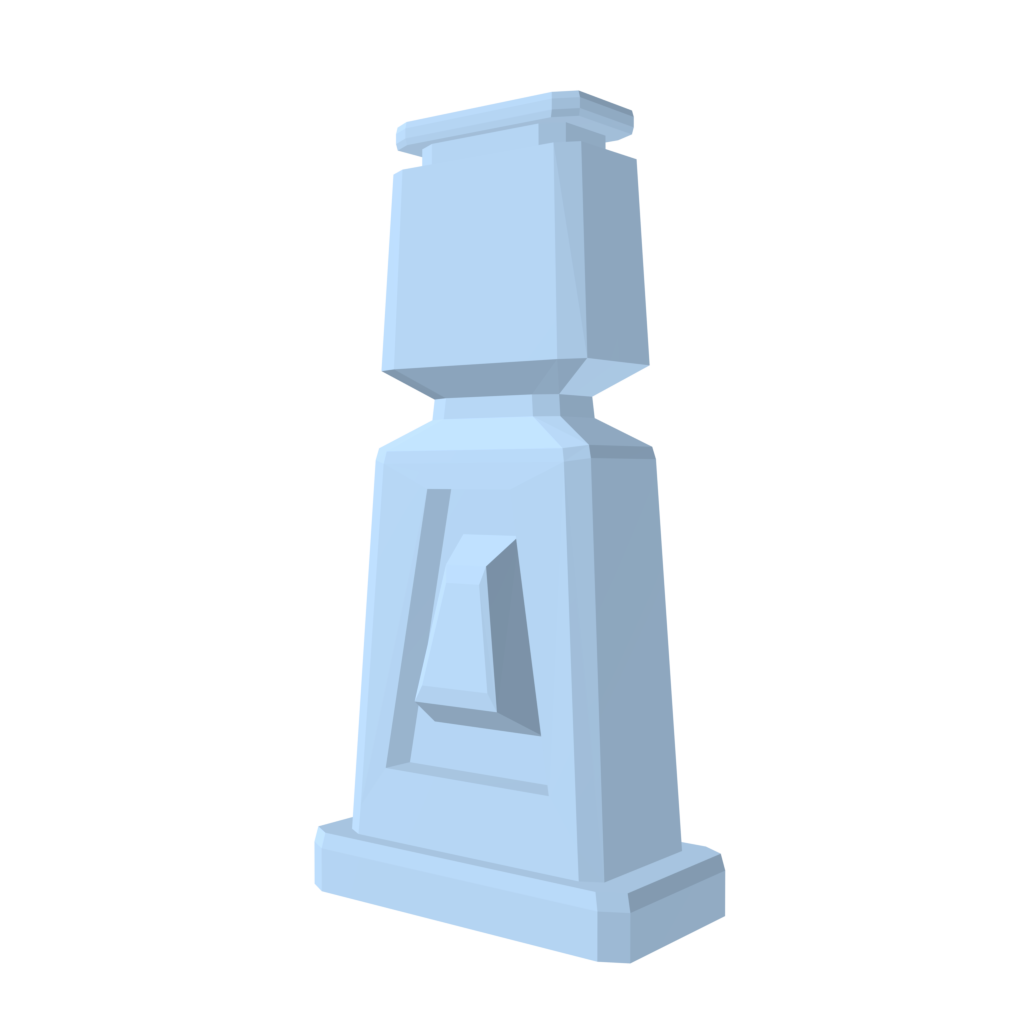}} & \adjustbox{valign=c}{\makecell{``abstract sculpture\\ with rectangular body, square cap,\\ smaller square base,\\ recessed trapezoidal\\ shape on vertical face''}} & 2805 \\
10 & \adjustbox{valign=c}{\includegraphics[width=.12\linewidth]{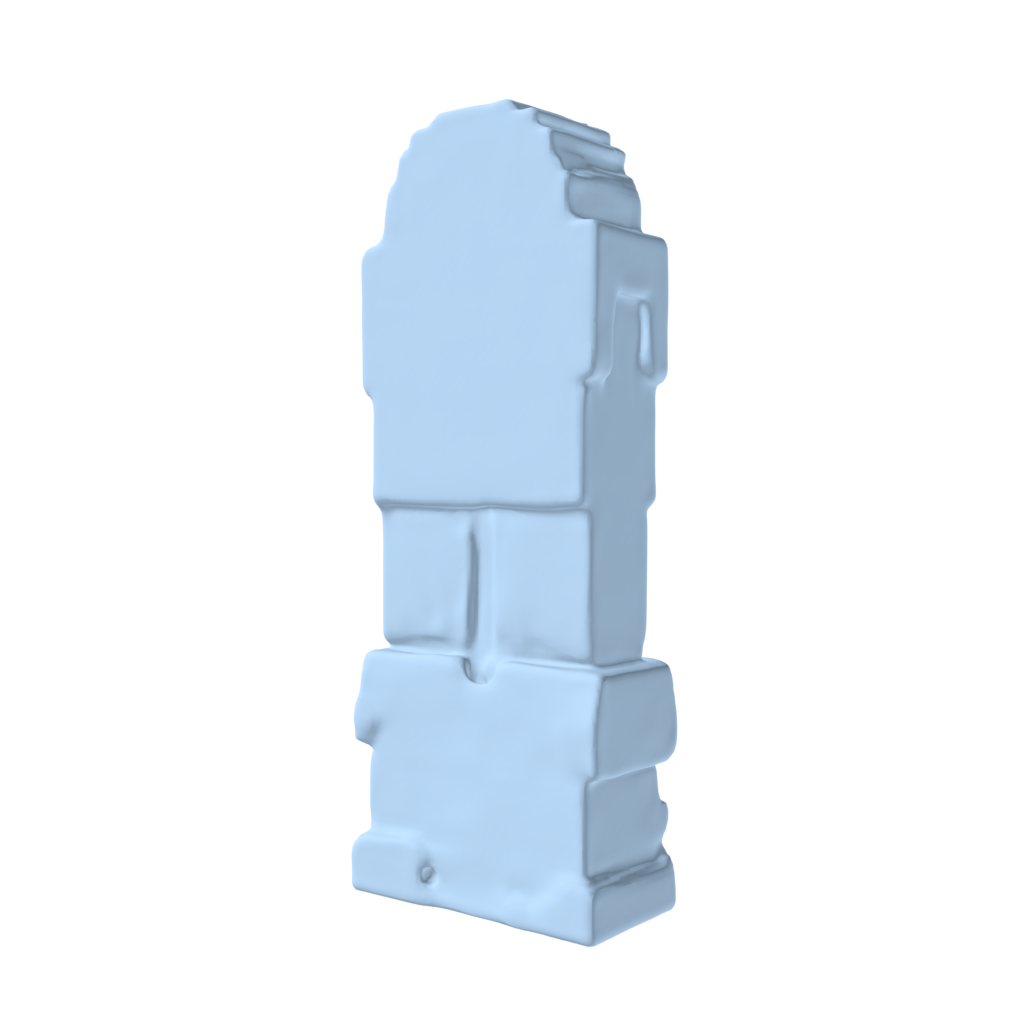}} & \adjustbox{valign=c}{\includegraphics[width=.12\linewidth]{figs/Appendix/GS-Cases/Figurine/GS-7-10-Ref.png}} & \adjustbox{valign=c}{\makecell{Same as $w=7$}} & 2188 \\
\end{tabular}
\caption{\textbf{Guidance scale case study for the prompt ``blocky humanoid figure standing on a base''.} 
The model generates a similar but lower-quality blocky figure with no guidance, reproduces the training shape when $w=3$, and emphasizes sub-phrases such as ``figure standing on a base'', but fails to generate the blocky head with larger $w$.}
\vspace{-1em}
\label{tab:gs-figurine}
\end{table}

Interestingly, in Figure~\ref{fig:gs-app}, as the guidance scale increases from 3 to 7, training FD moves in the opposite direction of $Z_U$, despite the two metrics being generally correlated. One possible explanation is that, in this case, increasing the guidance scale leads to lower diversity and higher similarity to training shapes in the generated outputs. Because FD measures similarity at the distribution level, reduced diversity among generated samples results in a higher training FD. In contrast, $Z_U$ is computed based on object-level similarity and is therefore less impacted by diversity.

\paragraph{Supplementary case studies.} 
Tables~\ref{tab:gs-basket} and~\ref{tab:gs-figurine} present two additional case studies involving different guidance scales. These results are consistent with the discussion in Section~\ref{subsec:GS}. We observe that generated shapes produced with higher guidance scales often fail to fully align with the complete prompt. Although these scales yield novel shapes that are distinct from the training set, we argue that while large guidance scales mitigate memorization, they do so at the cost of reduced prompt alignment.

\subsection{Latent Space}
\label{app:latent}
Vecset is produced by applying cross-attention from a set of latent queries to point cloud positional embeddings. Following 3DShape2VecSet~\citep{3DShape2Vecset}, the queries can be learnable vectors or sub-sampled point queries. Previous studies~\citep{zhang2024clay,chen2025dora} apply different sub-sampling rates to obtain different Vecset lengths. We adopt a similar point query design and compare three sequence lengths: 768, 1024, and 1280.

\begin{table}[htbp]
    \centering
    \small
    \setlength{\tabcolsep}{4.5pt} 
    \begin{tabular}{l l cc}
        \toprule
        latent shape & dataset   & CD {\scriptsize ($\times 10^3$)} $\downarrow$ & F-Score {\scriptsize ($t=0.01$)} $\uparrow$ \\
        \midrule
        $(768\times32)$   & Objaverse & 12.50 & 0.86 \\
        $(1024\times32)$  & Objaverse & 12.27 & 0.86 \\
        $(1280\times32)$  & Objaverse & \textbf{12.09} & \textbf{0.87} \\
        \bottomrule
    \end{tabular}
    \caption{
        \textbf{Reconstruction performance of VecSetX} trained on the entire Objaverse. Different Vecset lengths result in similar performance in terms of both CD and F-Score.
    }
    \vspace{-1em}
    \label{tab:vaeperformance}
\end{table}

\paragraph{Vecset autoencoder performance.} In Table~\ref{tab:vaeperformance}, we evaluate the autoencoder's reconstruction performance. Varying the Vecset length does not significantly affect reconstruction quality: all three lengths yield comparable performance.

\begin{figure}[h]
\centering
\setlength{\tabcolsep}{1.2pt}
\renewcommand{\arraystretch}{-0.5}
  \begin{tabular}{cccc}
    train & 768 & 1024 & 1280 \\
    \includegraphics[width=.25\linewidth]{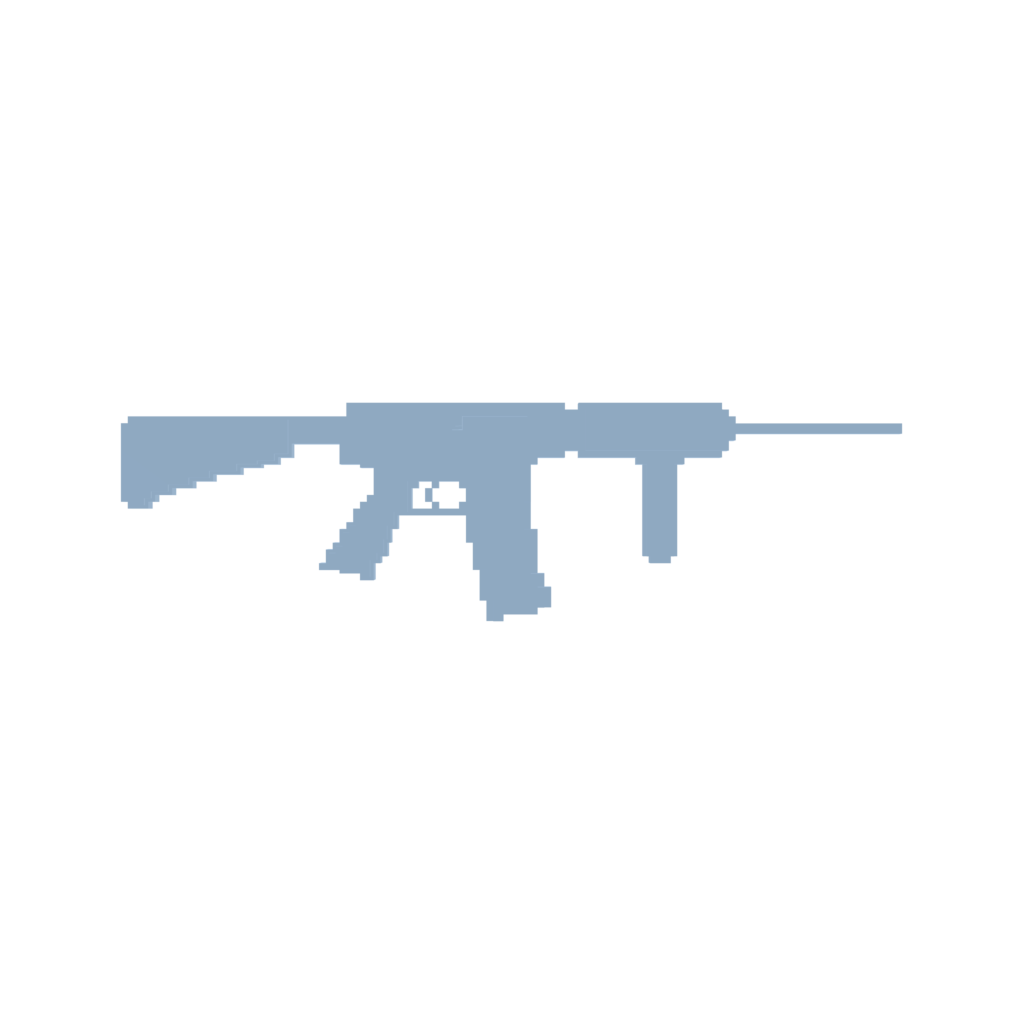} & 
    \includegraphics[width=.25\linewidth]{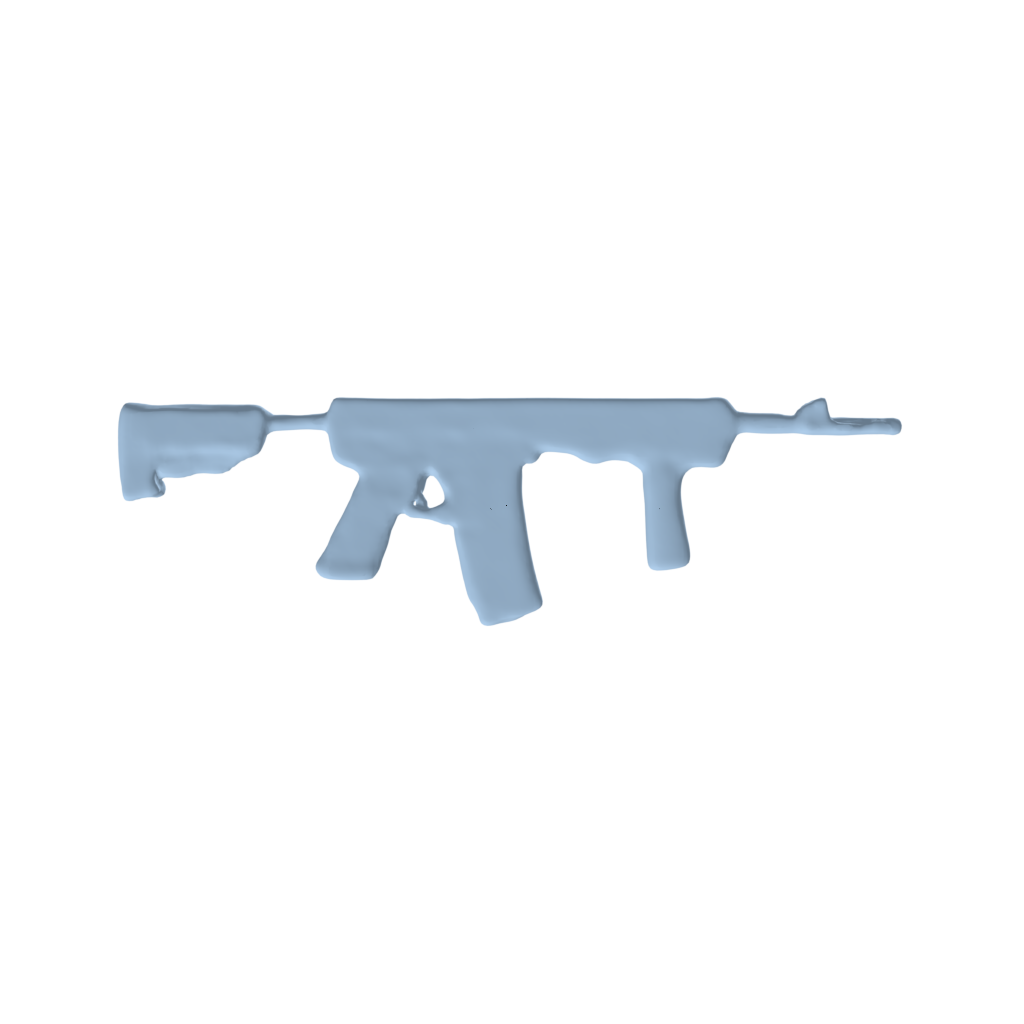} & 
    \includegraphics[width=.25\linewidth]{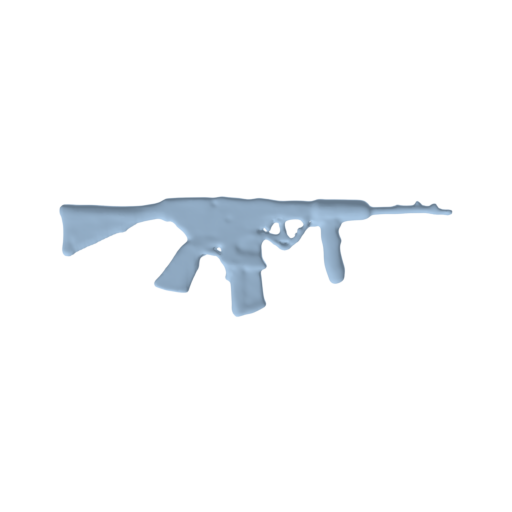} & 
    \includegraphics[width=.25\linewidth]{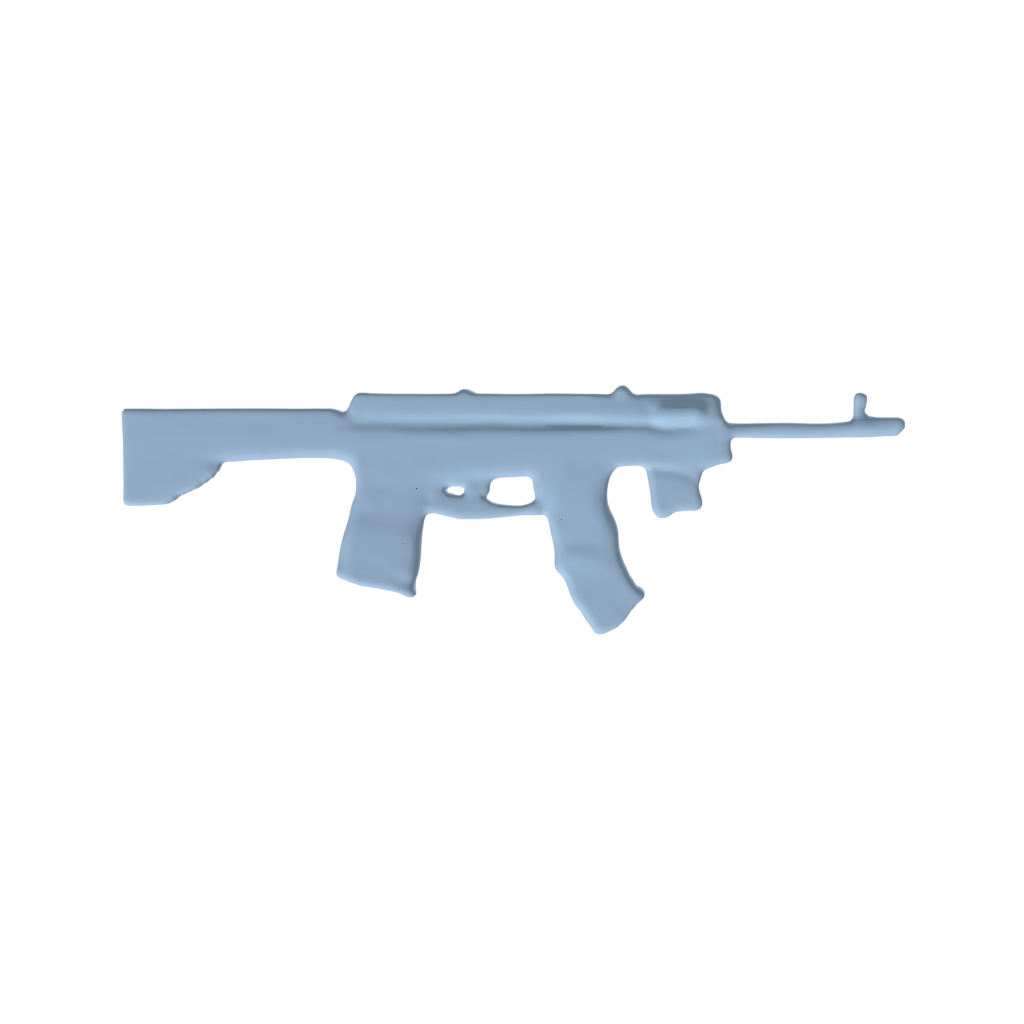} \\
  \end{tabular}
  \vspace{-0.25em}
  \caption{\textbf{Shapes generated with different Vecset lengths for the prompt ``an assault rifle with a stock, foregrip, and pistol grip; a body; and a long barrel''.} Vecsets of longer sequence lengths (1024 and 1280) generate high-quality shapes aligned with the training prompt, while exhibiting novel features (\eg, stocks and foregrips) that are different from the training shape.}
  \label{fig:latent_rifle}
\end{figure}

\begin{figure}[h]
\centering
\setlength{\tabcolsep}{1.2pt}
\renewcommand{\arraystretch}{-0.5}
  \begin{tabular}{cccc}
    train & 768 & 1024 & 1280 \\
    \includegraphics[width=.25\linewidth]{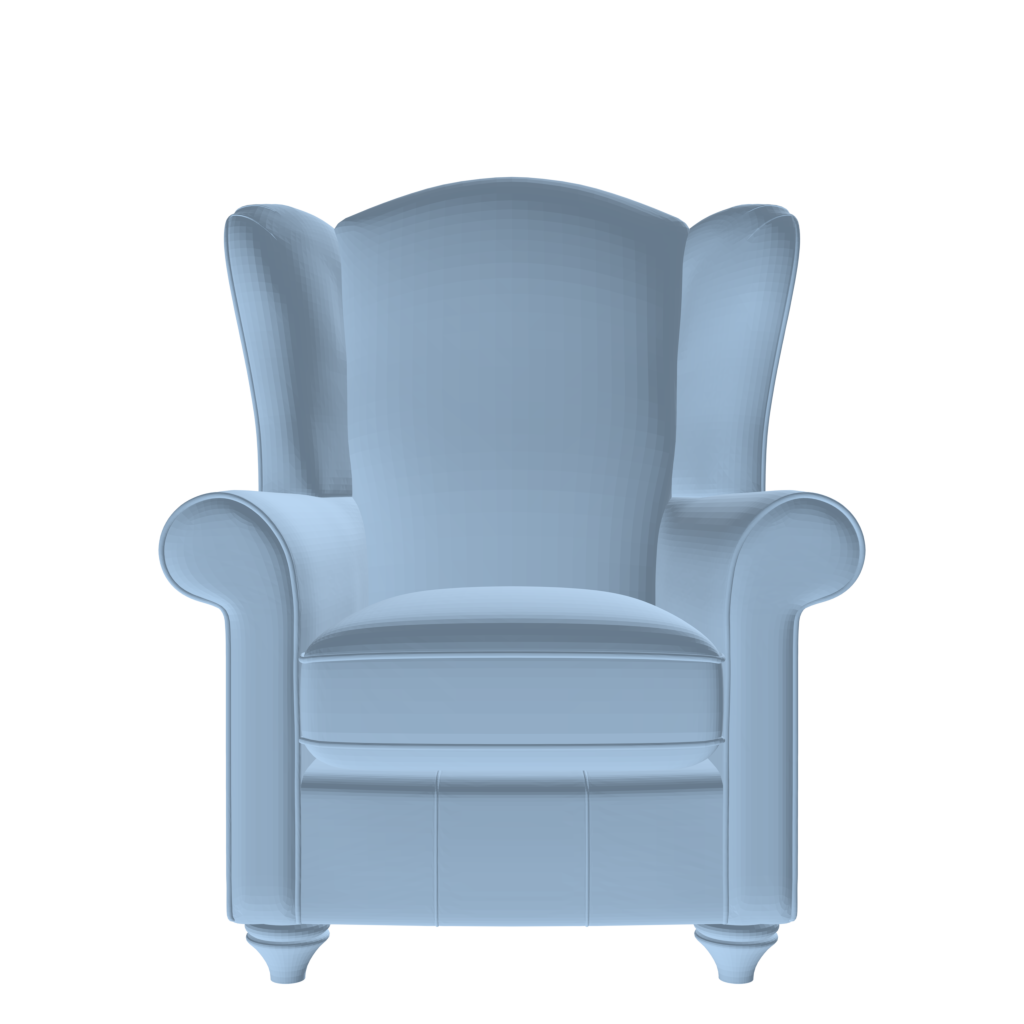} & 
    \includegraphics[width=.25\linewidth]{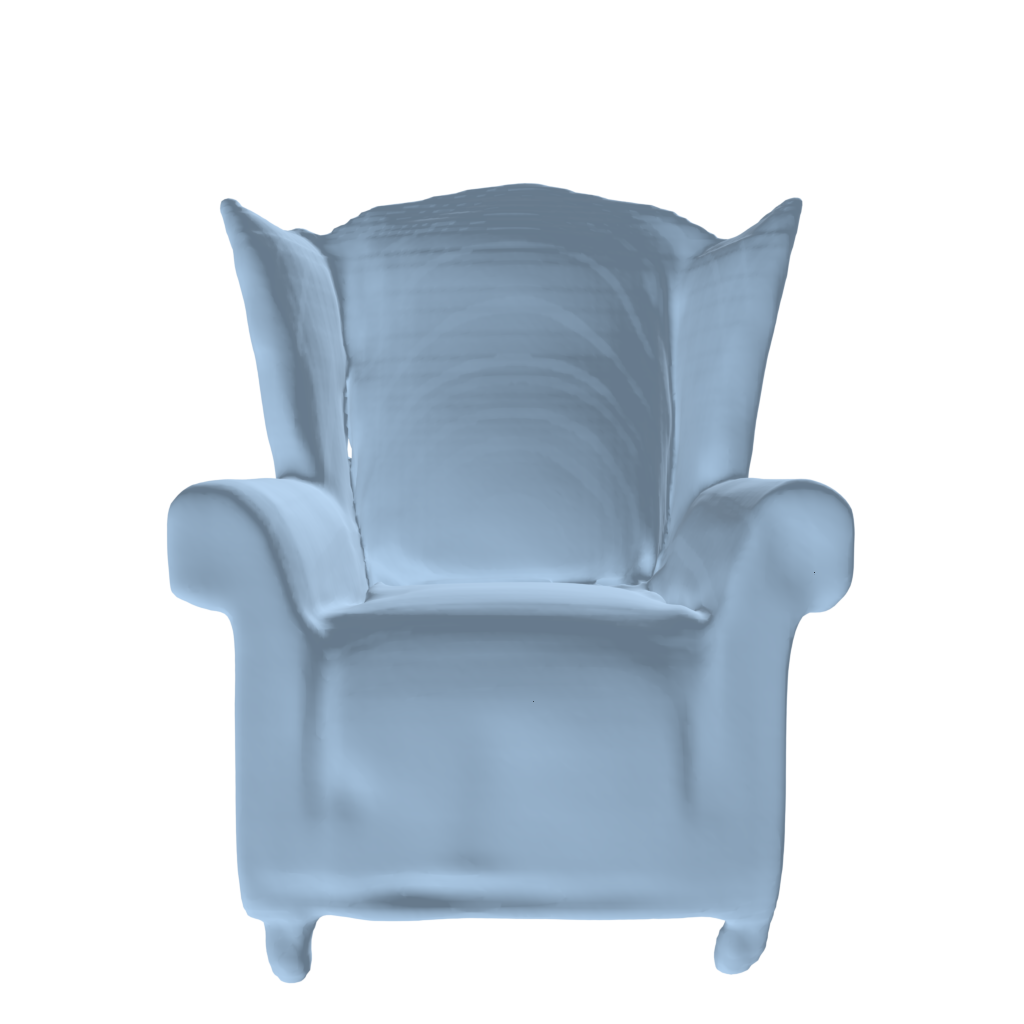} & 
    \includegraphics[width=.25\linewidth]{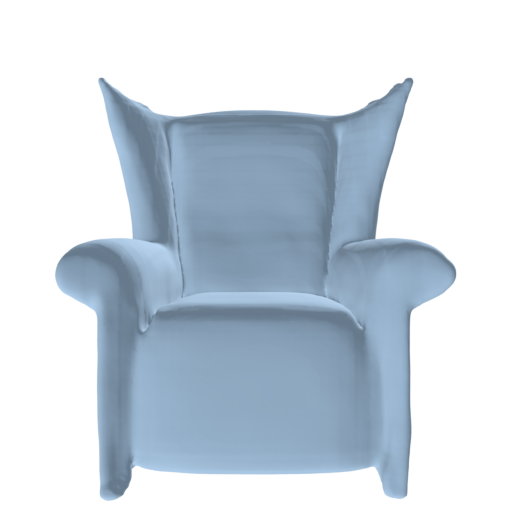} & 
    \includegraphics[width=.25\linewidth]{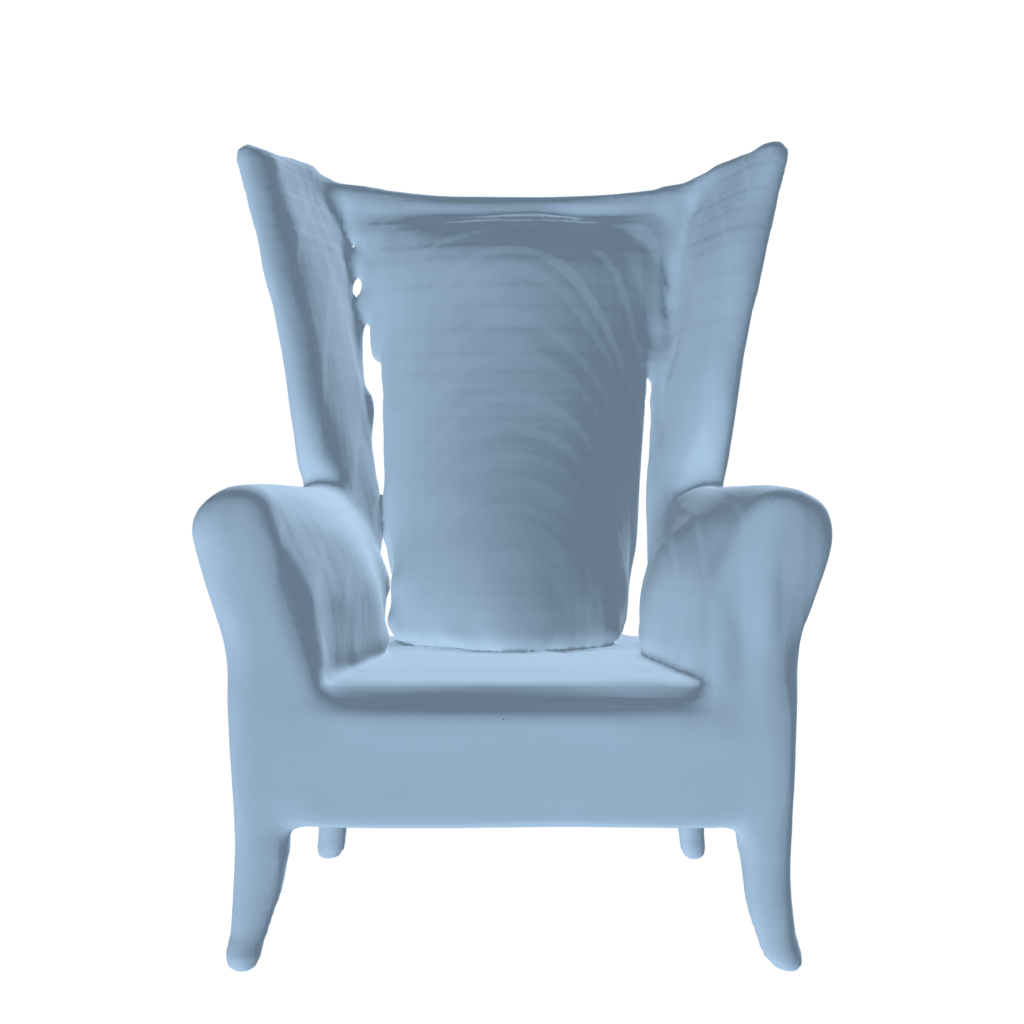} \\
  \end{tabular}
  \vspace{-0.25em}
  \caption{\textbf{Shapes generated with different Vecset lengths for the prompt ``wing chair''.} Vecsets of longer sequence lengths (1024 and 1280) generate high-quality shapes that are well aligned with the training prompt, while exhibiting novel features (\eg, arms and wings) that are different from the training shape.}
    \vspace{-1em}
  \label{fig:latent_chair}
\end{figure}

\paragraph{Supplementary case studies.} Figures~\ref{fig:latent_rifle} and~\ref{fig:latent_chair} present two additional case studies with different Vecset lengths. The results are consistent with the discussion in Section~\ref{subsec:latent}: shapes produced with longer Vecsets maintain high fidelity while showing reduced memorization.


\subsection{Rotation Augmentation}
\label{app:rot_aug}
\paragraph{Rotation invariance of distance metrics.} We evaluate the robustness of LFD and Uni3D by measuring the distance between an object's original pose and its rotated poses under each metric. Specifically, we apply random rotations sampled from $[0^{\circ}, 360^{\circ}]$ independently along the pitch (x), yaw (y), and roll (z) axes. If a metric is rotation invariant, the distance between the original and the rotated shapes should remain near zero. 

\begin{figure}[htbp]
    \centering
    \includegraphics[width=\linewidth]{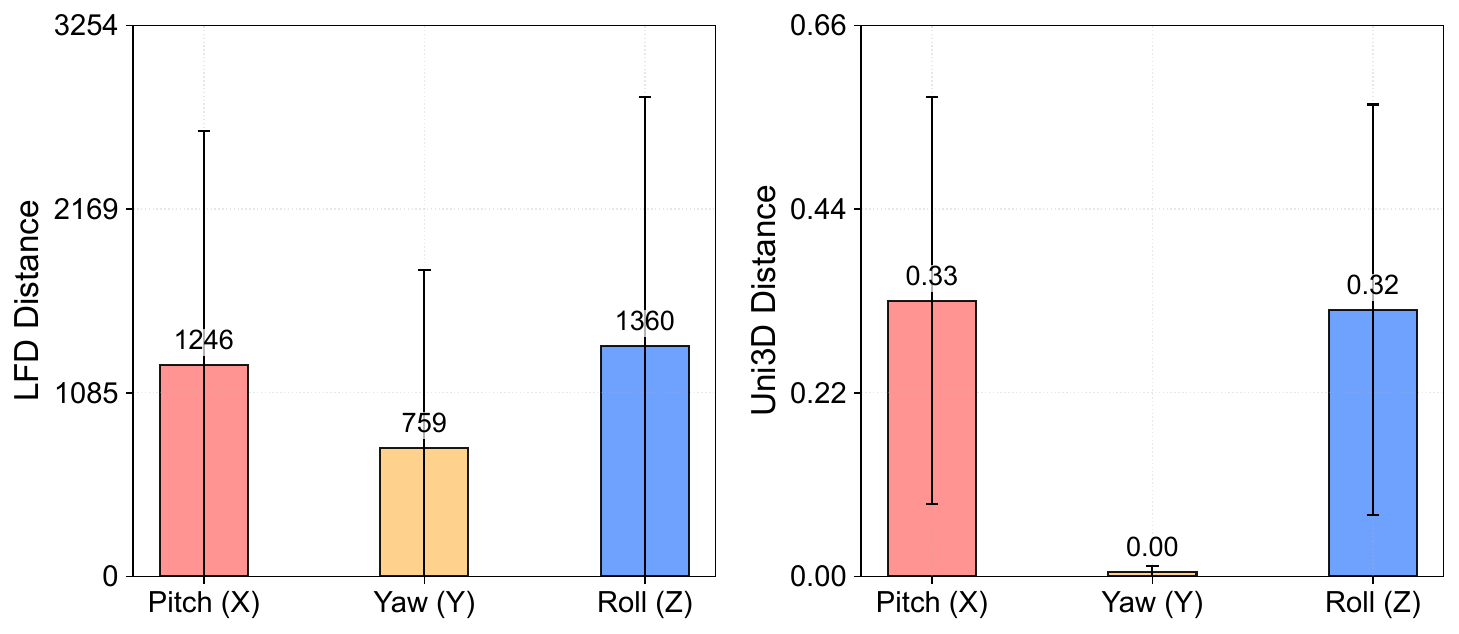}
    \caption{\textbf{Robustness of distance metrics to rotation.} Mean and standard deviation of LFD (left) and Uni3D (right) distances between original and randomly rotated objects. LFD exhibits significant variation across all rotation axes, indicating that it is not rotation-invariant. Uni3D demonstrates strong robustness (\ie, low distances) specifically under yaw rotation.}
    \label{fig:rot}
\end{figure}

As illustrated in Figure~\ref{fig:rot}, LFD is sensitive to rotation, with high distances between original and rotated objects for all axes. 
In contrast, objects rotated along the yaw axis have a near-zero distance to the original objects under Uni3D.

\paragraph{Rotation implementation.} 
To guarantee the accuracy of $Z_U$ and FD, we limit our experiments to yaw rotation. In our dataset, the y-axis corresponds to the yaw axis. Consequently, we restrict our rotations to four discrete angles around the y-axis ($0^\circ$, $90^\circ$, $180^\circ$, and $270^\circ$) to ensure Uni3D remains consistent. For LFD, we compute distances across these four distinct poses for each generated sample and report the minimum retrieval distance.

\section{Rendering Details}
\label{app:rendering}

We use Blender to render $N=12$ views for each object at $256 \times 256$ resolution. Prior to rendering, the geometry is normalized to fit within a unit cube centered at the origin. Camera poses are sampled uniformly from a spherical shell with radii $r \in [1.5, 2.2]$, oriented to face the object center. Lighting is similarly randomized for each view using an area light with varying energy and position. To ensure alignment across the dataset, we use a fixed random seed, resulting in identical camera trajectories and lighting conditions for every object.

\end{document}